%% file: main.tex
\pdfoutput=1
\documentclass[10pt,logo,copyright]{nvidiatechreport}
\input{packages}

\input{common.tex}

\definecolor{darkred}{rgb}{0.7, 0.0, 0.0}

\usepackage{pifont}
\usepackage{wrapfig}
\renewcommand{\today}{}
\usepackage[nameinlink]{cleveref}
\crefname{equation}{Eq.}{Eqs.}
\crefname{figure}{Fig.}{Figs.}
\crefname{section}{Sec.}{Sec.}
\crefname{appendix}{App.}{App.}
\crefname{table}{Tab.}{Tabs.}
\crefname{algorithm}{Algo}{Algo}
\crefname{thm}{Thm}{Thm}
\Crefname{thm}{Thm}{Thm}
\crefname{prop}{Prop}{Prop}
\usepackage{longtable}
\usepackage{wrapfig}
\usepackage{caption}
\usepackage{makecell}

\usepackage{booktabs}
\usepackage{multirow}
\usepackage{graphicx}
\usepackage{wrapfig}
\usepackage{caption}
\usepackage{enumitem}

\usepackage{xcolor}
\usepackage{hyperref} 
\usepackage{tikz}
\usetikzlibrary{trees} 
\usepackage[edges]{forest}
\usepackage[most]{tcolorbox}
\definecolor{nvidiaGreen}{HTML}{9dca63}
\definecolor{nvidiaGreenDark}{HTML}{4f8a10}
\definecolor{nvidiaGreenSoft}{HTML}{f3f9ec}
\newcommand{\crefnames}[3]{%
  \@for\next:=#1\do{%
    \expandafter\crefname\expandafter{\next}{#2}{#3}%
  }%
}

\usepackage{subcaption}
\tcbuselibrary{listings,breakable}
\usepackage{enumitem}

\newtcblisting{promptbox}[1]{%
  listing only, breakable, enhanced jigsaw,
  colback=nvidiaGreenSoft,
  colframe=nvidiaGreenDark,
  coltitle=white,
  colbacktitle=nvidiaGreenDark,
  title={#1}, fonttitle=\bfseries\small,
  before skip=4pt, after skip=4pt,
  top=2pt, bottom=2pt, left=4pt, right=4pt,
  listing options={
    basicstyle=\footnotesize\ttfamily,
    breaklines=true, breakatwhitespace=true,
    columns=fullflexible, keepspaces=true,
    aboveskip=0pt, belowskip=0pt,
    upquote=true,
  },
}

\definecolor{midnightgreen}{rgb}{0.0, 0.29, 0.33}
\definecolor{deepgreen}{HTML}{0aa344}
\definecolor{deeppurple}{HTML}{7030a0}
\definecolor{deepblue}{HTML}{171d91}
\definecolor{brown}{HTML}{843c0c}
\definecolor{shadered}{HTML}{ffe5e5}
\definecolor{shadegreen}{HTML}{e5f7ed}
\definecolor{msftBlack}{RGB}{0,0,0}
\definecolor{lightred}{RGB}{255,163,163}
\definecolor{deepred}{RGB}{146,0,0}

\newtcolorbox{boxL}{
    fontupper = \color{black},
    rounded corners,
    arc = 6pt,
    colframe = black!50, 
    boxrule = 0pt, 
    bottomrule = 4.5pt ,
    breakable,
}

\title{\textsc{VoLo}: A Physical Orchestrator for Open-\textsc{Vo}cabulary \textsc{Lo}ng-Horizon Manipulation}

\author{Siyi Chen$^{1,2}$, ~~Hugo Hadfield$^{1}$, ~~Alex Zook$^{1}$, ~~Mikaela Angelina Uy$^{1}$, 
~~Chan Hee Song$^{1}$, ~~Erwin Coumans$^{1}$, ~~Xuning Yang$^{1}$, ~~Faisal Ladhak$^{1}$, ~~Qing Qu$^{2}$, 
~~Stan Birchfield$^{1}$, ~~Jonathan Tremblay$^{1}$$^{\dagger}$, ~~Valts Blukis$^{1}$$^{\dagger}$ \\
\textsuperscript{1}NVIDIA \quad  \textsuperscript{2}University of Michigan \quad
$^{\dagger}$Project Leads
}

\begin{abstract}
Open-vocabulary long-horizon manipulation requires robots to reason over flexible instructions and complex multi-object scenes while adaptively planning, executing, monitoring, and recovering from failures. We address these demands with a closed agent loop in which a VLM orchestrates heterogeneous robot capabilities as interruptible tools. Unlike in virtual AI agents, the timing of decisions, actions and tool calls is important in a physical world that does not pause for reasoning. We refer to this setting as \emph{Physical Orchestration}, and propose \textsc{VoLoAgent},
a VLM that plans, monitors, and recovers by treating a VLA/WAM as an interruptible tool it steers mid-rollout alongside vision models and action primitives. To evaluate these long-horizon capabilities, we introduce \textsc{RoboVoLo}, a high-fidelity benchmark for open-vocabulary long-horizon manipulation across common sense, memory/state tracking, complex references, and world knowledge, with both task-level success and failure-mode diagnostics. Experiments show \textsc{VoLoAgent} substantially outperforms single VLA/VLM or tool-based systems, with validation on real-robot experiments. Project page: \url{https://chicychen.github.io/VoLo/}
\end{abstract}

\begin{document}

\maketitle

\vspace{-.35in}
\begin{figure}[!h]
  \centering
  \includegraphics[width=\linewidth]{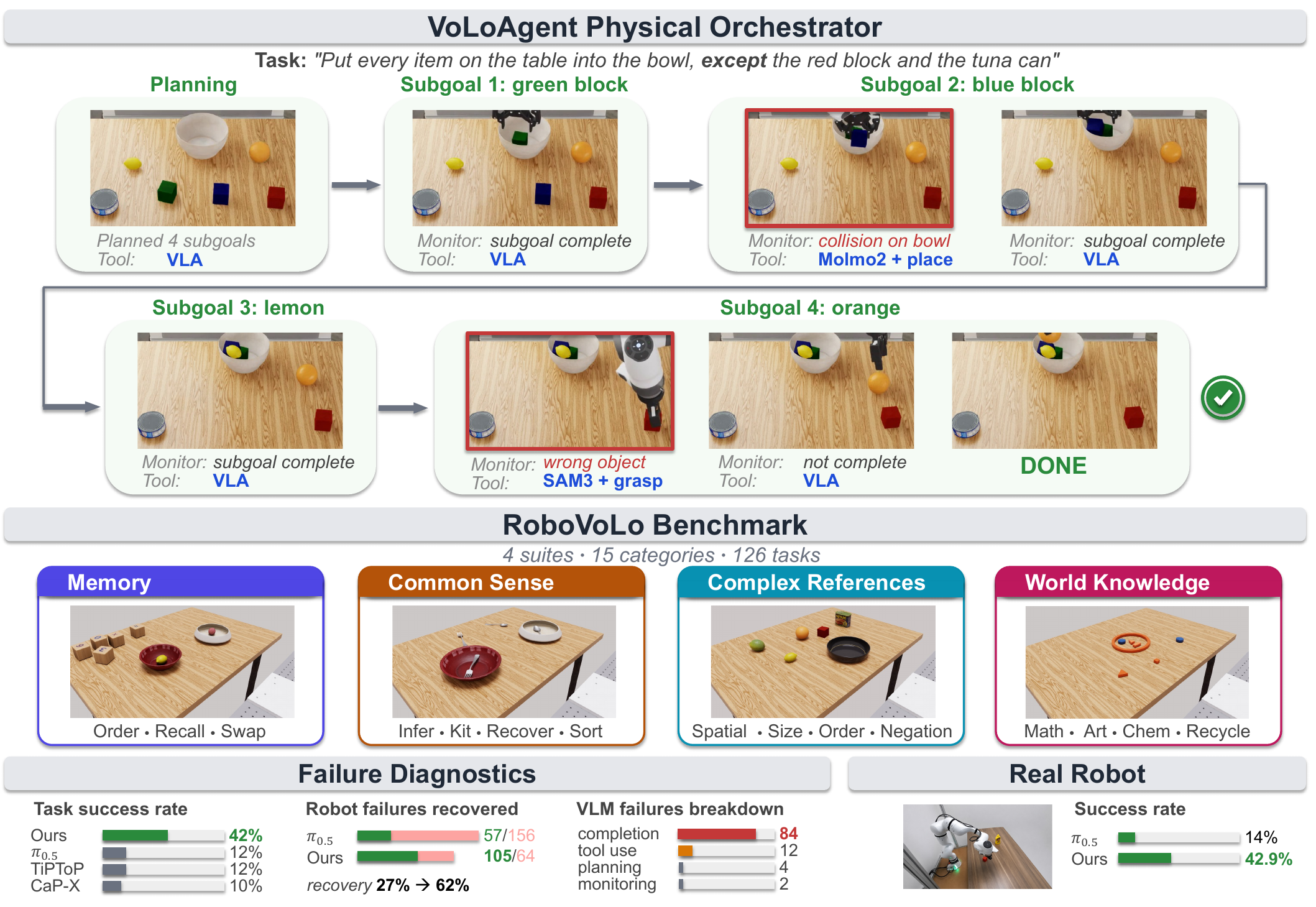}
  \vspace{-.2in}
  \caption{\textbf{\textsc{VoLo} overview.} \textsc{VoLoAgent} plans, monitors (e.g., subgoal complete), and uses tools (e.g., VLA, SAM3) to act and recover from failures (e.g., wrong object). \textsc{RoboVoLo} is a high-fidelity benchmark for evaluating and diagnosing open-vocabulary long-horizon manipulation.
  }
  \label{fig:teaser}
\end{figure}

\abscontent

\vspace{-.4in}
\section{Introduction}

Real-world manipulation is often open-vocabulary and long-horizon, rather than a template pick-and-place task. As illustrated in Fig.~\ref{fig:teaser}, when asked to ``put every item on the table into the bowl, except the red block and the tuna can,'' a robot must understand negative references such as ``except,'' plan over a sequence of objects, monitor whether each subgoal succeeds, and recover from failures such as picking the wrong object. These open-vocabulary long-horizon tasks require high-level capabilities including planning, reasoning over complex language, using world knowledge, spatial reasoning, and maintaining memory of the evolving scene. At the same time, they demand reliable embodied perception and precise low-level action skills.

Existing manipulation approaches only partially address these challenges. End-to-end vision-language-action (VLA) models~\citep{pi05,pistar06,pi07} and world action models (WAMs)~\citep{gr2,dreamzero,dreamdojo,cosmospolicy,lingbotva} exhibit precise manipulation, but lack robust planning, monitoring, and perception in the multi-object scenes typical of long-horizon tasks. LLM/VLM-driven code-as-policy methods~\citep{codeaspolicy,progprompt,capx}, including tool-augmented variants~\citep{spacetools}, support explicit reasoning over perception and classical control primitives, but are limited by fixed toolsets and control APIs for contact-rich manipulation, while largely overlooking monitoring and recovery. Recent hierarchical systems pair a VLM planner with a VLA executor~\citep{hamster,hirobot,agenticrobot,goal2skill,failsafe,criticloop}, but usually hard-wire this control flow rather than adaptively composing VLA/WAMs with perception, action, monitoring, and recovery tools. In short, the VLA is treated as a fixed executor rather than one interruptible capability among many.

We instead approach open-vocabulary long-horizon manipulation as \emph{physical orchestration}: unlike a virtual agent, which can pause the world while it thinks, a physical agent must decide \emph{when} to act, advance, or stop against a world that keeps moving (Sec.~\ref{sec:system:loop}).
We present \textsc{VoLoAgent}, an instantiation of this idea that unifies a VLA/WAM with perception models and grasp/place primitives as callable tools in a flexible VLM-managed agent loop, and outperforms hard-wired pipelines.


To study this regime, we introduce \textsc{RoboVoLo}, a high-fidelity benchmark for open-vocabulary long-horizon manipulation built on RoboLab~\citep{robolab}. Existing benchmarks~\citep{libero,robosuite,robolab,molmospaces2026} often focus on short-horizon skills, overlook open-vocabulary reasoning, or use simplified scenes, leaving limited room to study long-horizon state tracking and adaptive recovery. \textsc{RoboVoLo} spans four suites: common sense, memory, complex references, and world knowledge, comprising 15 task categories and 126 tasks in total. Comprehensive experiments show that \textsc{VoLoAgent} substantially outperforms standalone action models, code-as-policy systems, and TAMP-style baselines (\emph{i.e.,} task and motion planning). We further analyze both robot-level failures, such as wrong-object picks and stuck behavior, and VLM-level failures, such as planning mistakes, missed failure detection, and tool-use errors, to diagnose the strengths and limitations of tool-augmented robotic agents. Finally, we validate our findings on real Franka manipulation tasks, showing that orchestration substantially improves over a standalone action model.

We make the following contributions:

\begin{enumerate}[leftmargin=*]
    \item \textsc{VoLoAgent}, an adaptive tool-augmented robotic agent that uses a VLM to plan, reason, monitor, and recover by composing an interruptible VLA/WAM with perception models and classical action primitives callable tools in a single closed loop.
    \item \textsc{RoboVoLo}, a high-fidelity benchmark with 126 tasks for open-vocabulary long-horizon manipulation, spanning common sense, memory, references, and world knowledge, designed independently of the system.
    \item A large-scale empirical study comparing action models, code-as-policy systems, TAMP-style systems, and ablations of \textsc{VoLoAgent} orchestrator, complemented by real robotic experiments.
\end{enumerate}

\section{Related Work}

\paragraph{Vision-Language-Action and World Action Models.} End-to-end VLAs map observations and instructions directly to robot actions, achieving
strong dexterity at scale ~\citep{rt2,vima,openvla,rdt1b,pi05,pistar06,pi07,molmoact,molmoact2,molmobot}; world action models
(WAMs) extend this line by jointly predicting future video and actions~\citep{gr2,dreamzero,dreamdojo,cosmospolicy,lingbotva}. Recent variants interleave explicit reasoning, dual-system architectures, chain-of-thought planning, or depth-aware spatial tokens~\citep{molmoact,pi05,hirt}, and some push memory inside the policy via memory banks~\cite{memoryvla} or multi-frame chunking~\cite{cronusvla}. However, their action chunks still execute largely open-loop, limiting planning, reasoning, real-time monitoring, and tool-based recovery during execution. We instead use VLA/WAM as an interruptible tool inside a physical orchestrator.

\paragraph{Agentic and Hierarchical Robot Frameworks.} LLM- and VLM-driven program synthesis grounds high-level reasoning in robotic primitives via code generation~\citep{codeaspolicy,progprompt,capx} or closed-loop VLM verification~\citep{comerobot,innermonologue}, with TAMP-augmented variants guiding symbolic task-and-motion planners~\citep{vlmtamp,tiptop}. These remain limited by fixed primitive interfaces and largely overlook real-time monitoring and failure recovery.
A parallel line stacks a VLM planner above a VLA executor~\citep{hamster,hirobot,agenticrobot,goal2skill,manipulateanything, lei2026longhorizon, hivla, generalvla, hvlp_humanoid}, sometimes paired with a critic for failure detection and replanning~\citep{failsafe,criticloop,aha,robofac,novaplan,safevla,racer,replan,replanvlm,lera, reflectvlm, reconvla, fpc_vla, repo_vla}.
Concurrent work ~\cite{lei2026longhorizon} routes a VLM through a family of specialized VLAs. However, these systems still treat the VLM-VLA call as a hardwired pipeline; in contrast, our physical orchestrator treats the VLA/WAM as one interruptible tool among others, enabling real-time monitoring, mid-rollout intervention, and adaptive tool switching to perception or action primitive tools.

\paragraph{Long-Horizon Open-vocabulary Manipulation Benchmarks.}
Manipulation benchmarks span tabletop manipulation~\citep{robosuite,rlbench,libero,maniskill3},
household and kitchen environments~\citep{robocasa,behavior1k}, and
language-conditioned long-horizon tasks~\citep{calvin,vlabench,robocerebra},
with real-to-sim suites measuring policy transfer~\citep{simplerenv,molmospaces2026}.
While RoboCerebra~\citep{robocerebra} and VLABench~\citep{vlabench} stress
multi-step reasoning, they are low-fidelity, evaluate subtasks against a reset scene, and do
not measure memory carried across them; closer to our memory axis,
RMBench~\citep{rmbench} targets memory-dependent manipulation but scopes
it to short single-task contexts. \textsc{RoboVoLo}, built
on RoboLab~\citep{robolab}, instead requires reasoning over spatial state
accumulated by earlier subtasks, isolating persistent memory as a
measurable axis.


\begin{figure}[t]
  \centering
  \includegraphics[width=\linewidth]{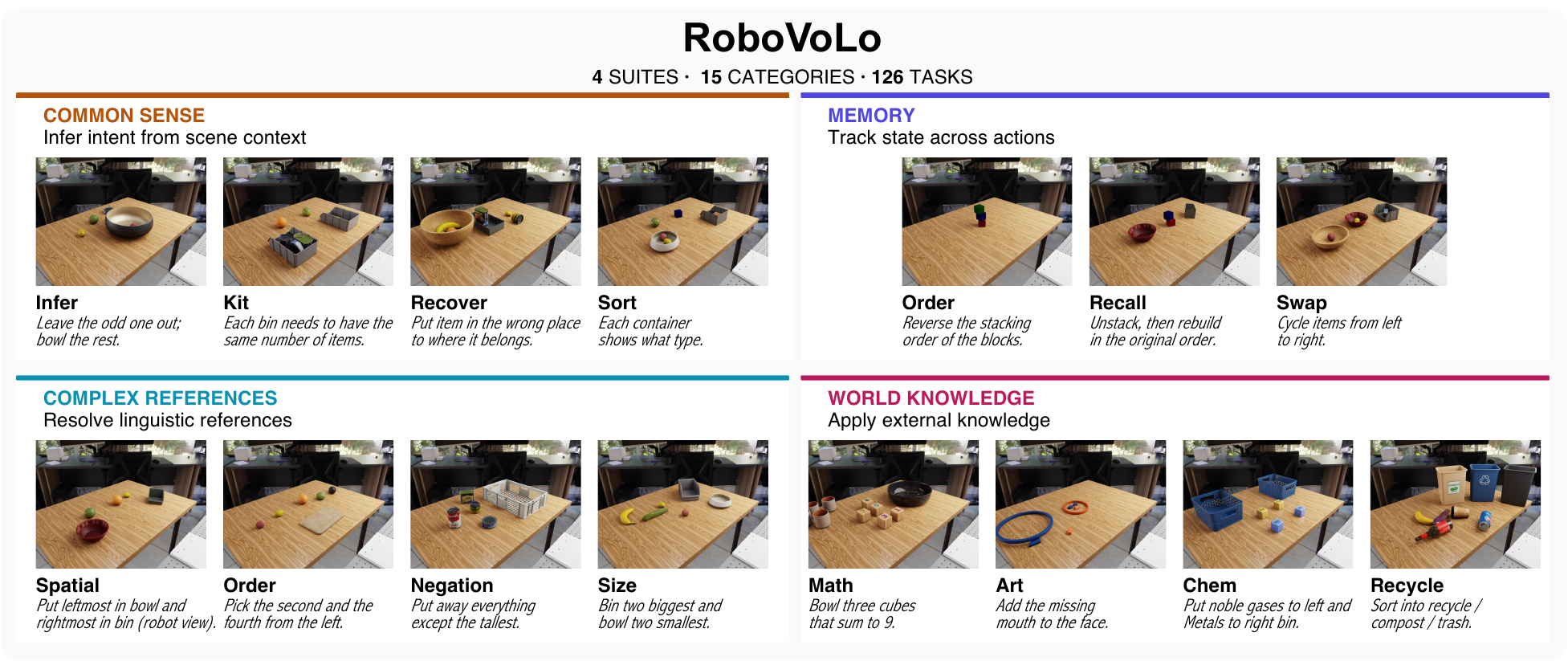}
  \caption{\textbf{\textsc{RoboVoLo} benchmark.} 126 long-horizon manipulation tasks across 15 categories, grouped into four capability suites: \emph{Common Sense} (infer intent from scene context), \emph{Memory} (track state across actions), \emph{Complex References} (resolve spatial, ordinal, size, and negation cues), and \emph{World Knowledge} (apply external knowledge spanning math, art, chemistry, and recycling). Each panel shows one representative task with its instruction.
  }
  \label{fig:bench}
\end{figure}

\section{\textsc{RoboVoLo} Benchmark}
\label{sec:bench}

\paragraph{Tasks and Scenes.} Long-horizon, open-vocabulary manipulation requires a robot to reason and act over many steps. It must ground intent in scene context, track state as the scene changes, resolve fine-grained references, and apply world knowledge to carry out each step while monitoring and recovering from failures. This coupling of reasoning and execution is largely unsolved, and current benchmarks do not isolate it. \textsc{RoboVoLo} fills that gap with 126 tasks that span four reasoning categories, each requiring a chain of grounded manipulation actions. The tasks are built so they cannot be solved by obvious instruction-independent behavior. Figure~\ref{fig:bench} summarizes the taxonomy of four main categories:

\begin{enumerate}[leftmargin=*]
    \item \textbf{Commonsense grounding.}
    Success depends on understanding the functional or contextual role of objects in the current environment, rather than following the instruction verbatim.
    \item \textbf{Memory.}
    These tasks require the policy to maintain information about earlier scene states during execution. Examples include restoring a previous arrangement, undoing a change, swapping objects, or rearranging objects relative to their initial configuration.
    \item \textbf{Complex references.}
    Evaluate fine-grained language understanding. Instructions contain spatial, ordinal, relational, size-based, or negative references that disambiguate objects.
    \item \textbf{World knowledge.}
    These tasks require general knowledge beyond the immediate geometry of the scene, covering domains like recycling, arithmetic, chemistry, and visual art.
\end{enumerate}

\paragraph{Simulator.}
\textsc{RoboVoLo} is built on RoboLab \citep{robolab}, a high-fidelity simulation environment based on NVIDIA Isaac Lab \citep{mittal2025isaaclab}.
To support these tasks, we expand RoboLab's asset library with $501$ new objects: $247$ household assets from NVIDIA's Lightwheel SimReady collection and $254$ task-specific assets, including $118$ chemical periodic-table element cubes, $120$ geometric art objects varying in color, shape, and size, and $16$ wooden math cubes with digits and operators. All assets include collision geometry and realistic physics materials, yielding a diverse collection spanning household, semantic, symbolic, and task-specific categories.


\section{\textsc{VoLoAgent} and Physical Orchestration}
\label{sec:system}

\subsection{Physical Orchestration}
\label{sec:system:loop}

Virtual AI agents assume a world that holds still while the agent thinks, whereas a physical agent must reason while the world keeps moving. This imposes a core requirement: the agent must \emph{monitor} the world for divergence between what it believes it has accomplished and the actual scene, \emph{halt} an in-flight action as quickly as possible if divergence is detected, and \emph{redirect} by choosing a correction: replanning, reissuing the action, or switching tools. Safe halting during reasoning may require an idling policy that for a fixed-base arm is simply stopping, but in general must keep the agent out of harm's way. We refer to this monitor--halt--redirect requirement as \emph{physical orchestration}. 

Prior closed-loop systems address parts of it: VLM-driven frameworks perform situated reasoning and failure recovery~\citep{comerobot}, key-frame agents recover from execution errors~\citep{clier}, and reactive controllers halt a moving base to recover mid-task~\citep{onthemove}, but each targets a subset of these capabilities or a fixed pipeline. With \emph{physical orchestration} we emphasize the need to handle all three together, for an open-vocabulary agent that switches tools mid-rollout, including interrupting asynchronous tools such as a learned visuomotor policy mid-rollout.

\begin{figure}[t]
  \centering
  \includegraphics[width=\linewidth]{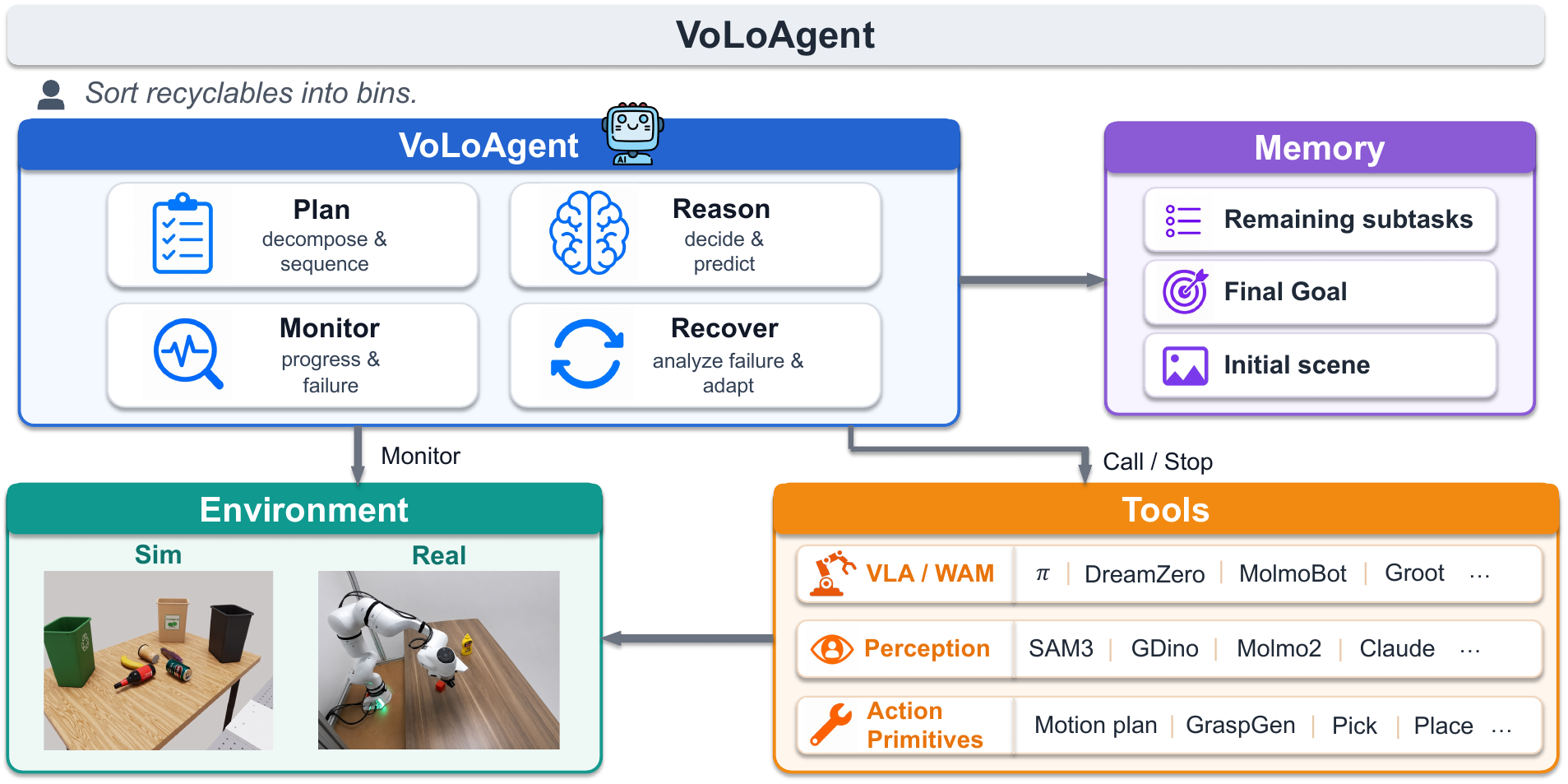}
  \caption{\textbf{\textsc{VoLoAgent} system.} A VLM agent plans, monitors, and orchestrates tools (VLA/WAM rollouts, perception models, grasp/place primitives) through one closed-loop control law. The agent can interrupt a VLA rollout and switch to a different tool when execution drifts.}
  \label{fig:sys}
\end{figure}

\subsection{\textsc{VoLoAgent} System}


\textsc{VoLoAgent} is a physical orchestrator: a single VLM agent that plans subtasks, monitors execution, and continuously routes among tools, deciding whether to continue, switch tools, advance, or recover. Unlike prior hierarchical systems that split control between a VLM planner and a VLA executor, here the VLA is one callable tool alongside perception models and grasp/place primitives, combined complementarily. It realizes the monitor--halt--recover loop through three design choices. \textbf{(P1) Asynchronous tools}: robot motion runs independent of the agent's reasoning, so the agent interleaves monitoring with execution rather than blocking. \textbf{(P2) Fast and slow memory}: a short monitor context (current observation, active subgoal, recent decisions) read as close to the motion timescale as possible (0.2Hz here), and a fuller deliberation context (task memory, scene history, tool catalog) consulted only at planning points, echoing dual-system VLA designs~\citep{pi05}. \textbf{(P3) Safety-aware idling}: holding the robot still when reasoning must continue mid-task.

We instantiate three complementary tool families: \textbf{VLA/WAM} (e.g.\ $\pi_{0.5}$, DreamZero) is a first-class visuomotor tool but can struggle with open-vocabulary grounding. \textbf{Perception tools} (GroundingDINO~\citep{groundingdino}, SAM2~\citep{sam2}, SAM3~\citep{sam3}, Molmo2~\citep{molmo,molmo2}) provide open-vocabulary detection and segmentation. \textbf{Action primitives} such as \texttt{grasp(target)} and \texttt{place(destination)} combine perception, GraspGen~\citep{graspgen}, and IK for geometry-grounded motion but remain rigid under contact-rich interaction. Full API signatures and prompts are in Appendices~\ref{app:system} and~\ref{app:prompts}. The VLM routes among these tools through the following phases:

\paragraph{Initial execution.} Given a user instruction and the initial scene, the agent decomposes the task into atomic subgoals and stores them with the final goal and initial scene in external memory. It then issues the first tool call, typically a VLA rollout for its continuous visuomotor control, and begins monitoring concurrently.

\paragraph{Monitoring \& routing.} At each monitor step the agent reads the latest observation with memory under the monitor context (P2) and selects one of $\{\textsc{continue},\ \textsc{next\_subgoal},\ \textsc{recovery}\}$. There is no fixed split between planner and executor; the same agent decides whether to keep the current tool running, advance, or pause for recovery.

\paragraph{Recovery.} On \textsc{recovery} the active tool is idled (P3) and the agent enters the deliberation context to pick one of: \textsc{continue} if the alarm was a false positive (resume the rollout), \textsc{replan} to re-issue the remaining subgoal decomposition, \textsc{rewrite} to run the VLA with a new subgoal instruction, or \textsc{grasp}\,/\,\textsc{place} to run the respective primitive on a perception-grounded target.

The loop terminates on timeout or task completion. A key emergent property is complementarity: action primitives \emph{inject} perception grounding into the VLA, so even a failed grasp leaves the gripper near the target with a clean view for the VLA to finish the pick (Sec.~\ref{sec:result:sankey},~\ref{sec:result:ablation}).

\begin{table*}[t]
\centering
\caption{Results of various methods on our benchmark (rows: \emph{Common Sense}, \emph{Memory}, \emph{Complex References}, \emph{World Knowledge}), as well as on the \emph{Robolab-Vague} benchmark. Methods (columns) are grouped by families: \emph{Single action model} (no orchestrator), \emph{Code-as-policy + VLM}, \emph{TAMP + VLM}, and \emph{VoLoAgent}. Each task is run for 3 episodes.  All values are success rate (\%, higher is better). \textbf{Bold} = best in row; \underline{underline} = second-best.}
\label{tab:main}
\footnotesize
\setlength{\tabcolsep}{4pt}
\renewcommand{\arraystretch}{1.15}
\resizebox{\linewidth}{!}{%
\begin{tabular}{l l ccccc cc c ccc}
\toprule
\multicolumn{2}{c}{} & \multicolumn{5}{c}{\emph{Single action model}} & \multicolumn{2}{c}{\emph{Code-as-policy}} & \emph{TAMP} & \multicolumn{3}{c}{\textbf{\textsc{VoLoAgent} (Ours)}} \\
\cmidrule(lr){3-7}\cmidrule(lr){8-9}\cmidrule(lr){10-10}\cmidrule(lr){11-13}
Suite & Category
  & $\pi_{0.5}$ & $\pi_0$-FAST & MolmoBot & MolmoAct2 & DreamZero
  & CaPX-s & CaPX-e
  & TiPToP
  & No VLA & Only VLA & \textbf{Full} \\
\midrule
\multirow{5}{*}{\textbf{Common Sense}}
  & Infer    & 0.00 & 9.52 & 14.29 & 0.00 & 19.05 & 9.52 & 14.29 & 4.76 & \underline{19.05} & \textbf{52.38} & \textbf{52.38} \\
  & Kit      & 16.67 & 4.17 & 0.00 & 0.00 & 12.50 & 12.50 & 16.67 & 8.33 & \underline{41.67} & 33.33 & \textbf{50.00} \\
  & Recover  & 4.17 & 0.00 & 12.50 & 12.50 & 20.83 & 37.50 & 29.17 & 0.00 & \textbf{62.50} & \underline{45.83} & \textbf{62.50} \\
  & Sort     & 23.81 & 0.00 & 0.00 & 0.00 & 0.00 & 0.00 & 0.00 & 0.00 & 0.00 & \underline{47.62} & \textbf{52.38} \\
  \rowcolor{gray!15} \cellcolor{white} & \textbf{Overall} & 11.11 & 3.33 & 6.67 & 3.33 & 13.33 & 15.56 & 15.56 & 3.33 & 32.22 & \underline{44.44} & \textbf{54.44} \\
\midrule
\multirow{4}{*}{\textbf{Memory}}
  & Order    &  12.50 & 25.00 & \underline{33.33} & 25.00 & 29.17 & 16.67 & 16.67 & 0.00 & 25.00 & 29.17 & \textbf{54.17}  \\
  & Recall   &  23.33 & 3.33 & 30.00 & 3.33 & 21.43 & 23.33 & 23.33 & 3.33 & 6.67 & \textbf{63.33} & \underline{56.67}  \\
  & Swap     &  3.33 & 0.00 & \underline{6.67} & 3.33 & 0.00 & \underline{6.67} & \underline{6.67} & 0.00 & \textbf{10.00} & \textbf{10.00} & 3.33  \\
  \rowcolor{gray!15} \cellcolor{white} & \textbf{Overall} &  13.10 & 8.33 & 22.62 & 9.52 & 15.85 & 15.48 & 15.48 & 1.19 & 13.10 & \underline{34.52} & \textbf{36.90}  \\
\midrule
\multirow{5}{*}{\textbf{Complex References}}
  & Spatial   &  14.81 & 11.11 & 0.00 & 7.41 & 11.11 & 7.41 & 7.41 & 25.93 & 7.41 & \underline{29.63} & \textbf{40.74}  \\
  & Counting  &  16.67 & 12.50 & 12.50 & 0.00 & 0.00 & 4.17 & 4.17 & 12.50 & 4.17 & \underline{45.83} & \textbf{54.17}  \\
  & Negation  &  16.67 & 0.00 & 0.00 & 0.00 & 0.00 & 0.00 & 0.00 & 20.83 & 25.00 & \underline{45.83} & \textbf{54.17}  \\
  & Size+Sort &  19.05 & 4.76 & 9.52 & 0.00 & 4.76 & 19.05 & 19.05 & 23.81 & 0.00 & \underline{42.86} & \textbf{57.14}  \\
  \rowcolor{gray!15} \cellcolor{white} & \textbf{Overall} &  16.67 & 7.29 & 5.21 & 2.08 & 4.17 & 7.29 & 7.29 & 20.83 & 9.38 & \underline{40.62} & \textbf{51.04}  \\
\midrule
\multirow{5}{*}{\textbf{World Knowledge}}
  & Art     &  0.00 & 0.00 & 0.00 & 0.00 & 0.00 & 0.00 & 0.00 & \textbf{16.67} & \textbf{16.67} & 4.17 & \underline{8.33}  \\
  & Chem    &  8.33 & 0.00 & 12.50 & 4.17 & 12.50 & 4.17 & 4.17 & \underline{50.00} & 29.17 & 41.67 & \textbf{54.17}  \\
  & Math    &  4.17 & 0.00 & 0.00 & 0.00 & 0.00 & 0.00 & 0.00 & 4.17 & \textbf{20.83} & 0.00 & \underline{12.50}  \\
  & Recycle &  \underline{25.00} & 0.00 & 4.17 & 0.00 & 0.00 & 4.17 & 4.17 & 20.83 & 0.00 & \textbf{37.50} & \underline{25.00}  \\
  \rowcolor{gray!15} \cellcolor{white} & \textbf{Overall} &  9.38 & 0.00 & 4.17 & 1.04 & 3.12 & 2.08 & 2.08 & \underline{22.92} & 16.67 & 20.83 & \textbf{25.00}  \\
\midrule
\multirow{4}{*}{\textbf{Robolab-Vague}}
  & Easy     & 19.79 & 10.94 & 13.76 & 6.25 & 19.79 & 16.67 & 15.10 & 29.69 & 19.79 & \textbf{35.94} & \underline{34.90} \\
  & Med      & 17.54 & 11.40 & 11.40 & 6.14 & 18.80 & 14.04 & 9.65 & 7.02 & 16.67 & \underline{26.32} & \textbf{30.70} \\
  & Hard     & 5.56 & 3.70 & 3.77 & 0.00 & 13.73 & 7.41 & 1.85 & 5.56 & 12.96 & \underline{16.67} & \textbf{24.07} \\
  \rowcolor{gray!15} \cellcolor{white} & \textbf{Overall} & 16.94 & 10.00 & 11.52 & 5.28 & 18.61 & 14.44 & 11.39 & 18.89 & 17.78 & \underline{30.00} & \textbf{31.94} \\
\bottomrule
\multicolumn{13}{l}{\footnotesize Method abbreviations: CaPX-s = CaP-X single, CAPX-e = CAP-X ensemble.}\\
\end{tabular}%
}
\end{table*}

\section{Experimental Results}
\label{sec:result}

\subsection{Setup}
\label{sec:result:setup}


\paragraph{Simulation benchmarks.}
We evaluate on four \textsc{RoboVoLo} suites covering 126 tasks and on the existing RoboLab~\cite{robolab} benchmark (120 tasks) with vague-choice instructions. All policy models use the DROID setup~\citep{droid}: a 7-DoF Franka Research 3 arm with a Robotiq 2F-85 gripper, external ZED 2i and wrist ZED mini cameras, and a 7-DoF joint-position plus binary-gripper action space. Camera poses and lighting match the real DROID configuration. Each task is evaluated over three fixed-seed trials to ensure identical initial states across systems. We also explored MuJoCo-based benchmarks, including LIBERO, RoboCerebra, and VLABench, but found them unsuitable due to limited generalist-policy support and insufficient realism; see Appendix~\ref{app:other-sims}.

\paragraph{\textsc{VoLoAgent}.}
 \textit{\textsc{VoLoAgent} (Full)} uses Claude Opus 4.6~\citep{claudeopus47} as the decision-making VLM with the following tools: $\pi_{0.5}$~\citep{pi05} as the VLA, SAM3~\citep{sam3} and Molmo2~\citep{molmo2} as perception tools, and GraspGen~\citep{graspgen} with multi-start IK plus depth-projected point placement for pick and place execution. The VLA and primitives run at $15$Hz, while the VLM monitors at $0.2$Hz from a front camera. We found this monitoring frequency reasonable for the pace of VLA motion, but increasing or adaptively varying it is an important future design goal.
 We compare two main ablations: \textit{\textsc{VoLoAgent} (No VLA)}, which only uses perception tools and \textsc{grasp}/\textsc{place} action primitives, and \textit{\textsc{VoLoAgent} (Only VLA)}, which disables all other tools and only relies on verbal steering of the VLA. Complete component ablations are in Sec.~\ref{sec:result:ablation}.

\paragraph{Baselines.}
We compare against three baseline families: (i) standalone action-model policies ($\pi_{0.5}$~\citep{pi05}, $\pi_0$-FAST~\citep{pi0fast}, MolmoBot~\citep{molmobot}, MolmoAct2~\citep{molmoact2}, DreamZero~\citep{dreamzero}), (ii) code-as-policy + VLM (CaP-X~\citep{capx}, single and ensemble), and (iii) TAMP + VLM (TiPToP~\citep{tiptop}).

\subsection{Main Results}
\label{sec:result:main}

Table~\ref{tab:main} shows that \textsc{VoLoAgent} achieves the best long-horizon open-vocabulary manipulation performance, outperforming single-model, code-as-policy, and TAMP baselines on every suite. The full system is significantly better than all methods ($p<0.05$), except the Only VLA ablation ($p=0.0598$), under a paired randomization test that asks whether one method consistently outperforms another across tasks~\citep{edgington2007randomization} (see Appendix~\ref{app:stat-detail}). 
Against the strongest baseline in each suite, \textsc{VoLoAgent} (Full) gains 
+38.9\% on Common Sense,
+30.2\% on Complex References,
+14.3\% on Memory, and
+13.1\% on Robolab-Vague; the exception is World Knowledge (+2.1\%), where the TAMP baseline's symbolic planning is competitive.
The gains come primarily from the planning, monitoring and recovery inherent in a physical orchestrator design, supplemented by the availability of complementary tools whose individual strengths cover others' blind spots.
The full system also exceeds its own strongest ablation on every suite by between +1.9\% and +10.4\%. 

\begin{figure}[h]
  \centering
  \includegraphics[width=\linewidth]{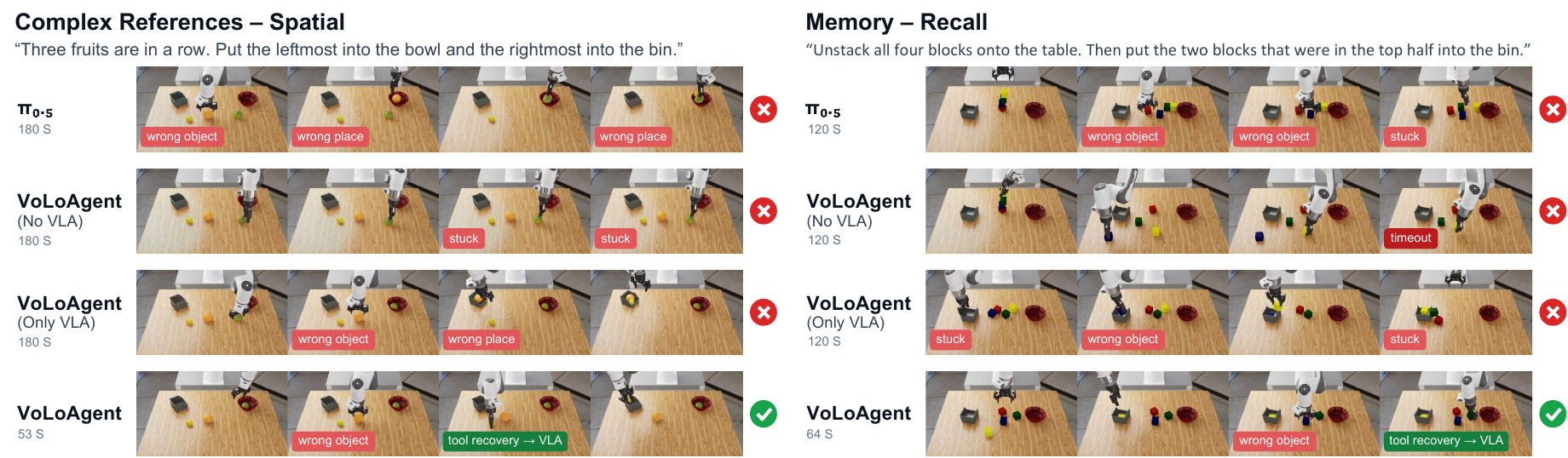}
  \caption{\textbf{Process comparison on two open-vocabulary long-horizon tasks}, one row per system. Red tags mark failure events and green tags mark grasp-tool recovery events. The behaviors shown are described in Sec.~\ref{sec:result:main}.}
  \label{fig:process}
\end{figure}

Figure~\ref{fig:process} illustrates these gains on two representative tasks. $\pi_{0.5}$ relies on visual priors and ignores open-vocabulary constraints, placing all objects into the same bowl. \textsc{VoLoAgent} (No VLA) grounds the instruction and plans subtasks, but its action primitives struggle with contact-rich picks and exhaust the step budget. \textsc{VoLoAgent} (Only VLA) can steer the VLA through prompts, but remains limited by the VLA's perception errors, such as grasping an orange instead of a lemon. The full system combines their strengths: when the VLA selects the wrong object, the grasp tool repositions the gripper on the correct target, and the VLA completes the contact-rich manipulation.


\subsection{Failure Mode Analysis}
\label{sec:result:sankey}

\paragraph{Metrics Definition.} We analyze failures along two axes. \textbf{World failures} measure state-level execution errors: wrong-object pick (\emph{WOP}), wrong-target placement (\emph{WTP}), and lack of end-effector progress for over 10s (\emph{Stuck}), each paired with a \texttt{recovery} event when resolved. \textbf{VLM failures} measure reasoning and action errors: incorrect planning, false or missed completion monitor, missed failure detection, and wrong tool calling. Metrics mainly use ground-truth simulation states and human-labeled task features; full definitions are in Appendix~\ref{app:metrics-defs}.

\begin{figure}[h]
\centering
\includegraphics[width=.85\linewidth]{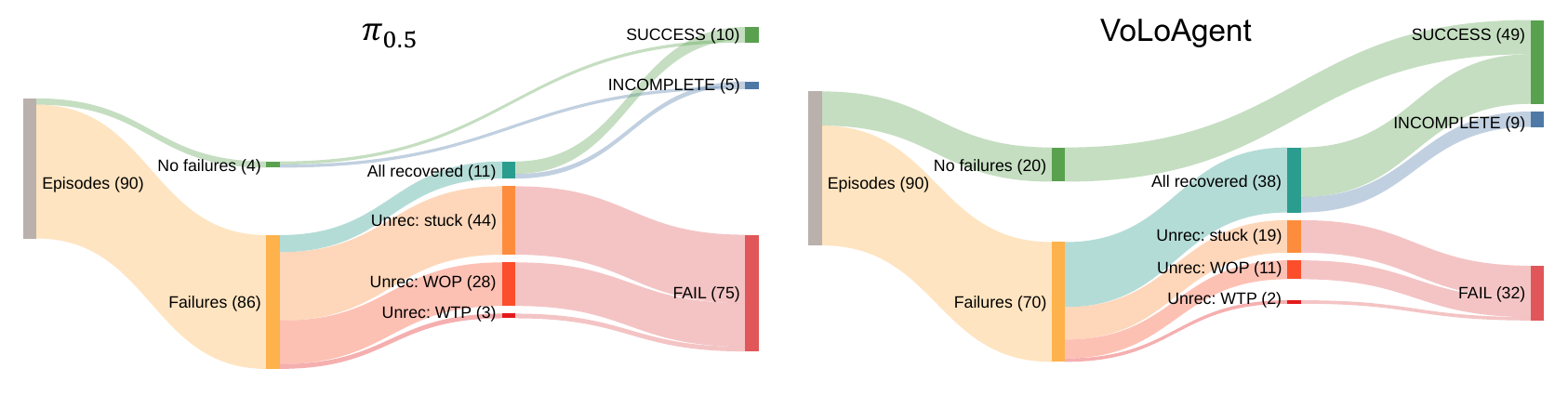}\
\caption{\textbf{World failure analysis} tracing episodes through failures, recovery, and outcomes for $\pi_{0.5}$ (left) and \textsc{VoLoAgent} (right). Major failure subtypes: \emph{stuck}, \emph{WOP}=wrong object picked, \emph{WTP}=wrong target place. Band thickness is proportional to the number of episodes. 
}
\label{fig:sankey}
\end{figure}

\paragraph{World Failures.} Figure~\ref{fig:sankey} traces $\pi_{0.5}$ and \textsc{VoLoAgent} through the failure-recovery pipeline for world failures. \textsc{VoLoAgent} has $5\times$ more failure-free episodes than $\pi_{0.5}$ ($20$ vs.\ $4$). Among episodes that do hit a failure, \textsc{VoLoAgent} recovers from $54\%$ ($38/70$) vs.\ only $13\%$ ($11/86$) for $\pi_{0.5}$, showing that \textsc{VoLoAgent} not only enhances direct success but also greatly improves failure recovery (see  Appendix~\ref{app:sankey} for \textsc{VoLoAgent} ablations).

\begin{figure}[h]
  \centering
  \includegraphics[width=\linewidth]{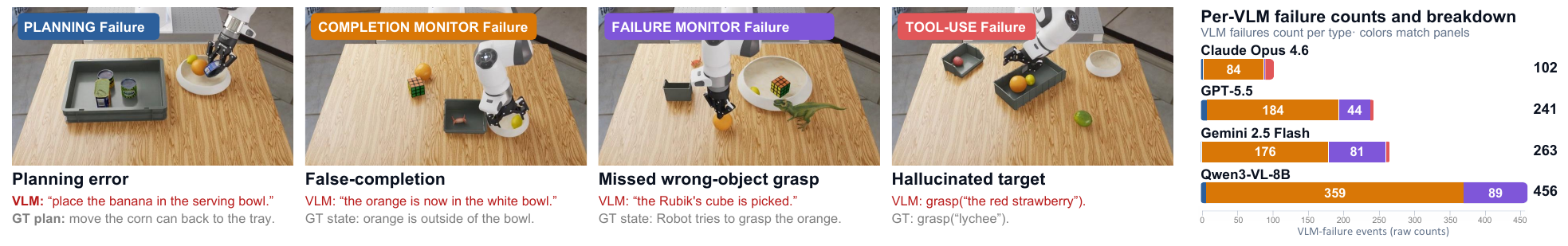}
  \caption{\textbf{VLM failure audit.} \emph{Left:} one example per failure type (Planning, Completion-monitor, Failure-monitor, Tool-use). \emph{Right:} per-VLM error counts across $n{=}90$ episodes; segment colors match the example tag colors. Qwen3-VL-8B reaches $23\%$ of the ceiling error counts, Claude Opus 4.6 only $5\%$. Error definitions in Appendix~\ref{app:metrics-defs}.
  } 
  \label{fig:vlm_failure}
\end{figure}

\paragraph{VLM Failures.} Figure~\ref{fig:vlm_failure} shows one qualitative example per VLM failure class (left) and per-VLM event counts across four frontier VLMs (right). \emph{Completion-monitor} errors dominate every backend, accounting for $>$$67\%$ of total events and increasing $4.3\times$ across VLM capability. \emph{Failure-monitor} errors are another major class for every VLM except Claude Opus 4.6: GPT-5.5 has $44$, Gemini 2.5 Flash $81$, and Qwen3-VL-8B $89$. \emph{Planning} errors are rare for every backend ($\leq 9$ events per $90$ episodes) as are \emph{tool-use} mismatches ($\leq 12$ events). Improving completion and failure monitoring are next steps to strengthening the physical orchestrator design.

\subsection{Component Ablations}
\label{sec:result:ablation}

We conduct comprehensive ablation studies, varying one component at a time while holding the rest at our default and study four axes: \textbf{System} comparing single VLA and VoLoAgent variants, 
\textbf{Perception} varying the 
\begin{wraptable}{r}{0.36\linewidth}
\centering
\captionsetup{font=footnotesize}
\caption{Component ablation, cross-suite \emph{Overall} RoboVoLo success rate (\%). Full breakdown in Table~\ref{tab:ablation-full}.}
\label{tab:ablation}
\footnotesize
\setlength{\tabcolsep}{4pt}
\renewcommand{\arraystretch}{1.05}
\resizebox{\linewidth}{!}{%
\begin{tabular}{l l c}
\toprule
Axis & Ablation & \textbf{Success rate} \\
\midrule
\multirow{3}{*}{System}
& $\pi_{0.5}$ (Pure VLA)      & 12.57 \\
& \textsc{VoLoAgent} (No VLA)          & 17.76 \\
& \textsc{VoLoAgent} (Only VLA)        & 34.97 \\
\midrule
\multirow{3}{*}{Perception}
& GDino+SAM2 / Molmo2         & 38.52 \\
& SAM3 / VLM-point            & 36.07 \\
& Exterior camera             & 36.94 \\
\midrule
\multirow{3}{*}{VLM model}
& GPT-5.5                     & 35.52 \\
& Gemini-2.5-Flash            & 31.97 \\
& Qwen3-VL-8B                 & 19.95 \\
\midrule
\multirow{3}{*}{VLA model}
& $\pi_{0}$-FAST              & 26.23 \\
& MolmoBot-DROID              & 24.86 \\
& DreamZero-DROID             & 21.86 \\
\midrule
\rowcolor{gray!15} \multicolumn{2}{l}{\textbf{\textsc{VoLoAgent}}} & \textbf{41.80} \\
\bottomrule
\end{tabular}%
}
\end{wraptable}
perception tools and camera view, and the choice of \textbf{VLM} and \textbf{VLA} model.
Table~\ref{tab:ablation} reports cross-suite \emph{Overall} success rates.
\textbf{System.} The full system reaches $41.80$, while Single-VLA, \emph{\textsc{VoLoAgent} (No VLA)}, and \emph{\textsc{VoLoAgent} (Only VLA)} score $12.57$\%, $17.76$\%, and $34.97$\% respectively.
\textbf{Perception.} The system is robust to the choice of perception tool, with all variants achieving substantial gains. It also remains strong with the DROID exterior camera view to the orchestrator, though performance drops slightly because objects are sometimes occluded in the exterior and wrist views. \textbf{VLM.} Frontier VLMs as orchestrator yield $+19$\% to $+29$\% over the VLA-only baseline; with weaker VLMs the gain becomes marginal, the open-weights Qwen3-VL-8B model drops to $+7$\%, aligning with its $4\times$ higher VLM-failure count in Fig.~\ref{fig:vlm_failure}. \textbf{VLA.} The orchestrator multiplies every VLA backbone by $2$--$6\times$ overall, and gains hold across every base policy.
We compared methods using task-paired success-rate differences, using aggregate success over all trials and two-sided exact sign-flip permutation tests over per-task success fractions. The only non-significant comparisons ($p<0.05$) to the full system were two perceptual ablations: GDino+SAM2 / Molmo2 and Exterior camera.

\begin{figure}[t]
  \centering
  \includegraphics[width=\linewidth]{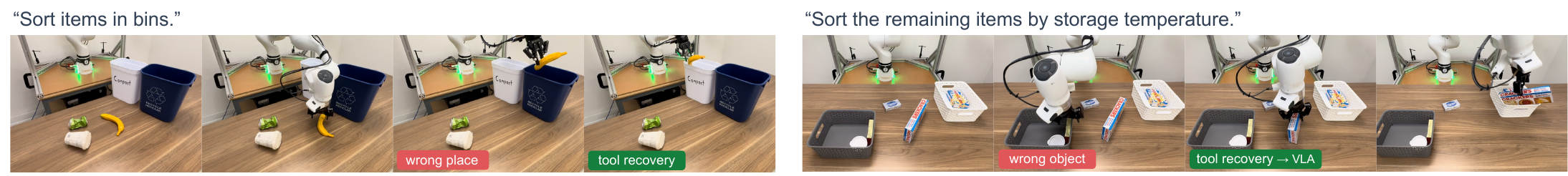}
  \caption{\textbf{Real robot examples.} \textsc{VoLoAgent} monitors and recovers from failures such as wrong place destination, wrong object pick in the real world as well.
  } 
  \label{fig:real_robot}
\end{figure}

\subsection{Real Robot Validation}
\label{sec:result:real}

\begin{wraptable}{r}{0.3\linewidth}
  \centering
  \captionsetup{font=footnotesize}
  \caption{Real-robot success rate (\%) with 95\% Wilson confidence  across $14$ tasks $\times\,3$ trials.}
  \label{tab:real-robot}
  \footnotesize
  \setlength{\tabcolsep}{3pt}
  \renewcommand{\arraystretch}{1.05}
  \resizebox{\linewidth}{!}{%
  \begin{tabular}{l c c}
  \toprule
  System & \textbf{Overall} & \textbf{95\% CI} \\
  \midrule
  $\pi_{0.5}$                  & 14.3\% & [6.7, 27.8] \\
  \textsc{VoLoAgent} (No VLA)           & 45.2\% & [31.2, 60.1] \\
  \textsc{VoLoAgent} (Only VLA)         & 40.5\% & [27.0, 55.5] \\
  \textbf{\textsc{VoLoAgent} (full)}    & 42.9\% & [29.1, 57.8] \\
  \bottomrule
  \end{tabular}%
  }
  \end{wraptable}

To evaluate whether \textsc{VoLoAgent} can operate beyond simulation, we deploy it on a real Franka FR3 with physical objects across a representative sample of $14$ \textsc{RoboVoLo} tasks, running $3$ matched-initial-state trials per task for $\pi_{0.5}$, \textsc{VoLoAgent} variants, and full \textsc{VoLoAgent}, for a total of 168 rollouts across variants. Full \textsc{VoLoAgent} achieves $\mathbf{42.9\%}$ success versus $14.3\%$ for $\pi_{0.5}$, a $3\times$ improvement that supports the physical applicability of our agent loop design (Table~\ref{tab:real-robot}). Figure~\ref{fig:real_robot} shows representative real-world recoveries from wrong-object picks and wrong-place drops. The intermediate variants achieve similar real-robot success ($45.2\%$ and $40.5\%$) with highly overlapping confidence intervals. Qualitatively, the grasp tool appears to work better in the real world than sim due to contact dynamics differences. Reaching statistical power to compare the ablations requires a larger real-robot study on substantially more tasks and trials per system. See Appendix~\ref{app:real-robot} for more details including full list of tasks.



\section{Conclusion and Limitations}
\label{sec:conclusion}

We introduced \textsc{VoLoAgent}, a physical orchestrator that unifies VLA/WAM rollouts, perception models, and grasp/place primitives in a VLM-managed closed loop, and \textsc{RoboVoLo}, a $126$-task benchmark for open-vocabulary long-horizon manipulation.
\textsc{VoLoAgent} outperforms existing baselines, with ablations showing that orchestration drives the gains. 

\paragraph{Limitations.}
Our failure analysis (Sec.~\ref{sec:result:sankey}) highlights completion monitoring accuracy as a key direction for improvement. The per-call latency ($\sim$1--5\,s for cloud VLMs) of the orchestrating VLM bounds reaction time and may miss fast failures, calling for fast local monitors.
\textsc{VoLoAgent} was demonstrated on a single-arm manipulator with a parallel-jaw gripper. Extending to bimanual,
dexterous-hand, or mobile embodiments is supported by the framework, but requires retraining or swapping the VLA.
Safe idling currently reduces to halting the arm, which does not generalize to embodiments that must act to stay safe (e.g., a balancing humanoid).

\setcitestyle{numbers}
\bibliographystyle{plainnat}
\bibliography{main}
\clearpage
\appendix

\begin{center}
{\LARGE \bfseries Appendix}
\end{center}

\input{corl_2026_template_submission/appendix_body}

\end{document}

%% file: packages.tex
\usepackage[authoryear,sort&compress,round]{natbib}

\usepackage[utf8]{inputenc} 
\usepackage[T1]{fontenc}    

\usepackage{parskip}        
\usepackage{url}            
\usepackage{hyperref}       
\usepackage{booktabs}       
\usepackage{amsfonts}       
\usepackage{nicefrac}       
\usepackage{microtype}      
\usepackage{xcolor}         
\usepackage[dvipsnames]{xcolor} 
\usepackage{graphicx}
\usepackage{animate}        
\usepackage{subcaption}
\usepackage{tabularx}
\usepackage{makecell}
\usepackage{adjustbox}
\usepackage{setspace}
\newcolumntype{M}[1]{>{\centering\arraybackslash}m{#1}}
\usepackage{float}
\usepackage{tikz}
\usetikzlibrary{positioning,shapes,arrows}
\usepackage{amsmath,amsfonts,bm, bbm,leftindex}
\usepackage{multirow}
\usepackage{comment}
\usepackage{gensymb}
\usepackage{lipsum}
\usetikzlibrary{arrows.meta, positioning, fit}
\usepackage[para]{threeparttable}
\usepackage{tikz}
\usetikzlibrary{tikzmark}



%% file: common.tex
\def\eqref#1{equation~\ref{#1}}

\def\1{\bm{1}}

\DeclareMathAlphabet{\mathsfit}{\encodingdefault}{\sfdefault}{m}{sl}
\SetMathAlphabet{\mathsfit}{bold}{\encodingdefault}{\sfdefault}{bx}{n}

\makeatletter
\let\save@mathaccent\mathaccent
\newcommand*\if@single[3]{%
  \setbox0\hbox{${\mathaccent"0362{#1}}^H$}%
  \setbox2\hbox{${\mathaccent"0362{\kern0pt#1}}^H$}%
  \ifdim\ht0=\ht2 #3\else #2\fi
  }
\newcommand*\rel@kern[1]{\kern#1\dimexpr\macc@kerna}
\newcommand*\widebar[1]{\@ifnextchar^{{\wide@bar{#1}{0}}}{\wide@bar{#1}{1}}}
\newcommand*\wide@bar[2]{\if@single{#1}{\wide@bar@{#1}{#2}{1}}{\wide@bar@{#1}{#2}{2}}}
\newcommand*\wide@bar@[3]{%
  \begingroup
  \def\mathaccent##1##2{%
    \let\mathaccent\save@mathaccent
    \if#32 \let\macc@nucleus\first@char \fi
    \setbox\z@\hbox{$\macc@style{\macc@nucleus}_{}$}%
    \setbox\tw@\hbox{$\macc@style{\macc@nucleus}{}_{}$}%
    \dimen@\wd\tw@
    \advance\dimen@-\wd\z@
    \divide\dimen@ 3
    \@tempdima\wd\tw@
    \advance\@tempdima-\scriptspace
    \divide\@tempdima 10
    \advance\dimen@-\@tempdima
    \ifdim\dimen@>\z@ \dimen@0pt\fi
    \rel@kern{0.6}\kern-\dimen@
    \if#31
      \overline{\rel@kern{-0.6}\kern\dimen@\macc@nucleus\rel@kern{0.4}\kern\dimen@}%
      \advance\dimen@0.4\dimexpr\macc@kerna
      \let\final@kern#2%
      \ifdim\dimen@<\z@ \let\final@kern1\fi
      \if\final@kern1 \kern-\dimen@\fi
    \else
      \overline{\rel@kern{-0.6}\kern\dimen@#1}%
    \fi
  }%
  \macc@depth\@ne
  \let\math@bgroup\@empty \let\math@egroup\macc@set@skewchar
  \mathsurround\z@ \frozen@everymath{\mathgroup\macc@group\relax}%
  \macc@set@skewchar\relax
  \let\mathaccentV\macc@nested@a
  \if#31
    \macc@nested@a\relax111{#1}%
  \else
    \def\gobble@till@marker##1\endmarker{}%
    \futurelet\first@char\gobble@till@marker#1\endmarker
    \ifcat\noexpand\first@char A\else
      \def\first@char{}%
    \fi
    \macc@nested@a\relax111{\first@char}%
  \fi
  \endgroup
}
\makeatother


%% file: corl_2026_template_submission/appendix_body.tex
We provide additional details and extended results in the
supplementary materials:
\begin{itemize}\setlength{\itemsep}{1pt}
  \item \textbf{Appendix~\ref{app:future}:} Future directions.
  \item \textbf{Appendix~\ref{app:system}:} Agentic system and tool implementation details.
  \item \textbf{Appendix~\ref{app:prompts}:} VLM prompt templates.
  \item \textbf{Appendix~\ref{app:bench-tasks}:} \textsc{RoboVoLo}
    benchmark details and visualizations.
  \item \textbf{Appendix~\ref{app:sim-setup}:} Simulation setup and compute details.
  \item \textbf{Appendix~\ref{app:stat-detail}:} Statistic test details.
  \item \textbf{Appendix~\ref{app:real-robot}:} Real robot setup
    and per task breakdowns.
  \item \textbf{Appendix~\ref{app:other-sims}:} Eploration on other simulation environments.
  \item \textbf{Appendix~\ref{app:ablation-full}:} Per suite
    breakdown of the component ablation.
  \item \textbf{Appendix~\ref{app:sankey}:} Additional Sankey
    diagrams.
  \item \textbf{Appendix~\ref{app:metrics-defs}:} Failure mode
    definitions.
\end{itemize}

\vspace{4pt}

\section{Future Directions}
\label{app:future}

\textsc{VoLoAgent} is currently restricted to a single-arm parallel-jaw
gripper on a tabletop. Extending to bimanual coordination,
dexterous-hand manipulation, or mobile-base / humanoid embodiments
is a natural next step; the agent loop and tool API are
embodiment-agnostic, but the action-primitive tools (grasp, place)
would need re-implementations that respect the new kinematics and
contact model. Moreover, distilling the completion- and failure-monitor calls into a smaller and stronger local checker is another potential direction for the future, because completion-monitor errors are the dominant failure category on every VLM we evaluated (including the open-weights Qwen3-VL-8B), and a specialized checker can plausibly cut latency by an order of magnitude.


\section{Agentic System — Tools and API}
\label{app:system}

This appendix expands the agent loop and tool catalog of
the main paper. Appendix~\ref{app:system:proxy} describes the
proxy architecture and how it isolates the orchestrator from the
eval client and VLA so that new simulators or policy backends can be
plugged in without touching the agent loop. Appendix~\ref{app:system:tools}
specifies the tool catalog and the action-primitive pipelines.

\subsection{Proxy Architecture and Extensibility}
\label{app:system:proxy}

\paragraph{Process layout.}
\textsc{VoLoAgent} is implemented as a stand-alone proxy that sits
between an existing eval client and an existing VLA/WAM policy server.
The four components run as separate processes, typically on separate
GPUs and in separate Conda / venv environments because of dependency
conflicts between the perception stack, the policy stack, and the
simulator:

\begin{itemize}\setlength{\itemsep}{1pt}
\item \emph{Eval client} (e.g.\ robolab IsaacLab driver, LIBERO /
VLABench runner): owns the simulator, sends observations, applies
actions.
\item \emph{Orchestrator} (this work): runs the VLM agent loop,
dispatches action primitives \texttt{grasp}\,/\,\texttt{place}, and
proxies the VLA channel.
\item \emph{VLA / WAM policy server} (e.g.\ openpi for $\pi_{0.5}$,
DreamZero, MolmoBot): exposes a chunked-action endpoint over its
native wire protocol.
\item \emph{Tool server}: hosts GraspGen + SAM3 + Molmo2 in a
separate Conda env; reached by the orchestrator over an
HTTP RPC.
\end{itemize}
None of these processes share Python imports; the only coupling is
on the wire. This isolation lets the GraspGen stack and a 14B WAM
coexist on the same node without dependency conflicts, and lets us
swap the eval client or VLA without rebuilding the orchestrator.

\paragraph{Pluggable transports: Frontend and Backend.}
The orchestrator defines two abstract interfaces: \texttt{Frontend}
accepts eval-client connections in a specific wire protocol and
exposes a \texttt{FrontendSession} that yields canonical observations
and consumes canonical actions; \texttt{Backend} opens a connection
to the VLA server in some (possibly different) protocol and
translates canonical observations to the VLA's native schema and back.
Strategies in between always see a single canonical schema (the
openpi observation/action layout, msgpack-numpy encoded), so adding a
new VLA family requires only a new $\sim$200-line protocol module ---
no changes to the agent loop or the tool catalog.

The repository ships four protocol modules:
\begin{itemize}\setlength{\itemsep}{1pt}
\item \emph{OpenPI WebSocket} ($\pi_{0}$, $\pi_{0.5}$, paligemma) ---
the native canonical case, essentially a pass-through codec.
\item \emph{GR00T ZMQ} --- translates between the canonical schema
and GR00T's native dict structure with a 180
$\times$ 320 wide-aspect frame.
\item \emph{OpenVLA REST} --- adapts a JSON-numpy single-step REST
endpoint, padding 7-D joint deltas to the canonical 8-D chunk and
replicating the single action across the chunk since OpenVLA does
not chunk natively.
\item \emph{File-IPC} --- routes observations and actions through
atomically-renamed msgpack files on shared storage. We use this on
osmo to drive a 14B DreamZero server hosted on a separate H100 pool
that cannot open a TCP socket to the L40 pool running Isaac Sim.
\end{itemize}

\paragraph{Simulator extensibility via \texttt{--env} presets.}
The second axis of extensibility targets the simulator side. The
\texttt{--env} flag selects a small set of defaults that adapt the
orchestrator to a specific simulator without code changes: image-key
conventions (\texttt{exterior\_image\_1\_left} vs.\
\texttt{observation/image}), action-chunk length (8-step joint-position
chunks for IsaacLab/PhysX vs.\ 5-step OSC\_POSE replans for LIBERO),
and the cadence of monitor calls (\texttt{check\_interval} halved on
LIBERO so monitor frequency in real time stays comparable). All
preset values are overridable per flag, so a new simulator typically
requires only a new preset entry plus the right image keys ---
everything else (agent loop, prompts, tool catalog) is untouched. We
have exercised this path with robolab / IsaacLab, LIBERO, RoboCasa
and VLABench (see Appendix~\ref{app:other-sims}).

\paragraph{Mode and recovery extensibility.}
Within a fixed (frontend, backend, env) triple, the orchestration
behaviour is controlled by three orthogonal CLI flags. \texttt{--mode}
selects the strategy (\texttt{passthrough}, \texttt{subgoal},
\texttt{tool\_chain}, \texttt{next\_goal}, \ldots);
\texttt{--failure-monitor} selects the monitor backend (\texttt{vlm},
\texttt{gt}, \texttt{signal\_primary}, \ldots); and
\texttt{--recovery-mode} selects the action vocabulary that the
monitor exposes (\texttt{replan}, \texttt{replan\_grasp},
\texttt{replan\_tools}, \ldots; see Appendix~\ref{app:prompts}).
The three \textsc{VoLoAgent} variants reported in the main paper
are reached by toggling these flags only; no orchestrator code path
is variant-specific, which keeps the system easy to extend with new
strategies, monitors, or recovery vocabularies.

\subsection{Tool Catalog and API}
\label{app:system:tools}

\paragraph{Tool catalog.}
\begin{itemize}\setlength{\itemsep}{1pt}
\item \texttt{vla(prompt, obs)} $\rightarrow$ \texttt{action\_chunk}
[$N$\,steps]: forwards the current observation and the instruction
prompt to the policy server. Runs asynchronously (P1) at $15$\,Hz; the
agent can \emph{halt} an in-flight chunk when the monitor returns
\textsc{recovery}.
\item \texttt{grasp(target)} $\rightarrow$ \texttt{action\_chunk}
[$M$\,steps]: classic action primitive that grasps the named object.
\texttt{target} is a natural-language phrase produced by the VLM and
used by the perception stack to detect the target object.
\item \texttt{place(destination)} $\rightarrow$ \texttt{action\_chunk}
[$M$\,steps]: classic action primitive that places the currently-held
object at the named destination. \texttt{destination} is a
natural-language phrase produced by the VLM and used by the perception
stack to localize the placement region.
\end{itemize}
Perception models (GroundingDINO~\citep{groundingdino},
SAM2~\citep{sam2}, SAM3~\citep{sam3}, Molmo2~\citep{molmo2}) are
invoked \emph{inside} \texttt{grasp}/\texttt{place} to ground the
natural-language target before motion planning (see ``Action-primitive
pipelines'' below). Hiding intermediate perception outputs from the VLM
keeps the agent's reasoning loop short and fast, in contrast to
SpaceTools-style designs~\citep{spacetools} that re-prompt the VLM on
every intermediate detection.

\paragraph{Action-primitive pipelines.}
\texttt{grasp(target)} runs: open-vocabulary detection/segmentation
(e.g., GroundingDINO~\citep{groundingdino} or SAM3~\citep{sam3})
$\rightarrow$ depth-aware point-cloud crop $\rightarrow$
GraspGen~\citep{graspgen} 6-DoF pose $\rightarrow$ multi-start Franka
IK $\rightarrow$ Cartesian trajectory streamed back to the eval client
in the same chunk format as \texttt{vla}.
\texttt{place(destination)} runs: 2-D destination point
(Molmo2-point~\citep{molmo2} or a direct VLM point) $\rightarrow$
depth projection to a 3-D world target $\rightarrow$ IK with the
current end-effector orientation preferred (top-down fallback)
$\rightarrow$ release at a fixed clearance above the target.

\paragraph{IK algorithm details.}
Both primitives use the same Cartesian-to-joint solver: damped
least-squares (DLS) IK with null-space joint centering, wrapped in a
multi-start ladder for robustness on cluttered tabletops. Given a
target end-effector pose $T^{\star} \in SE(3)$ and a seed configuration
$q_0 \in \mathbb{R}^7$, each iteration computes the analytic Jacobian
$J(q)$, the 6-D pose error $e \in \mathbb{R}^6$ (position + axis-angle
rotation), and the DLS step
$\Delta q = J^{\!\top}(JJ^{\!\top} + \lambda^2 I)^{-1} e
+ (I - J^{+}J)\,\alpha\,(q_{\text{mid}} - q)$,
with damping $\lambda{=}5\!\times\!10^{-3}$ and null-space gain
$\alpha{=}0.5$ pulling toward the joint mid-range $q_{\text{mid}}$.
Steps are clamped to $\|\Delta q\|_2 \leq 0.2$\,rad and configurations
are projected onto the Franka joint limits at every iteration.
Convergence is declared when position error $<$\,$5\!\times\!10^{-4}$\,m
and rotation error $<$\,$5\!\times\!10^{-3}$\,rad, up to $500$
iterations. The multi-start wrapper retries seeds in a fixed ladder ---
caller seed, $\pm\pi$ wrist flip on joint 7, a canonical
well-conditioned reset, and up to three random draws from
$q_{\text{mid}} \pm 1.5$\,rad --- and accepts the first solution that
satisfies the internal tolerances together with FK-checked rotation
error $<$\,$2^{\circ}$ (which guards against the $\sim$$180^{\circ}$
axis-angle singularity that can otherwise false-positive convergence)
and an optional joint-branch consistency cap that rejects solutions
jumping to a far IK branch incompatible with joint-space trajectory
interpolation. The resulting joint trajectory is interpolated and
chunked into \texttt{action\_chunk}s of the same shape the policy
server emits, so the eval client treats primitive and \texttt{vla}
outputs identically.

\paragraph{Internal VLM calls.} The decisions above are produced by
fixed-template VLM prompts (subgoal decomposition, completion check,
failure check, \texttt{grasp}-target naming,
\texttt{place}-destination naming, rewrite, replan). These are
agent-internal procedures rather than tools the agent can choose
among, and are detailed in Appendix~\ref{app:prompts}.


\section{VLM Prompts}
\label{app:prompts}

\textsc{VoLoAgent}'s VLM calls fall into three fixed-template families,
each rendered as a JSON-mode request so the orchestrator can
deterministically parse the response: \emph{plan} (issued once at
episode start, C.1); the per-step \emph{monitoring and acting} call
(C.2), whose exact form depends on the agent variant — tool-chain
mode (\textsc{VoLoAgent} (No VLA)), replan-only mode, and the
replan-with-tools mode used by the full \textsc{VoLoAgent} all share
the same skeleton but expose different action vocabularies; and the
\emph{replan} call (C.3, issued only when monitoring-and-acting
returns the \textsc{replan} action). The Molmo2-point query used
inside \texttt{place(destination)} (Appendix~\ref{app:system}) is a
separate single-purpose VLM call and is shown last (C.4). All prompts
appear verbatim below; full source lives at
\texttt{vlm\_orchestrator/strategies/} and
\texttt{vlm\_orchestrator/failure\_handlers/}.

\paragraph{C.1 Plan} (\texttt{DECOMPOSE\_SHARED\_PROMPT}).
Issued once per episode. Takes the user instruction and the initial
scene image and returns a fresh ordered list of atomic subgoals. The
agent stores the resulting list in slow-context memory (P2).
\textbf{All three monitoring-and-acting variants below
(C.2(a)\,/\,(b)\,/\,(c)) share this C.1 prompt verbatim} — in code,
\texttt{ToolChainStrategy} inherits from \texttt{SubgoalStrategy}, so
the planning call is the same path; only the per-step monitoring call
in C.2 differs across variants.

\begin{promptbox}{Subgoal decomposition prompt}
You are a vision-language assistant helping a robot arm plan tabletop manipulation tasks. You will see the robot's camera view and a task instruction. Produce an ordered list of subgoals the robot should execute one at a time.

Planning principles:
- Atomic actions. Each subgoal is one pick-and-place / push / reorient / open / close action.
- Implicit prerequisites. Add steps the instruction omits when physics requires them (clear an obstacle, reorient an object too large for a side grasp, etc.).
- Avoid unnecessary re-handling. Where physics permit, place each object directly in its final pose on the first lift, instead of staging it at a temporary location.
- Naming. Refer to objects by colour + type ("the yellow banana", "the grey bin"). For visually-identical duplicates use a generic phrase ("a Rubik's cube"). Keep each instruction short (<=15 words) and start with an action verb.
- Ordering. Mark `ordered: true` when a later step depends on an earlier one; mark `false` when steps are independent (e.g. sorting many items into one container).
- Trivial tasks. A single pick-and-place with no obstacles should be one subgoal with `ordered: false`.

Output ONLY valid JSON of the form:
{"subgoals": [{"instruction": "...", "ordered": true|false}, ...]}
\end{promptbox}

\paragraph{C.2 Monitoring and acting.}
The per-step decision call that produces the orchestration action of
the main paper. We use three variants of this prompt depending
on the agent's tool catalog: tool-chain mode (no VLA, only
\texttt{grasp}/\texttt{place}), replan-only mode (VLA only), and
replan-with-tools mode (VLA + \texttt{grasp}/\texttt{place}, used by
\textsc{VoLoAgent} (Full)). The three variants share the same inputs
(overall task, current subgoal, BEFORE / NOW images, last-tool
outcome where applicable) and a status/action JSON output, but expose
different action vocabularies.

\medskip
\noindent\textbf{C.2(a) Tool-chain monitoring and acting}
(\texttt{TOOL\_CHAIN\_SYSTEM\_PROMPT}). Used in
\textsc{VoLoAgent} (No VLA). The agent picks a \texttt{subgoal\_action}
$\in \{\textsc{advance}, \textsc{continue}, \textsc{replan},
\textsc{abort}\}$ \emph{and} a tool $\in \{\texttt{grasp},
\texttt{place}, \texttt{noop}\}$ in the same JSON, with strict pairing
rules (any non-\textsc{continue} action must pair with
\texttt{noop}). The prompt is excerpted to its rules; full anti-example
list lives in \texttt{tool\_chain.py}.

\begin{promptbox}{Tool-chain monitoring-and-acting prompt}
You are driving a robot arm. At every step you (a) judge how the last tool call went and where the plan stands, then (b) pick the next tool to run. Output BOTH decisions in a single JSON object.

Inputs you receive:
  - the overall task instruction
  - the planned subgoal list with the current subgoal index marked
  - a BEFORE image (start of the episode)
  - a NOW image (current scene)
  - the last tool that ran, its arguments, and its outcome
    (DONE / FAILED + failure_reason). These are null on the first cycle.

SUBGOAL ACTION -- judges what has ALREADY happened to the current subgoal as of the NOW image. Pick exactly one:
  "continue" -- the current subgoal is not yet finished in the NOW image. Use this whenever you still need to run a tool (grasp or place) to advance it. THIS IS THE DEFAULT.
  "advance"  -- the current subgoal is ALREADY satisfied in the NOW image. Pair with tool="noop".
  "replan"   -- something went wrong or the plan no longer fits the NOW scene. Pair with tool="noop".
  "abort"    -- task is unrecoverable. Pair with tool="noop".

NEXT TOOL (the action to run THIS cycle):
  grasp(target: str)
    Pick up `target`. Use a noun phrase the detector can localize ("red block", "blue bowl") -- no pronouns, no relations.
  place(destination: str, held_object_hint: str)
    Place the held object at `destination`. PHRASING RULES:
      1. Inside a container -> "in <container>" (e.g. "in the white bowl").
      2. Beside / near an anchor -> prepend "empty space" (e.g. "empty space next to the green lemon").
      3. Open table -> "empty space on the table".
      4. On a surface object -> "on <surface>" (e.g. "on the wire rack shelf").
  noop
    Skip a cycle. REQUIRED for subgoal_action in {advance, replan, abort}.

PAIRING RULES (enforced):
  - "continue" -> tool MUST be "grasp" or "place".
  - "advance" / "replan" / "abort" -> tool MUST be "noop".

OUTPUT: a single JSON object, no markdown, exactly four top-level keys:
  {
    "subgoal_action": "advance" | "continue" | "replan" | "abort",
    "tool":           "grasp"   | "place"    | "noop",
    "args":           { <tool-specific> },
    "reason":         "<one short sentence describing what HAS happened>"
  }
\end{promptbox}

\medskip
\noindent\textbf{C.2(b) Replan-only and (c) Replan-with-tools
monitoring and acting} (\texttt{VLMFailureHandler}). Used in
\textsc{VoLoAgent} (Only VLA) with \texttt{recovery\_mode="replan"}
and in \textsc{VoLoAgent} (Full) with
\texttt{recovery\_mode="replan\_tools"} respectively. The two variants
share a single prompt template; in code,
\texttt{\_build\_system\_prompt} constructs the prompt by
conditionally appending one bullet per available action to the
\texttt{ACTION:} block. The shared skeleton is shown below with a
\texttt{<ACTION BLOCK>} placeholder, followed by the per-action
bullets that are emitted in the order the conditions fire.

\begin{promptbox}{Replan / replan-with-tools monitoring-and-acting prompt (shared skeleton)}
You are monitoring a robot arm executing a tabletop manipulation task.

You receive two sets of images:
- BEFORE images: the scene at the START of the episode (for reference).
- NOW images: the current scene.

You also receive the overall task instruction, the current subgoal, and any remaining subgoals after the current one.

HOW TO REASON:
- Use the BEFORE images and overall task to understand the goal.
- Use the NOW images, current subgoal, and remaining subgoals to understand the current state and what still needs to happen.
- Decide whether the robot can keep executing the current subgoal from the NOW scene. If yes -> "in_progress" / "continue". If the current subgoal is already done -> "complete" / "next". If something actively prevents progress (wrong object in gripper, object dropped, robot stuck) -> "failure".
- The subgoal list may have been replanned mid-episode. The current subgoal and remaining subgoals already reflect what needs to be done from the current state -- do not treat them as stale.

Your job:
1. Assess status -- look at the NOW images and determine the status of the CURRENT SUBGOAL.
2. Decide action -- choose the best action for the robot.

STATUS:
- "complete" -- the current subgoal is done (the target object is at its destination in the NOW images).
- "failure"  -- the robot is clearly stuck, holding the wrong object, or the target object is unreachable.
- "in_progress" -- the subgoal is not yet complete but is still achievable.

ACTION:
<ACTION BLOCK>

IMPORTANT RULES:
- Only say "complete" if you see CLEAR visual evidence the subgoal is done in the NOW images.
- Say "in_progress" if the robot has not yet started or is still working on the current subgoal -- this is NORMAL, not a failure.
- Only say "failure" if the robot is clearly stuck, has picked up the WRONG object, or has been making no progress for a long time.
- When in doubt, say "in_progress" with action "continue".
- Do NOT replan just because the scene looks different from BEFORE. Only replan if the current subgoal is impossible or the plan no longer makes sense given the NOW scene.

Output ONLY valid JSON (no markdown, no explanation):
{"status": "...", "action": "...", "reason": "brief explanation",
 "grasp_target": "...", "place_destination": "...",
 "place_held_object": "..."}
\end{promptbox}

\noindent The \texttt{<ACTION BLOCK>} above is filled by appending the
following bullets, each gated on whether the corresponding action is
in \texttt{available\_actions}. \textbf{Replan-only} (b) emits the
\texttt{next} / \texttt{continue} / \texttt{replan} bullets only;
\textbf{replan-with-tools} (c) additionally emits the
\texttt{grasp\_tool} and \texttt{place\_tool} bullets.

\begin{promptbox}{Action bullets (conditionally appended to the shared skeleton)}
# always present in (b) and (c):
- "next"     -- this subgoal is COMPLETE. The target object has moved to its destination. Advance to the next subgoal.
- "continue" -- the robot is making progress or the subgoal is not yet complete. Keep working with the current instruction.
- "replan"   -- something went wrong (object dropped, stuck, wrong object picked, no visible progress) OR the current plan is not working. Replan the remaining subtasks from the current scene state.

# (c) only -- appended when ACTION_GRASP is available:
- "grasp_tool" -- fire only for WRONG-OBJECT recovery: the gripper holds (or is closing on) an object whose identity differs from the subgoal's target. Do NOT fire while the robot is still attempting the correct object -- that is in-progress execution; emit "continue". When firing, set "grasp_target" to the noun phrase of the correct object.

# (c) only -- appended when ACTION_PLACE is available:
- "place_tool" -- fire only for WRONG-DESTINATION recovery: the held object is heading to or already released at a destination different from the subgoal's destination. Do NOT fire while the robot is en route to the correct destination -- that is in-progress execution; emit "continue". When firing, set "place_destination" to a FREEFORM destination phrase. PHRASING RULES:
    1. Inside a container -> "in <container>" (e.g. "in the white bowl").
    2. Beside / near an anchor -> prepend "empty space" (e.g. "empty space next to the green lemon").
    3. Open table -> "empty space on the table".
    4. On a surface object -> "on <surface>" (e.g. "on the wire rack shelf").
  Optionally also set "place_held_object" with the noun phrase of the object in the gripper.
  LIMITATION: place_tool is unreliable for high containers (e.g. a shelf, a high vase). For those destinations prefer "continue" (let the VLA finish) or "replan" instead of firing place_tool.
\end{promptbox}

\noindent The C.2(a) tool-chain prompt and the C.2(b)/(c) prompts are
disjoint paths in the codebase: C.2(a) drives \texttt{tool\_chain.py},
while C.2(b)/(c) drive the subgoal strategy with the failure handler.

\paragraph{C.3 Replan} (\texttt{RECYCLE\_SYSTEM\_PROMPT}).
Issued only when monitoring-and-acting (C.2) returns the
\textsc{replan} action. Unlike C.1, replan is \emph{not} a fresh
re-decomposition: it takes the BEFORE image, the NOW image, the
original instruction, and the \emph{original subgoal list}, and
returns the subset of those subgoals that still need to be executed
(plus a \texttt{done} flag if every subgoal is already satisfied).
The prompt is framed as a completion checker that, in the same call,
emits the remaining subgoals as its primary output --- hence the
JSON schema below. It explicitly forbids inventing new subgoals or
rephrasing existing ones, which keeps the agent on the original plan
after transient failures (dropped object, mis-grasp) instead of
drifting through paraphrases on every recovery.

\begin{promptbox}{Replan prompt}
You check whether a robot finished a pick-and-place task by comparing BEFORE and NOW images of the scene.

Rules:
- Only consider objects matching the task instruction. Ignore all others (figurines, decorations, containers, robot arm, etc.).
- An object is DONE if it moved from its BEFORE table position into the target container / stacking position.
- An object is REMAINING if it is still clearly visible on the table in the NOW image, or if the physics engine reports it as regressed.
- Do NOT invent objects you cannot clearly see on the table in NOW.
- Rubik's cubes are standard multi-colored cubes -- never name them by a single face color. Just call them "Rubik's cube".
- If you are unsure whether an object is still on the table or already in the correct position, assume it is DONE (be conservative -- fewer subgoals is better than hallucinated ones).
- IMPORTANT: When an ORIGINAL SUBGOAL LIST is provided, you MUST select subgoals ONLY from that list. Do NOT rephrase, rewrite, or invent new subgoal instructions. Return the exact text of the original subgoals that still need to be completed.

You MUST respond with ONLY a single JSON object -- no analysis, no bullet points, no markdown, no explanation before or after.

JSON format (task not done):
{"remaining": ["green cube"],
 "subgoals": [{"instruction": "Pick up the green cube and place it in the bin",
               "target_object": "green cube"}],
 "done": false}

JSON format (task complete):
{"remaining": [], "subgoals": [], "done": true}
\end{promptbox}

\paragraph{C.4 Place-tool 2-D point query.}
When \texttt{place(destination)} fires (Appendix~\ref{app:system}) the
2-D destination point is obtained either by a direct VLM-point query
(below) or by \emph{Molmo2-point}~\citep{molmo2}, which uses the same
phrasing but returns a sharper pixel coordinate for tight placements.
The output is then projected to 3-D via the depth image.

\begin{promptbox}{Place-destination 2-D point query}
You are given a robot's camera view and a destination phrase. Return the pixel coordinate (x, y) where the gripper should release the held object so that, after release, the object satisfies the destination phrase.

Rules:
- Coordinates are in image pixels with (0, 0) at the top-left.
- For "in <container>", point at the centre of the container's interior.
- For "empty space next to <anchor>", point at a clear table region next to the anchor, not on top of any object.
- For "on <surface>", point at a stable region of the surface.
- If multiple candidates match, prefer the one closest to the gripper.

Destination phrase: <PHRASE>

Output ONLY valid JSON: {"x": <int>, "y": <int>, "reason": "..."}
\end{promptbox}

\paragraph{Front-camera variant.}
When \texttt{--use-front-camera} is active, the orchestrator appends a short note
to each prompt above instructing the VLM that the front-camera view is
both left/right and front/behind \emph{flipped} relative to the robot
frame, and asking it to describe targets by visual features (colour,
type, proximity to landmarks) rather than directional words. Three
near-identical variants of this note exist in code, one per prompt
family.

\begin{promptbox}{Front-camera note appended to C.1 Plan and C.3 Replan (\texttt{\_FRONT\_CAM\_NOTE})}
NOTE on camera orientation: the "Front camera" view is BOTH L/R and front/behind FLIPPED from the robot's perspective. image-LEFT <-> robot-RIGHT, image-TOP <-> robot-BEHIND.

To avoid frame confusion, describe targets by VISUAL FEATURES (color, type, proximity to landmarks) in every output field -- NOT by left/right/front/behind. Example: "the grey container next to the red block" instead of "the right bin". If the user's instruction uses a direction, first apply the flip to find which object they mean in the image, then re-describe by visual features.
\end{promptbox}

\begin{promptbox}{Front-camera note appended to C.2(b)/(c) monitoring-and-acting (failure-handler variant)}
NOTE on camera orientation: the "Front camera" view is BOTH L/R and front/behind FLIPPED from the robot's perspective. image-LEFT <-> robot-RIGHT, image-TOP <-> robot-BEHIND.

To avoid frame confusion, describe targets by VISUAL FEATURES (color, type, proximity to landmarks) in every output field -- NOT by left/right/front/behind. Example: "the grey container next to the red block" instead of "the right bin". If the subgoal uses a direction, first apply the flip to find which object it means in the image, then re-describe by visual features.
\end{promptbox}

\begin{promptbox}{Front-camera note appended to C.2(a) tool-chain (\texttt{\_TOOL\_CHAIN\_FRONT\_CAM\_NOTE})}
NOTE on camera orientation: the "Front camera" view is BOTH L/R and front/behind FLIPPED from the robot's perspective. image-LEFT <-> robot-RIGHT, image-TOP <-> robot-BEHIND.

To avoid frame confusion, describe targets by VISUAL FEATURES (color, type, proximity to landmarks) in every `target` / `destination` -- NOT by left/right/front/behind. Example: "the grey container next to the red block" instead of "the right bin". If the subgoal uses a direction, first apply the flip to find which object it means in the image, then re-describe by visual features.
\end{promptbox}


\section{\textsc{RoboVoLo} Benchmark Details}
\label{app:bench-tasks}

\textsc{RoboVoLo} comprises 126 tasks across four suites, grouped
into 15 task categories. Each
suite targets one diagnostic axis: scene-context grounding for
Common Sense, state tracking for Memory, language-reference
resolution for Complex References, and external-knowledge
application for World Knowledge. Each task is authored so that
instruction-independent behaviour (e.g.\ a fixed ``put everything
in the bin'' policy) cannot succeed. Table~\ref{tab:bench-cats}
defines all 15 categories: their testing purpose, the cognitive
skill being probed, the number of tasks, and one representative
instruction. The full task list with per-task initial-state
screenshots is in the released benchmark repository.

\begin{table}[h!]
\centering
\small
\setlength{\tabcolsep}{4pt}
\renewcommand{\arraystretch}{1.15}
\caption{\textbf{\textsc{RoboVoLo} task categories.} 15 categories
across four suites, with the testing purpose and a representative
instruction. \emph{\#} is the number of distinct tasks per
category; each is run for $T{=}3$ seeded trials.}
\label{tab:bench-cats}
\resizebox{\linewidth}{!}{%
\begin{tabular}{@{}llp{5.6cm}cp{4.6cm}@{}}
\toprule
\textbf{Suite} & \textbf{Category} & \textbf{Testing purpose} & \textbf{\#} & \textbf{Example instruction} \\
\midrule
\multirow{4}{*}{\shortstack[l]{\textbf{Common}\\\textbf{Sense (CS)}}}
 & Infer    & Infer the implicit goal from scene context when the instruction is under-specified. & 8 & \textit{``Sort the cans on the table into the correct bowls.''} \\
 & Kit      & Assemble a coherent set from scattered objects (kitting / packing). & 8 & \textit{``Set up the table for breakfast.''} \\
 & Recover  & Detect an out-of-place object and restore the expected configuration. & 8 & \textit{``Something fell, put it back where it belongs.''} \\
 & Sort     & Group objects by category, function, or container affinity. & 8 & \textit{``One item is in the wrong container -- move it to where it belongs.''} \\
\midrule
\multirow{3}{*}{\textbf{Memory}}
 & Order    & Track or reproduce an ordered sequence (e.g.\ stack reversal, line ordering, cyclic rotation). & 10 & \textit{``Reverse the two-block stack.''} \\
 & Recall   & Recall an earlier scene state hidden by intermediate manipulations. & 10 & \textit{``Unstack all four blocks onto the table, then put the two that were on top into the bin.''} \\
 & Swap     & Exchange the positions of two objects without losing track of either. & 10 & \textit{``Swap the contents of the left and right bins.''} \\
\midrule
\multirow{4}{*}{\shortstack[l]{\textbf{Complex}\\\textbf{Refs (CR)}}}
 & Spatial   & Resolve spatial-relation references (left/right, behind, between, nearest). & 8 & \textit{``Put the fruit behind the bowl into the bin.''} \\
 & Counting  & Resolve ordinal / counting references (leftmost, second, every-other, $n$-th from end). & 8 & \textit{``Five fruits are lined up. Put the second and the fourth from the left onto the cutting board.''} \\
 & Negation  & Handle negative references (everything except $X$, items not in a group). & 8 & \textit{``Move every item except the cans into the bin.''} \\
 & Size+Sort & Resolve size-based references and combine with sorting. & 8 & \textit{``Put the smaller fruits in the small bowl and the larger ones in the large bowl.''} \\
\midrule
\multirow{4}{*}{\shortstack[l]{\textbf{World}\\\textbf{Know. (WK)}}}
 & Art       & Compose stylised pictures by arranging shape/colour primitives. & 8 & \textit{``Use the blocks to make a stick figure.''} \\
 & Chem      & Assemble chemical formulas from periodic-table element cubes. & 8 & \textit{``Build the chemical formula for water from the periodic-table cubes.''} \\
 & Math      & Solve simple arithmetic by arranging digit/operator cubes. & 8 & \textit{``Use the cubes to make an equation that equals seven.''} \\
 & Recycle   & Sort items by material / recyclability. & 8 & \textit{``Place the recyclable items in the blue bin and the trash in the grey bin.''} \\
\bottomrule
\end{tabular}%
}
\end{table}

\begin{figure}[h!]
  \centering
  \begin{subfigure}[t]{0.235\linewidth}
    \centering
    \includegraphics[width=\linewidth]{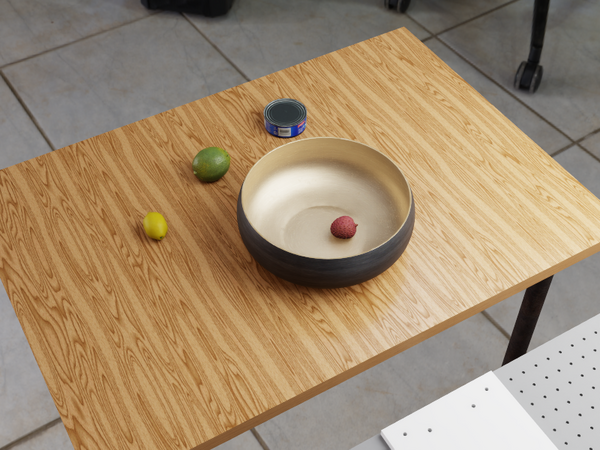}
    \caption{\textit{Infer.} ``Most items on the table are the same kind. Put them in the bowl and leave the odd one out.''}
  \end{subfigure}
  \hfill
  \begin{subfigure}[t]{0.235\linewidth}
    \centering
    \includegraphics[width=\linewidth]{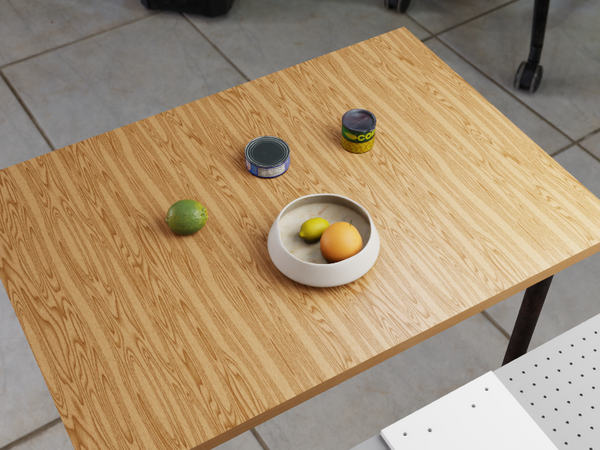}
    \caption{\textit{Infer.} ``One item on the table belongs with the group in the bowl. Put it there. Leave the rest.''}
  \end{subfigure}
  \hfill
  \begin{subfigure}[t]{0.235\linewidth}
    \centering
    \includegraphics[width=\linewidth]{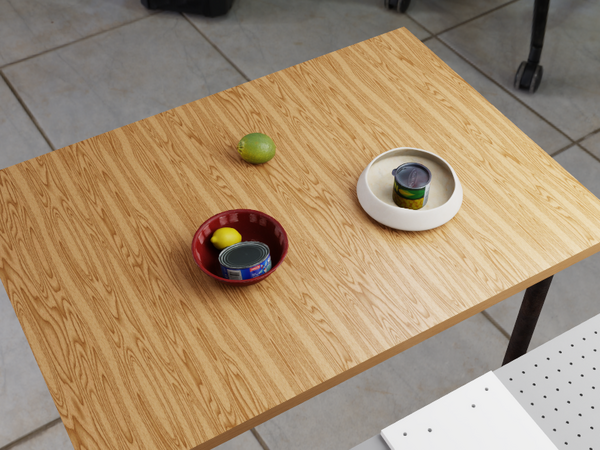}
    \caption{\textit{Kit.} ``Each bowl should contain a can and a fruit.''}
  \end{subfigure}
  \hfill
  \begin{subfigure}[t]{0.235\linewidth}
    \centering
    \includegraphics[width=\linewidth]{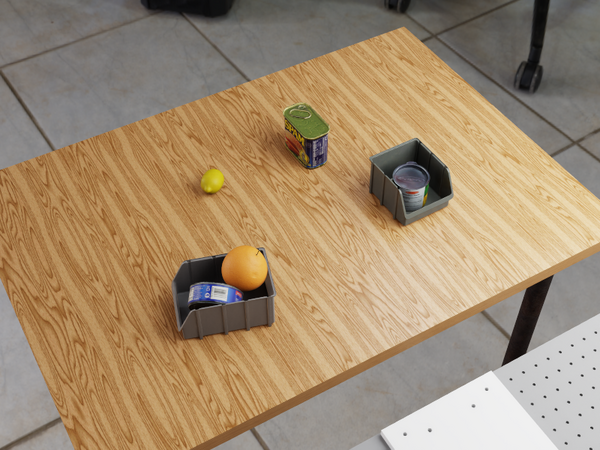}
    \caption{\textit{Kit.} ``One bin has a complete set. Make the other one match.''}
  \end{subfigure}
  \\[4pt]
  \begin{subfigure}[t]{0.235\linewidth}
    \centering
    \includegraphics[width=\linewidth]{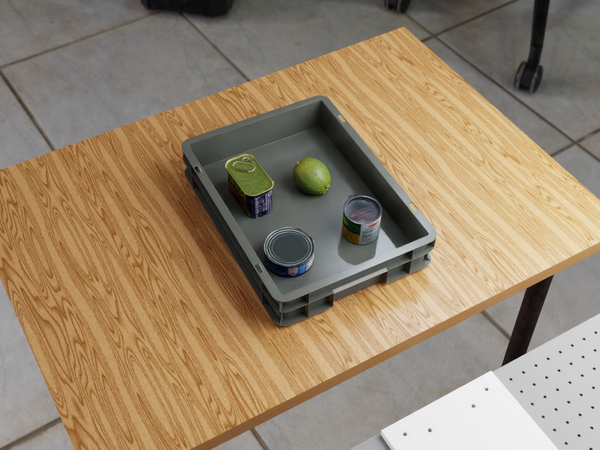}
    \caption{\textit{Recover.} ``One item in the bin doesn't belong with the others. Take it out.''}
  \end{subfigure}
  \hfill
  \begin{subfigure}[t]{0.235\linewidth}
    \centering
    \includegraphics[width=\linewidth]{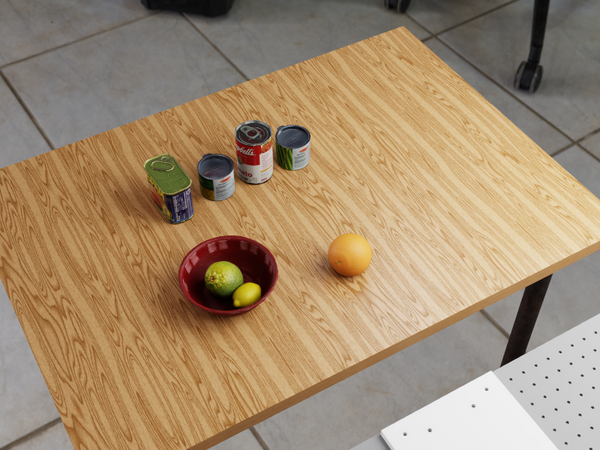}
    \caption{\textit{Recover.} ``An object that belongs in the bowl has fallen out. Put it back.''}
  \end{subfigure}
  \hfill
  \begin{subfigure}[t]{0.235\linewidth}
    \centering
    \includegraphics[width=\linewidth]{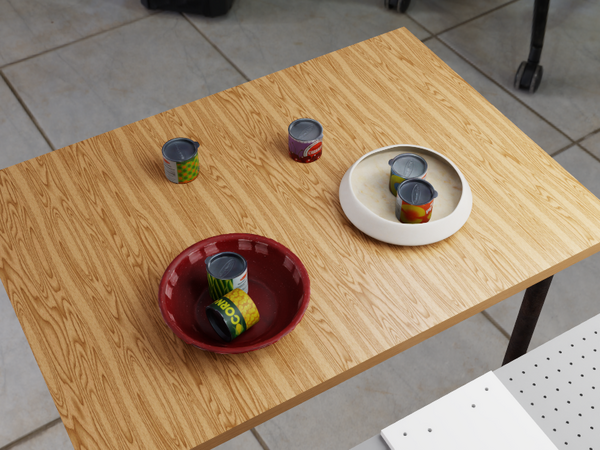}
    \caption{\textit{Sort.} ``Sort the cans on the table into the correct bowls.''}
  \end{subfigure}
  \hfill
  \begin{subfigure}[t]{0.235\linewidth}
    \centering
    \includegraphics[width=\linewidth]{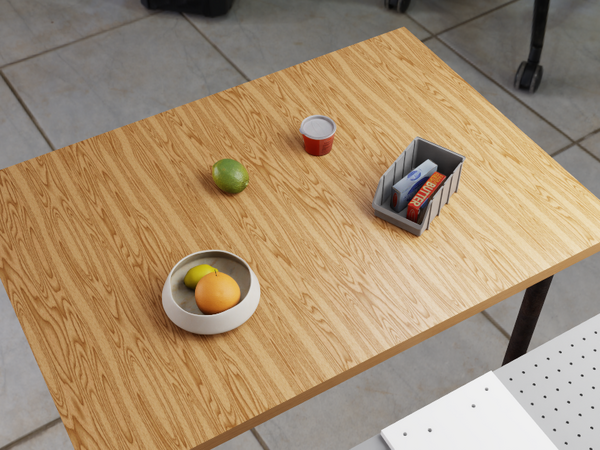}
    \caption{\textit{Sort.} ``Each container holds a category. Put the table items where they belong.''}
  \end{subfigure}
  \caption{\textbf{Common Sense suite -- task examples.} 8 representative initial-scene views from the Common Sense suite of \textsc{RoboVoLo}; each panel shows the task category (italics) and its instruction.}
  \label{fig:bench-cs}
\end{figure}

\begin{figure}[h!]
  \centering
  \begin{subfigure}[t]{0.31\linewidth}
    \centering
    \includegraphics[width=\linewidth]{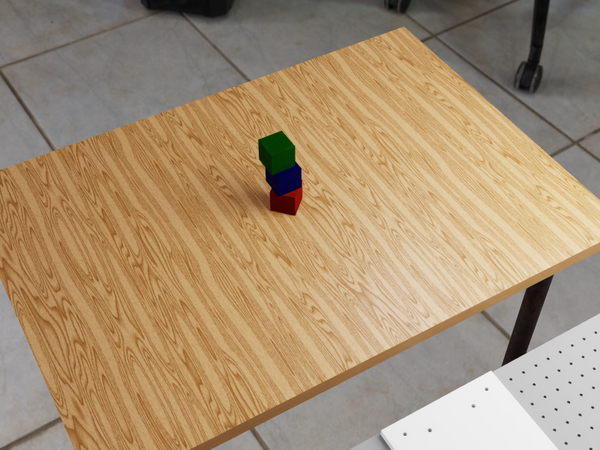}
    \caption{\textit{Order.} ``Reverse the stacking order of the three colored blocks.''}
  \end{subfigure}
  \hfill
  \begin{subfigure}[t]{0.31\linewidth}
    \centering
    \includegraphics[width=\linewidth]{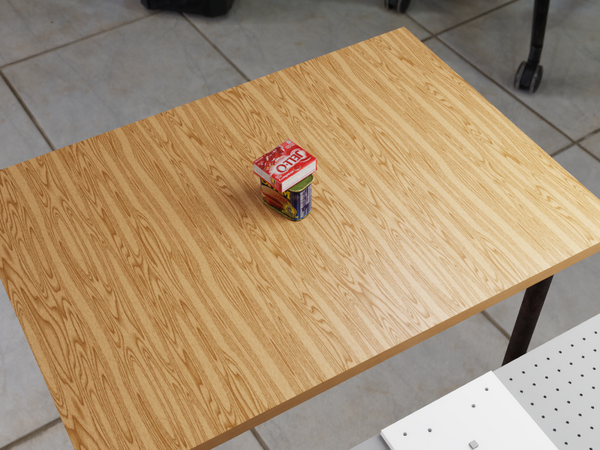}
    \caption{\textit{Order.} ``Reverse the two-block stack.''}
  \end{subfigure}
  \hfill
  \begin{subfigure}[t]{0.31\linewidth}
    \centering
    \includegraphics[width=\linewidth]{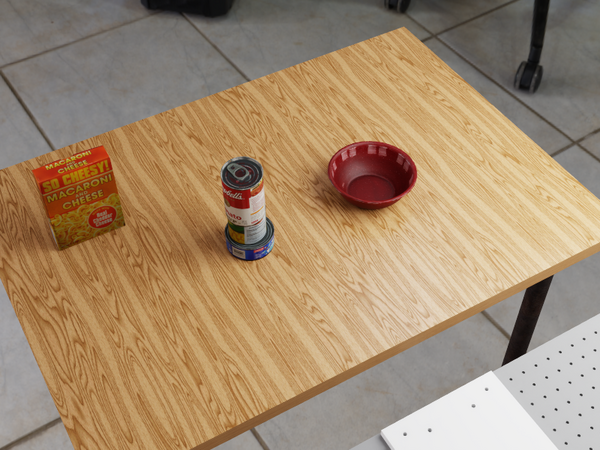}
    \caption{\textit{Recall.} ``Unstack all the cans, then put only the top and bottom cans from the original stack into the bowl -- leave the middle can on the table.''}
  \end{subfigure}
  \\[4pt]
  \begin{subfigure}[t]{0.31\linewidth}
    \centering
    \includegraphics[width=\linewidth]{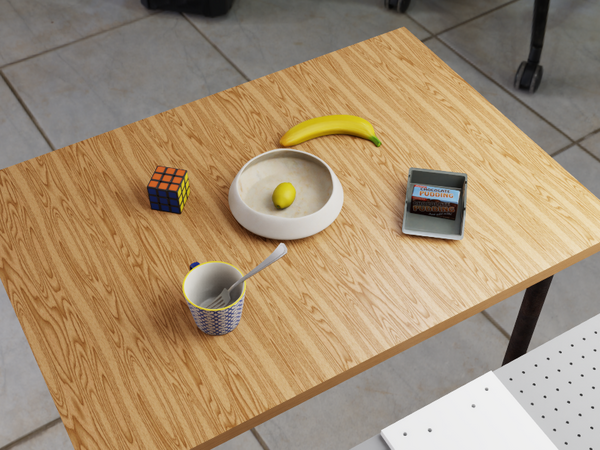}
    \caption{\textit{Recall.} ``Take all objects out of the three containers and place them on the table, then put the objects that were originally on the table into the serving bowl.''}
  \end{subfigure}
  \hfill
  \begin{subfigure}[t]{0.31\linewidth}
    \centering
    \includegraphics[width=\linewidth]{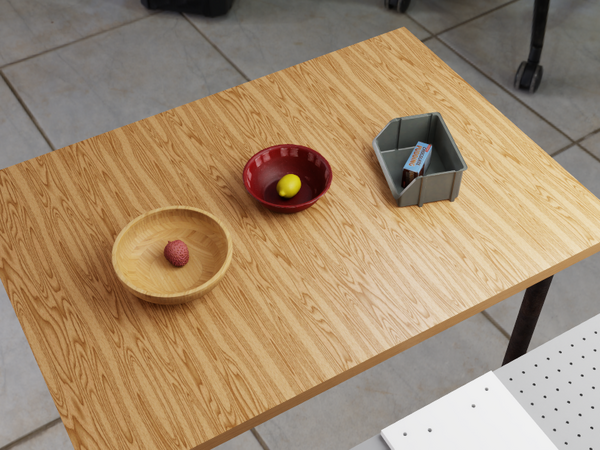}
    \caption{\textit{Swap.} ``Cycle the items through three containers: move the wooden bowl's item to the plastic bowl, the plastic bowl's item to the bin, and the bin's item to the wooden bowl.''}
  \end{subfigure}
  \hfill
  \begin{subfigure}[t]{0.31\linewidth}
    \centering
    \includegraphics[width=\linewidth]{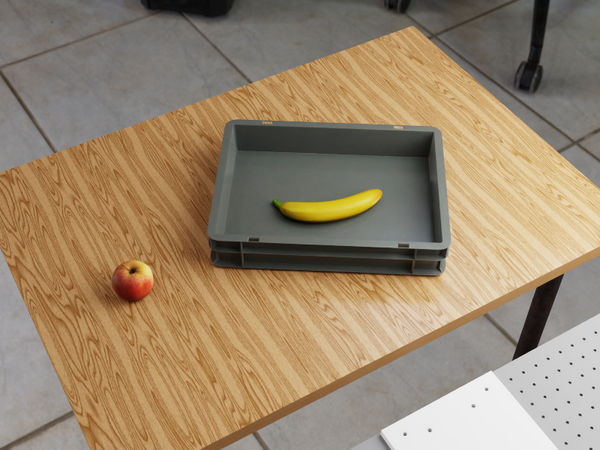}
    \caption{\textit{Swap.} ``Take the item out of the container and place it on the table. Then put the item that was on the table into the container.''}
  \end{subfigure}
  \caption{\textbf{Memory suite -- task examples.} 6 representative initial-scene views from the Memory suite of \textsc{RoboVoLo}; each panel shows the task category (italics) and its instruction.}
  \label{fig:bench-mem}
\end{figure}

\begin{figure}[h!]
  \centering
  \begin{subfigure}[t]{0.235\linewidth}
    \centering
    \includegraphics[width=\linewidth]{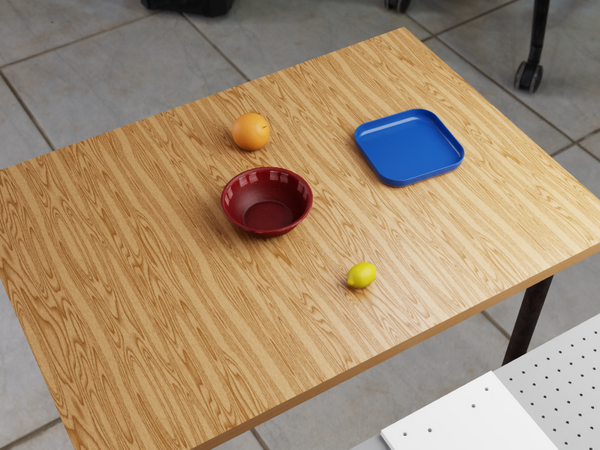}
    \caption{\textit{Spatial.} ``Put the fruit that is behind the bowl, from the robot's perspective, onto the plate.''}
  \end{subfigure}
  \hfill
  \begin{subfigure}[t]{0.235\linewidth}
    \centering
    \includegraphics[width=\linewidth]{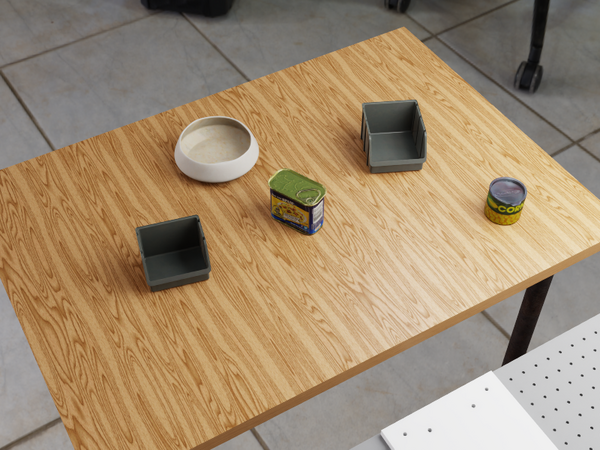}
    \caption{\textit{Spatial.} ``Put only the item that is between the two bins into the bowl.''}
  \end{subfigure}
  \hfill
  \begin{subfigure}[t]{0.235\linewidth}
    \centering
    \includegraphics[width=\linewidth]{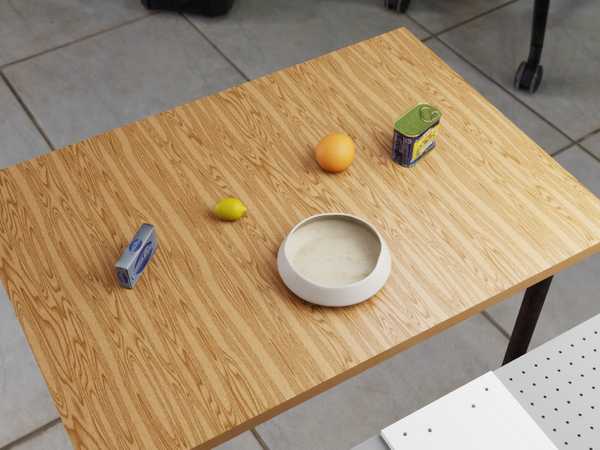}
    \caption{\textit{Counting.} ``Four items are in a row. Put the first and the last into the bowl.''}
  \end{subfigure}
  \hfill
  \begin{subfigure}[t]{0.235\linewidth}
    \centering
    \includegraphics[width=\linewidth]{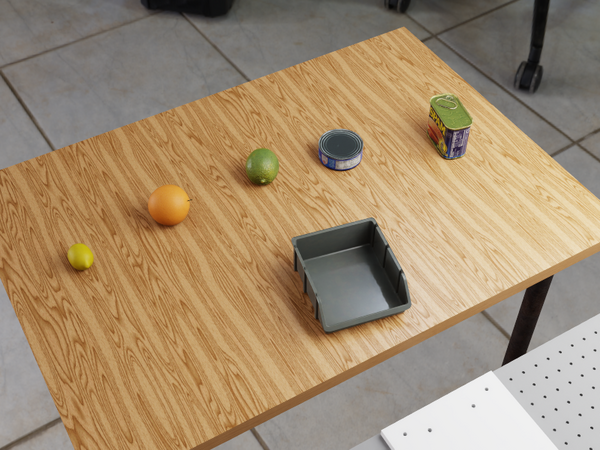}
    \caption{\textit{Counting.} ``Five items are in a row. Put the three leftmost ones into the bin.''}
  \end{subfigure}
  \\[4pt]
  \begin{subfigure}[t]{0.235\linewidth}
    \centering
    \includegraphics[width=\linewidth]{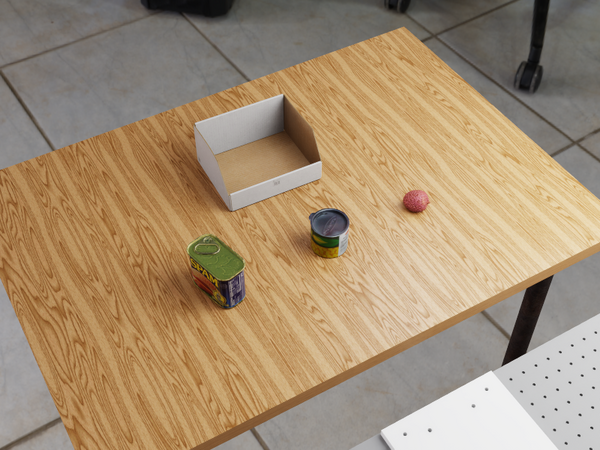}
    \caption{\textit{Size+Sort.} ``Put the largest and the smallest items into the bin -- leave the middle-sized one on the table.''}
  \end{subfigure}
  \hfill
  \begin{subfigure}[t]{0.235\linewidth}
    \centering
    \includegraphics[width=\linewidth]{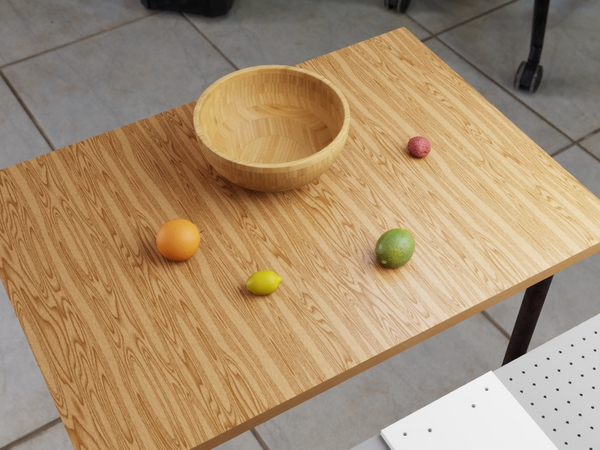}
    \caption{\textit{Size+Sort.} ``Move all fruits that are smaller than the orange into the bowl.''}
  \end{subfigure}
  \hfill
  \begin{subfigure}[t]{0.235\linewidth}
    \centering
    \includegraphics[width=\linewidth]{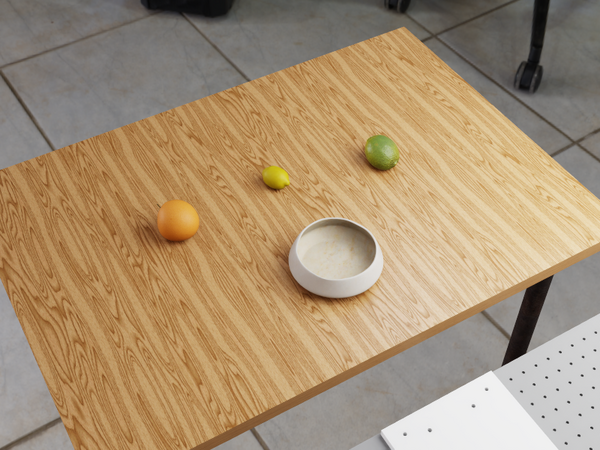}
    \caption{\textit{Negation.} ``Put all the fruits into the bowl except the orange.''}
  \end{subfigure}
  \hfill
  \begin{subfigure}[t]{0.235\linewidth}
    \centering
    \includegraphics[width=\linewidth]{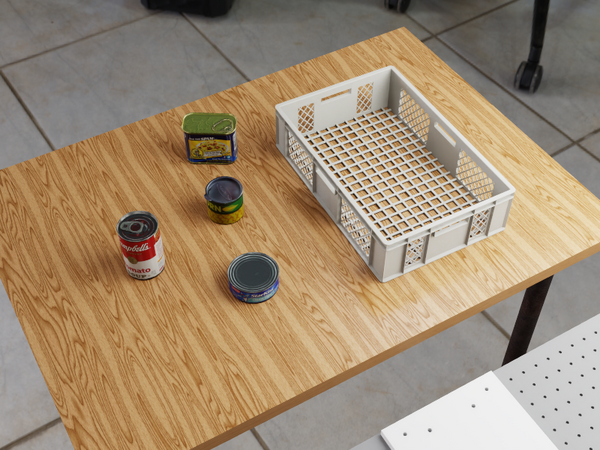}
    \caption{\textit{Negation.} ``Put all the cans into the crate except the tallest one.''}
  \end{subfigure}
  \caption{\textbf{Complex References suite -- task examples.} 8 representative initial-scene views from the Complex References suite of \textsc{RoboVoLo}; each panel shows the task category (italics) and its instruction.}
  \label{fig:bench-cr}
\end{figure}

\begin{figure}[h!]
  \centering
  \begin{subfigure}[t]{0.235\linewidth}
    \centering
    \includegraphics[width=\linewidth]{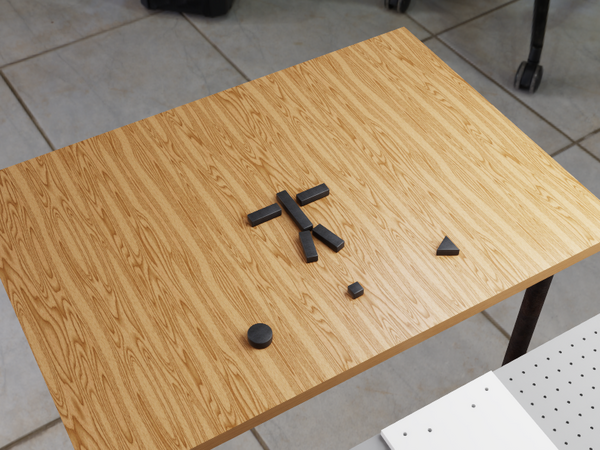}
    \caption{\textit{Art.} ``Complete the stick figure by placing the missing head.''}
  \end{subfigure}
  \hfill
  \begin{subfigure}[t]{0.235\linewidth}
    \centering
    \includegraphics[width=\linewidth]{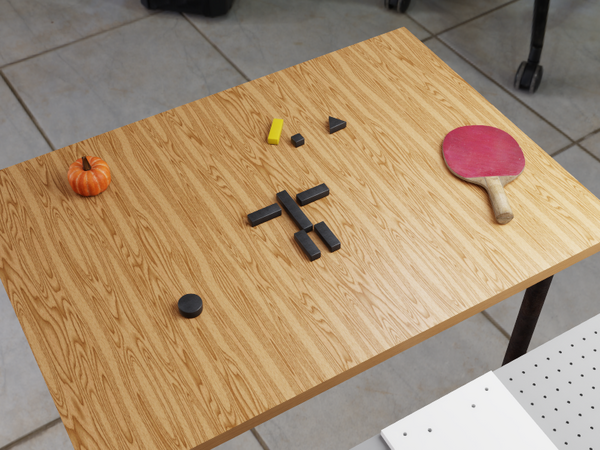}
    \caption{\textit{Art.} ``Complete the stick figure by placing the missing head.''}
  \end{subfigure}
  \hfill
  \begin{subfigure}[t]{0.235\linewidth}
    \centering
    \includegraphics[width=\linewidth]{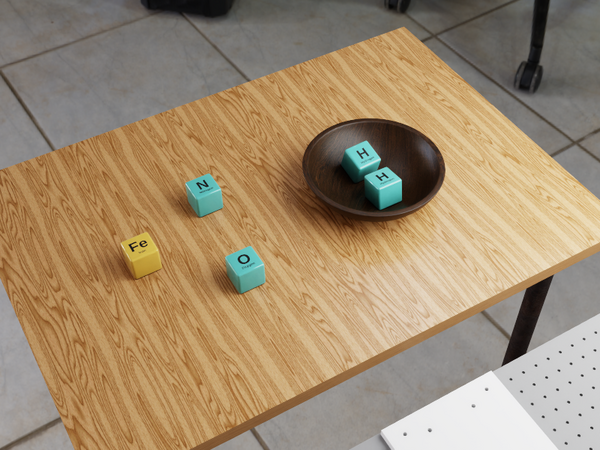}
    \caption{\textit{Chem.} ``Complete the water molecule by adding the missing element to the bowl.''}
  \end{subfigure}
  \hfill
  \begin{subfigure}[t]{0.235\linewidth}
    \centering
    \includegraphics[width=\linewidth]{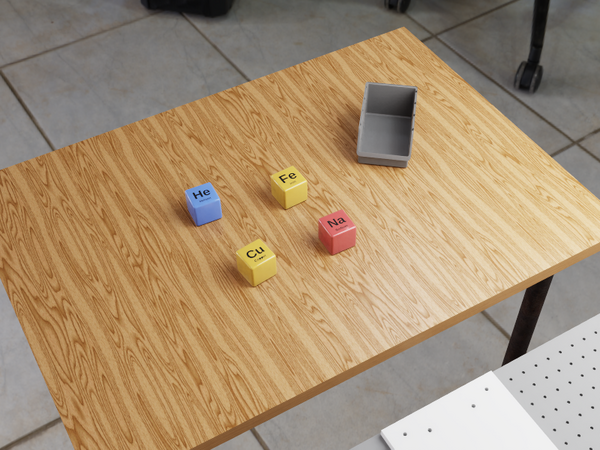}
    \caption{\textit{Chem.} ``Put the non-reactive noble gas into the bin.''}
  \end{subfigure}
  \\[4pt]
  \begin{subfigure}[t]{0.235\linewidth}
    \centering
    \includegraphics[width=\linewidth]{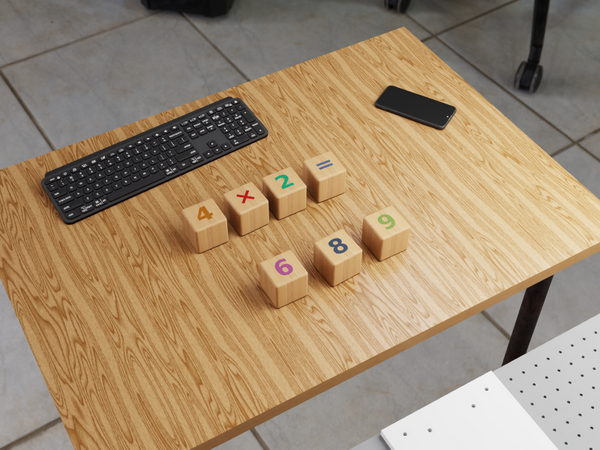}
    \caption{\textit{Math.} ``Move a cube to complete the equation.''}
  \end{subfigure}
  \hfill
  \begin{subfigure}[t]{0.235\linewidth}
    \centering
    \includegraphics[width=\linewidth]{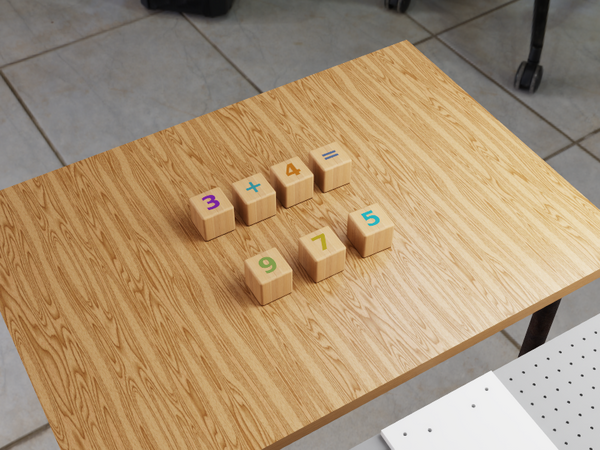}
    \caption{\textit{Math.} ``Move a cube to complete the equation.''}
  \end{subfigure}
  \hfill
  \begin{subfigure}[t]{0.235\linewidth}
    \centering
    \includegraphics[width=\linewidth]{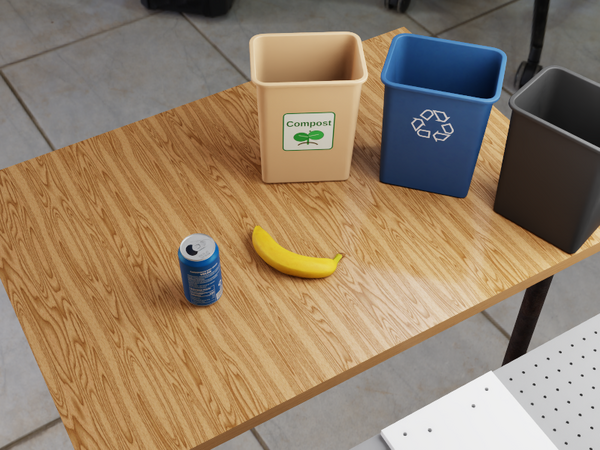}
    \caption{\textit{Recycle.} ``Sort 2 items into correct bins.''}
  \end{subfigure}
  \hfill
  \begin{subfigure}[t]{0.235\linewidth}
    \centering
    \includegraphics[width=\linewidth]{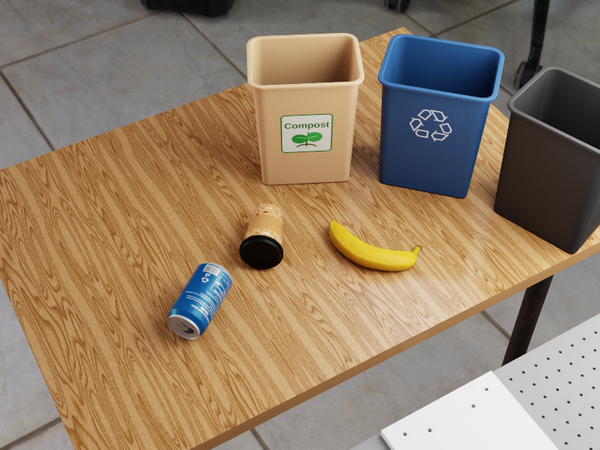}
    \caption{\textit{Recycle.} ``Sort 3 items into correct bins.''}
  \end{subfigure}
  \caption{\textbf{World Knowledge suite -- task examples.} 8 representative initial-scene views from the World Knowledge suite of \textsc{RoboVoLo}; each panel shows the task category (italics) and its instruction.}
  \label{fig:bench-wk}
\end{figure}

\paragraph{Task examples per suite.}
Figures~\ref{fig:bench-cs}--\ref{fig:bench-wk} show two representative initial-scene views per category for each suite, drawn from the front-camera view used by the orchestrator. Each panel's subcaption gives the task category (matching Table~\ref{tab:bench-cats}) and the natural-language instruction issued to the agent.

\section{Simulation Setup Details}
\label{app:sim-setup}

\textsc{RoboVoLo} is implemented on top of RoboLab~\citep{robolab} on
NVIDIA Isaac Sim / Isaac Lab~\citep{mittal2025isaaclab} with PhysX as
the physics engine. This appendix complements the main paper with
the embodiment, observation, and evaluation-protocol details needed
to reproduce a run.

\paragraph{Robot embodiment and action space.}
All policy models use the DROID configuration~\citep{droid}: a 7-DoF
Franka Research 3 arm with a Robotiq 2F-85 parallel-jaw gripper.
Actions are 8-D --- seven joint-position targets plus one binary
gripper open/close --- streamed in 8-step chunks at the
$\pi_{0.5}$~\citep{pi05} control rate of $15$\,Hz. The orchestrator's
\texttt{grasp}\,/\,\texttt{place} primitives emit chunks of the same
shape so the eval client treats them identically to VLA outputs.

\paragraph{Cameras and observations.}
The simulator exposes three RGB cameras whose intrinsics and
extrinsics match the real DROID cell: an exterior ZED 2i (used by
the VLA), a wrist ZED mini (used by the VLA), and a front-mounted
egocentric ZED 2i (used by the orchestrator's monitor and tools when
\texttt{--use-front-camera} is set; see Appendix~\ref{app:prompts}).
Camera images are $224\!\times\!224$ after canonical resize.
Synthetic depth from PhysX is forwarded alongside RGB; the
\texttt{grasp} primitive consumes the front-camera depth, and the
\texttt{place} primitive consumes the front-camera depth together
with a Molmo2 2-D point. The VLA does \emph{not} see depth.

\begin{figure}[h]
  \centering
  \includegraphics[width=\linewidth]{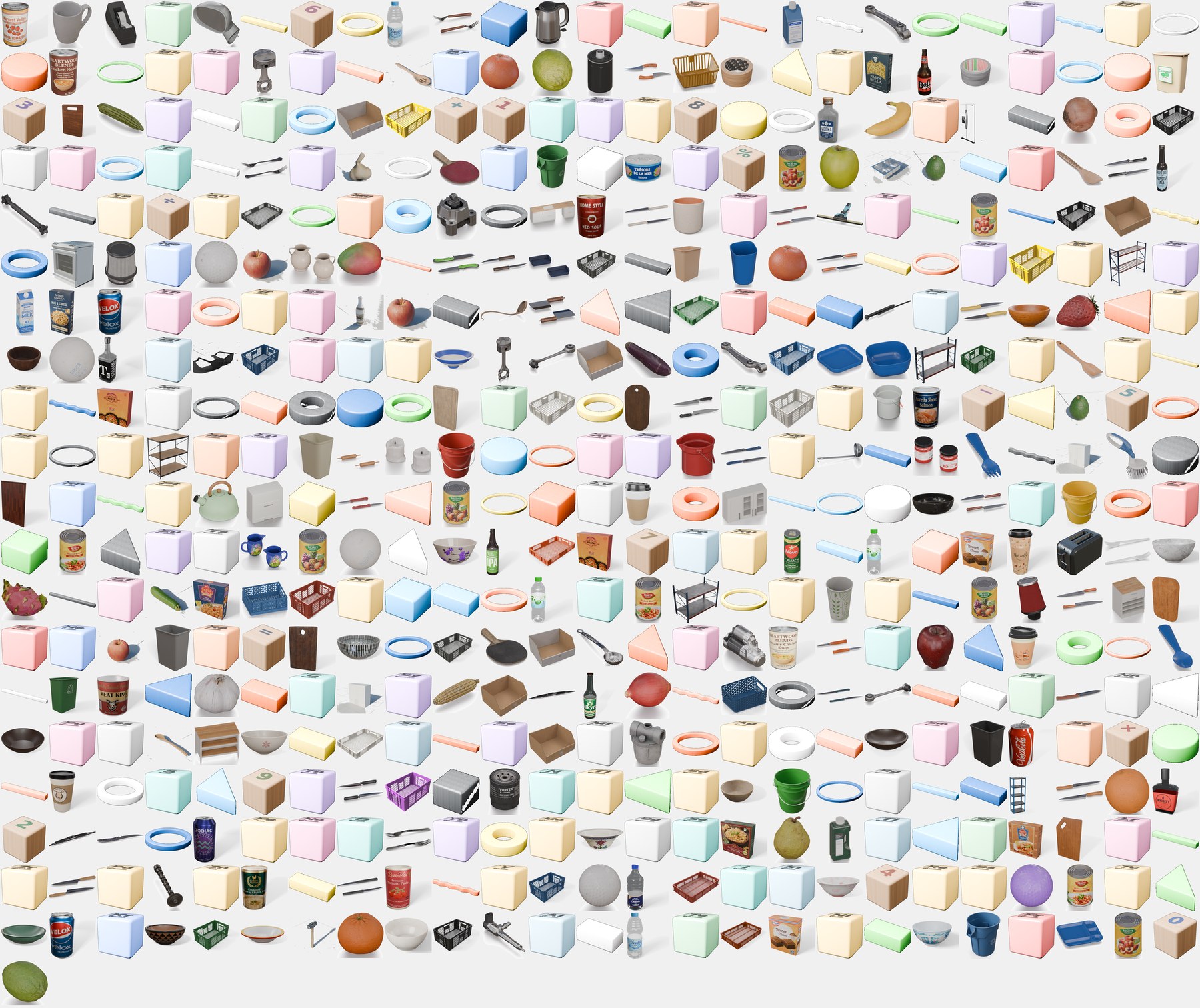}
  \caption{\textbf{The 501 new \textsc{RoboVoLo} assets.} Every object
  added on top of RoboLab's existing library is shown: $247$
  Lightwheel SimReady household items plus $254$ task-specific
  assets ($118$ periodic-table element cubes, $120$ geometric art
  primitives, $16$ math digit/operator cubes). Each tile is the
  Isaac Sim render of a single asset, randomly ordered.}
  \label{fig:asset-gallery}
\end{figure}

\paragraph{Scene assets.}
\textsc{RoboVoLo} expands RoboLab's asset library with the $501$
objects described in the main paper ($247$ Lightwheel SimReady
household items and $254$ task-specific assets: $118$ periodic-table
element cubes, $120$ geometric art primitives, and $16$ wooden
digit/operator cubes for math). Figure~\ref{fig:asset-gallery}
visualises all $501$ items at once, each rendered in isolation in
Isaac Sim and tiled into a single panel. All assets carry collision
geometry and physically-realistic mass / friction / restitution so
PhysX rigid-body dynamics produce contact behaviour usable by a
depth-based grasp pipeline.

\paragraph{Trial protocol and success criterion.}
Each task is evaluated for $T{=}3$ trials. Trials share the canonical
initial scene (object identity and slot assignment); within-trial
randomization perturbs initial object poses, lighting, and any
distractor placements according to a fixed seed indexed by trial,
so trial $k$ of system $A$ and trial $k$ of system $B$ start from
identical states and admit paired comparison
(Appendix~\ref{app:stat-detail}). Success is determined by the
RoboLab task's authored termination predicate (object-in-container,
final pose tolerance, etc.); episodes that do not satisfy it before
the per-task simulation-step budget elapses count as failures.

\paragraph{Compute.}
A single trial of \textsc{VoLoAgent} (Full) consumes roughly one L40
GPU for the simulator + grasp server and one H100 (or a second L40
for $\pi_{0.5}$, or two H100 for $14$B DreamZero) for the VLA, plus a
few cents of cloud-VLM inference. The 126-task $\times$ $3$-trial
\textsc{RoboVoLo} sweep takes $\sim\!18$ GPU-hours per system.


\section{Statistic Test Details}
\label{app:stat-detail}

This appendix documents the statistical procedures behind every
significance claim in the main paper: paired sign-flip randomization tests for
the simulation Main-Results and Component-Ablation tables in the main
paper (\S\ref{app:stat-sim}), and Wilson score confidence intervals
for the Real-Robot table in the main paper (\S\ref{app:stat-real}).

\subsection{Paired Sign-Flip Randomization Test (Main-Results and Component-Ablation tables in the main paper)}
\label{app:stat-sim}

\paragraph{Setup.} Fix two methods $A$ and $B$ and the set of tasks
$\mathcal{T}$ on which both methods were run with $K{=}3$ matched-seed
trials per task. For task $i\in\mathcal{T}$ let $s^A_i, s^B_i \in
\{0,1,\dots,K\}$ denote the number of successful trials and define the
per-task success fractions
\begin{equation}
\hat{p}^A_i \;=\; \frac{s^A_i}{K}, \qquad
\hat{p}^B_i \;=\; \frac{s^B_i}{K}, \qquad
d_i \;=\; \hat{p}^A_i - \hat{p}^B_i \;\in\; \{-1, -\tfrac{2}{3}, -\tfrac{1}{3}, 0, \tfrac{1}{3}, \tfrac{2}{3}, 1\}.
\label{eq:diff}
\end{equation}
The test statistic is the mean per-task difference
\begin{equation}
\bar{d} \;=\; \frac{1}{N}\sum_{i=1}^{N} d_i, \qquad N = |\mathcal{T}|.
\label{eq:tstat}
\end{equation}

\paragraph{Null hypothesis and randomization.} Under
$H_0\!:\!A \stackrel{d}{=} B$ on every task, the within-task labels
$(\hat{p}^A_i,\hat{p}^B_i)$ are exchangeable, which by linearity of
expectation flips the sign of $d_i$ with probability $\tfrac12$
independently for each $i$~\citep{edgington2007randomization}. The
randomization distribution of $\bar{d}$ is therefore
\begin{equation}
\bar{d}^\star(\bm{\varepsilon}) \;=\; \frac{1}{N}\sum_{i=1}^{N} \varepsilon_i\, d_i,
\qquad \bm{\varepsilon}\in\{-1,+1\}^{N}.
\label{eq:flip}
\end{equation}
Tasks with $d_i{=}0$ contribute zero in every flip and are kept,
matching standard sign-flip convention. The two-sided p-value for the
observed $\bar{d}_{\mathrm{obs}}$ is the tail mass
\begin{equation}
p \;=\; \Pr_{\bm{\varepsilon}\sim\mathrm{Unif}\{\pm1\}^N}
\!\Big[\,\big|\bar{d}^\star(\bm{\varepsilon})\big| \;\ge\; |\bar{d}_{\mathrm{obs}}|\,\Big].
\label{eq:pval}
\end{equation}

\paragraph{Computation.} For $N \le 24$ we evaluate
Eq.~\ref{eq:pval} \emph{exactly} by enumerating all $2^N$ sign
assignments. For larger $N$ we use a Monte-Carlo estimator with
$B{=}2{\times}10^5$ uniform random sign vectors and the unbiased
estimator of \citet{phipson2010permutation}:
\begin{equation}
\hat{p} \;=\; \frac{1 \;+\; \#\{\,b : |\bar{d}^\star_b| \ge |\bar{d}_{\mathrm{obs}}|\,\}}{1 \;+\; B}.
\label{eq:phipson}
\end{equation}
The $+1$ on numerator and denominator prevents $\hat{p}{=}0$ for
finite $B$ and guarantees $\hat{p}$ remains a valid p-value (i.e.
$\Pr(\hat{p} \le \alpha \mid H_0) \le \alpha$).

\paragraph{Scope.} Comparisons against \textbf{Full} in
the Main-Results table of the main paper pool tasks across all five benchmark suites
(\textit{Common Sense} 32, \textit{Memory} 30, \textit{Complex
References} 32, \textit{World Knowledge} 32, \textit{Robolab-Vague}
120). Ablation
comparisons in the main-paper Component-Ablation table pool only the four
\textsc{RoboVoLo} suites.

\paragraph{p-values for the main table.} Each row pairs
\textbf{Full} against one column of the main-paper Main-Results table; $\bar{d}_{\mathrm{obs}}$
is the mean per-task success-rate difference (in percentage points)
on the paired tasks.

\begin{center}
\footnotesize
\setlength{\tabcolsep}{6pt}
\renewcommand{\arraystretch}{1.05}
\begin{tabular}{l r r r}
\toprule
\textbf{Full} vs.\ baseline & $\bar{d}_{\mathrm{obs}}$ (pp) & $p$ & mode \\
\midrule
$\pi_{0.5}$               & $+22.18$ & $<10^{-4}$ & MC \\
$\pi_0$-FAST              & $+29.61$ & $<10^{-4}$ & MC \\
MolmoBot                  & $+26.69$ & $<10^{-4}$ & MC \\
MolmoAct2                 & $+32.37$ & $<10^{-4}$ & MC \\
DreamZero                 & $+23.28$ & $<10^{-4}$ & MC \\
CaP-X (single)            & $+24.79$ & $<10^{-4}$ & MC \\
CaP-X (ensemble)          & $+26.31$ & $<10^{-4}$ & MC \\
TiPToP                    & $+21.21$ & $<10^{-4}$ & MC \\
\textsc{VoLoAgent} (No VLA)    & $+19.15$ & $<10^{-4}$ & MC \\
\textsc{VoLoAgent} (Only VLA)  & $\phantom{0}+4.41$ & $0.0599$ & MC \\
\bottomrule
\end{tabular}
\end{center}

\paragraph{p-values for the ablation table.} On the four
\textsc{RoboVoLo} suites:

\begin{center}
\footnotesize
\setlength{\tabcolsep}{6pt}
\renewcommand{\arraystretch}{1.05}
\begin{tabular}{l l r r}
\toprule
Axis & Ablation (vs.\ Full) & $\bar{d}_{\mathrm{obs}}$ (pp) & $p$ \\
\midrule
\multirow{3}{*}{System}
& $\pi_{0.5}$ (Pure VLA)         & $+29.23$ & $<10^{-4}$ \\
& \textsc{VoLoAgent} (No VLA)    & $+24.04$ & $<10^{-4}$ \\
& \textsc{VoLoAgent} (Only VLA)  & $\phantom{0}+6.83$  & $0.0466$ \\
\midrule
\multirow{3}{*}{Perception}
& GDino$+$SAM2 / Molmo2          & $\phantom{0}+3.28$  & $0.3012$ \\
& SAM3 / VLM-point               & $\phantom{0}+5.74$  & $0.0437$ \\
& Exterior camera                & $\phantom{0}+5.28$  & $0.0942$ \\
\midrule
\multirow{3}{*}{VLM model}
& GPT-5.5                        & $\phantom{0}+6.28$  & $0.0480$ \\
& Gemini-2.5-Flash               & $\phantom{0}+9.84$  & $0.0051$ \\
& Qwen3-VL-8B                    & $+21.86$ & $<10^{-4}$ \\
\midrule
\multirow{3}{*}{VLA model}
& $\pi_0$-FAST                   & $+15.57$ & $<10^{-4}$ \\
& MolmoBot-DROID                 & $+16.94$ & $<10^{-4}$ \\
& DreamZero-DROID                & $+19.95$ & $<10^{-4}$ \\
\bottomrule
\end{tabular}
\end{center}

The two non-significant ablation rows are the perception swap
GDino$+$SAM2 / Molmo2 and the exterior-camera variant, matching the
ablation discussion in the main paper.

\subsection{Wilson Score Confidence Intervals}
\label{app:stat-real}

For the real-robot table we report a per-system success rate and a
two-sided $95\%$ confidence interval. With $n{=}42$ trials per system
($14$ tasks $\times\,3$ matched-initial-state trials), the
normal-approximation interval is unreliable for the small success
counts at the tails (e.g., $\pi_{0.5}$ with $\hat{p}\!=\!0.143$
violates the $n\hat{p}(1-\hat{p}) \ge 5$ rule of thumb). We therefore
use the \citet{wilson1927probable} score interval, which inverts the
standard score test on the binomial proportion and remains valid for
small $n$ and proportions near $0$ or $1$.

For $s$ successes out of $n$ trials and a two-sided level $1-\alpha$
($\alpha{=}0.05$, $z_{1-\alpha/2}{=}1.96$), the Wilson interval is
\begin{equation}
\mathrm{CI}_{1-\alpha}(s,n) \;=\;
\frac{1}{1+\tfrac{z^2}{n}} \!\left[\;
  \hat{p} + \tfrac{z^2}{2n}
  \;\pm\; z\sqrt{\,\tfrac{\hat{p}(1-\hat{p})}{n} + \tfrac{z^2}{4n^2}\,}
\;\right],
\qquad \hat{p}=\frac{s}{n}.
\label{eq:wilson}
\end{equation}
This is the interval reported in the main-paper Real-Robot table; it
contains the maximum-likelihood estimate $\hat{p}$, is contained in
$[0,1]$ by construction, and has approximately nominal coverage even
for $\hat{p}$ close to $0$ or $1$~\citep{brown2001interval}.

\paragraph{No paired test on the real-robot ablations.} Because the
intermediate-variant intervals overlap heavily with Full in the
Real-Robot table of the main paper, the main paper reports only
the per-system intervals and does not assert significance for the
within-\textsc{VoLoAgent} ablation comparisons. A larger sample (more tasks and/or more trials
per task) would be required to reach the statistical power needed to
distinguish the three \textsc{VoLoAgent} variants on the real robot.


\section{Real-Robot Setup and Results}
\label{app:real-robot}

\paragraph{Hardware.} A single physical DROID cell~\citep{droid}: a
7-DoF Franka Research 3 arm with a Robotiq 2F-85 parallel-jaw
gripper, an exterior ZED 2i, and a wrist-mounted ZED mini. Camera
extrinsics, lighting, and the joint-position $+$ binary-gripper
action space match the simulation setup of Appendix~\ref{app:sim-setup},
so the same VLA checkpoints, the same orchestrator code, and the same
\textsc{grasp} / \textsc{place} primitives run unchanged from sim to
real (no retraining, no per-cell calibration).

\paragraph{Per-task results.} We sampled $14$ tasks from the four
\textsc{RoboVoLo} suites that are physically reproducible with
the props available in the lab. Initial object placements were arranged
once per task and reset to the same configuration before each of the
$3$ trials per system, so all four systems see identical scenes
(same objects, same poses, same lighting). This gives
$14\times 3 = 42$ trials per system and $4\times 42 = 168$ trials
total. Table~\ref{tab:real-robot-per-task}
lists every task with its source suite and the success counts
behind the main-paper Real-Robot table. 

\begin{table}[h!]
\centering
\caption{Real-robot per-task success counts ($n{=}3$ trials per cell).
Same matched initial states across all four systems. Column totals
match the per-system rates in the main-paper Real-Robot table.}
\label{tab:real-robot-per-task}
\footnotesize
\setlength{\tabcolsep}{4pt}
\renewcommand{\arraystretch}{1.05}
\resizebox{\linewidth}{!}{%
\begin{tabular}{l c c c c}
\toprule
\textbf{Task} & $\pi_{0.5}$ & \textsc{VoLoAgent} (No VLA) & \textsc{VoLoAgent} (Only VLA) & \textbf{\textsc{VoLoAgent} (Full)} \\
\midrule
\texttt{KitCanFruitPair}        & 0/3 & 2/3 & 3/3 & 2/3 \\
\texttt{KitLunchPairs}          & 0/3 & 3/3 & 3/3 & 3/3 \\
\texttt{SortFridgeVsPantry}     & 0/3 & 0/3 & 1/3 & 0/3 \\
\texttt{SortProduceAndDairy}    & 0/3 & 1/3 & 3/3 & 2/3 \\
\texttt{SpatialByStackOrder}    & 0/3 & 0/3 & 0/3 & 0/3 \\
\texttt{SwapBinReplaceFruits}   & 0/3 & 3/3 & 1/3 & 3/3 \\
\texttt{UnstackSelect}          & 2/3 & 3/3 & 3/3 & 3/3 \\
\texttt{XrExceptOrange}         & 1/3 & 2/3 & 2/3 & 3/3 \\
\texttt{XrExtremesToBin}        & 0/3 & 0/3 & 0/3 & 0/3 \\
\texttt{XrFirstLast}            & 1/3 & 1/3 & 0/3 & 1/3 \\
\texttt{XrLeftmostRightmost}    & 0/3 & 2/3 & 1/3 & 1/3 \\
\texttt{XrSecondFourth}         & 2/3 & 1/3 & 0/3 & 0/3 \\
\texttt{XrSplitByBlock}         & 0/3 & 1/3 & 0/3 & 0/3 \\
\texttt{RecycleSort3}           & 0/3 & 0/3 & 0/3 & 0/3 \\
\midrule
\textbf{Total}        & \textbf{6/42} & \textbf{19/42} & \textbf{17/42} & \textbf{18/42} \\
\textbf{Success rate} & 14.3\% & 45.2\% & 40.5\% & 42.9\% \\
\bottomrule
\end{tabular}%
}
\end{table}


\section{Other Simulation Environments}
\label{app:other-sims}

We explored the \textsc{VoLoAgent} stack on several widely-used
MuJoCo-based benchmarks via the same WebSocket proxy: LIBERO~\citep{libero},
RoboCerebra~\citep{robocerebra}, and VLABench~\citep{vlabench}. In
addition, we authored our own set of long-horizon composite tasks on
top of the LIBERO scene set (``LIBERO -- author-designed composite
tasks''; same kitchen / desk scenes as LIBERO but with multi-step
instructions of the same flavor as our \textsc{RoboVoLo} suites:
literal, vague, and creative phrasings of the same scene-level
goal); we also examined the related LIBERO-Plus
suite~\citep{huang2025libero}. We tested the released
$\pi_{0.5}$ checkpoint for corresponding
benchmarks. Two recurring findings emerged
on \emph{every} (benchmark, model) combination we tried; we report
the LIBERO-based results as an example below, and note where the
same pattern recurred on RoboCerebra and VLABench.

\subsection{Finding 1: VLAs over-fit to scene, not generalizable to instruction}
\label{app:other-sims:overfit}

To test whether the VLA actually responds to changes in language,
we held the LIBERO scene configuration fixed and varied only the
prompt: \textit{explicit} (literal ``first $X$ then $Y$''),
\textit{vague} (synonyms), \textit{creative} (intent verbs),
\textit{creative-v2} (scenario utterances), and \textit{no prompt}
(empty). All five prompt styles share identical BDDL goal
predicates; only the language differs. Binary success was $0/10$
in every condition (3-step chains exhaust the $700$-step budget),
so we report \textbf{PSR} (predicate satisfaction rate over $24$
underlying goal predicates).

\begin{table}[h!]
\centering
\caption{$\pi_{0.5}$ (LIBERO) on $10$ author-designed composite
tasks ($24$ BDDL goal predicates), with only the prompt varied.
PSR spans only $5$\,pp from the empty prompt ($30\%$) to the
literal composite ($35\%$).}
\label{tab:other-sims-libero}
\footnotesize
\setlength{\tabcolsep}{10pt}
\renewcommand{\arraystretch}{1.05}
\begin{tabular}{l c}
\toprule
\textbf{Prompt style} & \textbf{$\pi_{0.5}$ (LIBERO) PSR} \\
\midrule
\textit{explicit}     (``first $X$ then $Y$ then $Z$'')   & 35.0\% \\
\textit{vague}        (synonyms)                          & 41.7\% \\
\textit{creative}     (intent verbs / category)           & 35.0\% \\
\textit{creative-v2}  (scenario-level utterances)         & 35.0\% \\
\textit{no prompt}    (empty instruction)                 & 30.0\% \\
\bottomrule
\end{tabular}
\end{table}

The decisive row is \textit{no prompt}: with the language channel
\emph{entirely removed}, the policy still satisfies $30\%$ of the
goal predicates --- essentially the same level it reaches when
given a literal multi-step description of what to do. The four
prompted rows are within a few points of one another (within
single-trial noise), and the empty-prompt row is within that same
band. The reason is not that prompt rewriting has a small but real
ceiling on this suite; it is that this $\pi_{0.5}$ checkpoint has
been trained on too narrow a scene distribution and has \emph{stopped
conditioning on language at all} --- it executes the same
scene-driven trajectory whatever the instruction says, including
when there is no instruction. Until the VLA itself generalizes to
instructions, the orchestrator has nothing to steer through.

We observed the same effect on every other (benchmark, model)
combination we tried and
spot-checks on RoboCerebra~\citep{robocerebra} and
VLABench~\citep{vlabench} reproduce the no-prompt $\approx$ prompted
gap with the released sim-trained checkpoints, showing limited generalist policy.

\subsection{Finding 2: vision tools do not transfer to non-photoreal sims}
\label{app:other-sims:visual}

The \textsc{grasp} / \textsc{place} primitives in
Appendix~\ref{app:system} rely on open-vocabulary detectors
(GroundingDINO, SAM2, SAM3) and pointing models (Molmo2) that were
trained almost exclusively on real-world imagery, and our ultimate
target is real-world deployment. Isaac Sim's PathTracer renderer
produces photometrically realistic images, so these vision tools
transfer. The MuJoCo-based simulators do not: their rendered scenes
have a large appearance gap relative to real scenes, leading to
failures such as
(i) GroundingDINO frequently returning empty boxes or wrong-class
boxes for everyday objects (mug, plate, basket) once they are
rendered in MuJoCo,
(ii) SAM2 / SAM3 either refusing to segment or attaching the mask to
the wrong instance, and
(iii) Molmo2's pointing collapsing to the image center on
flat-shaded scenes.
We observed the same three failure modes on RoboCerebra and VLABench
scenes (both also MuJoCo-rendered) without further tuning, so we
attribute the gap to the renderer rather than to any individual
benchmark's asset choices --- this is the empirical evidence behind
the main paper's ``insufficient realism'' claim.

In contrast, RoboLab uses the DROID setup~\citep{droid}, whose VLA
checkpoints are trained on a broad cross-embodiment real-robot
dataset and remain language-conditional, so the orchestrator's
instruction rewrites actually steer behavior. RoboLab also runs on
Isaac Sim, whose PathTracer rendering closes the visual gap to the
real world enough that GroundingDINO / SAM3 / Molmo2 transfer.
Together, these properties motivated us to design \textsc{RoboVoLo}
on top of RoboLab: we study agentic robot manipulation that is
aligned with a physical robot, rather than chasing failures induced
by sim-only artifacts of the underlying simulator or executor.


\section{Per-suite Component Ablations}
\label{app:ablation-full}

Table~\ref{tab:ablation-full} reports the full per-suite breakdown of the component ablation summarized in the main-paper Component-Ablation table (\emph{Overall} column only) in the main text.

\begin{table*}[h!]
\centering
\caption{Per-suite component ablations. Each axis varies one component while the rest of the system is held at our default. The final \textbf{\textsc{VoLoAgent}} row is the full system reference; its values are constant across all axes. All values are success rate (\%, higher is better).}
\label{tab:ablation-full}
\small
\setlength{\tabcolsep}{8pt}
\renewcommand{\arraystretch}{1.2}
\resizebox{\linewidth}{!}{%
\begin{tabular}{l l c c c c c}
\toprule
Axis & Configuration & \textbf{Common Sense} & \textbf{Memory} & \textbf{Complex Ref.} & \textbf{World Know.} & \textbf{Overall} \\
\midrule
\multirow{3}{*}{System}
& $\pi_{0.5}$ (Pure VLA)        & 11.11 & 13.10 & 16.67 & 9.38 & 12.57 \\
& \textsc{VoLoAgent} (No VLA)          & 32.22 & 13.10 & 9.38 & 16.67 & 17.76 \\
& \textsc{VoLoAgent} (Only VLA)        & 44.44 & 34.52 & 40.62 & 20.83 & 34.97 \\
\midrule
\multirow{3}{*}{Perception}
& GDino+SAM2 / Molmo2                     & 54.44 & 30.95   & 43.75 & 25.00 & 38.52 \\
& SAM3 / VLM-point                        & 43.33 & 35.71   & 43.75 & 21.88 & 36.07 \\
& Exterior camera                         & 41.67 & 30.95   & 52.08 & 22.92 & 36.94 \\
\midrule
\multirow{5}{*}{VLM model}
& Claude Sonnet 4.6                       & 48.89 & 36.90   & 41.67   & 18.75   & 36.34   \\
& GPT-5.5                                 & 42.22 & 35.71   & 42.71   & 21.88   & 35.52   \\
& GPT-5-mini                              & 54.44 & 22.62   & 38.54   & 22.92   & 34.70   \\
& Gemini-2.5-Flash                        & 45.56 & 26.19   & 36.46   & 19.79   & 31.97   \\
& Qwen3-VL-8B                      & 33.33 & 19.05   & 19.79 & 8.33  & 19.95 \\
\midrule
\multirow{4}{*}{VLA model}
& $\pi_{0}$-FAST                          & 50.00 &  25.00  &  19.79  &  11.46  &  26.23  \\
& MolmoBot-DROID                          & 37.78 &  27.38  &  19.79  &  15.62  &  24.86  \\
& MolmoAct2-DROID                         & 30.00 &  10.71   &  3.12  &  7.29  &  12.57  \\
& DreamZero-DROID                   & 50.00 & 10.71 & 13.54 & 13.54 & 21.86 \\
\midrule
\rowcolor{gray!15} \multicolumn{2}{l}{\textbf{\textsc{VoLoAgent}}} & \textbf{54.44} & \textbf{36.90} & \textbf{51.04} & \textbf{25.00} & \textbf{41.80} \\
\bottomrule
\end{tabular}%
}
\end{table*}

\section{Additional Outcome-flow Diagrams}
\label{app:sankey}

\begin{figure}[h!]
\centering
\includegraphics[width=\linewidth]{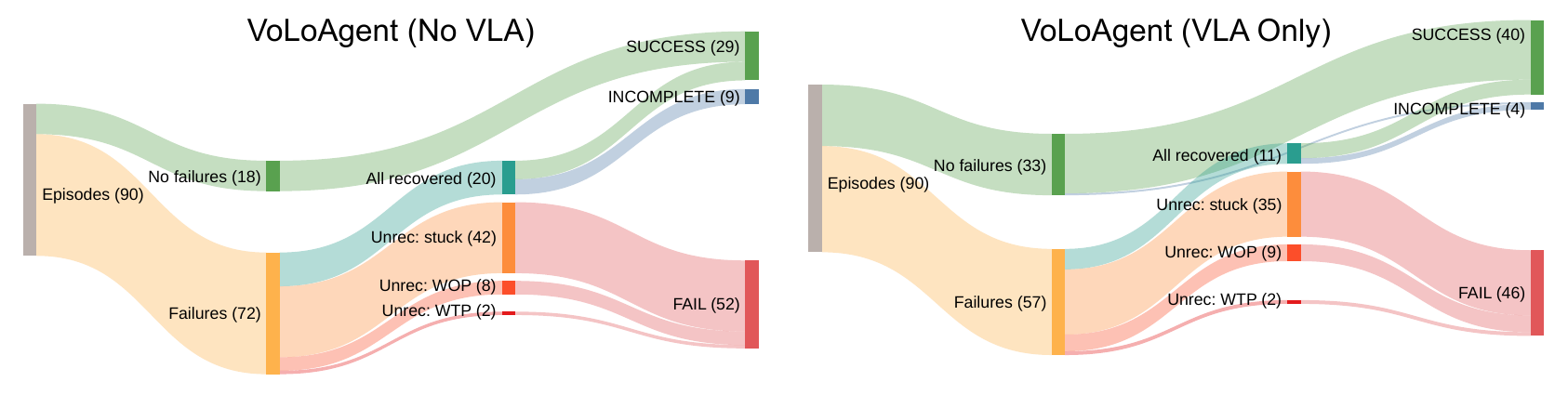}
\caption{Outcome flow on Common Sense ($n{=}90$) for the two intermediate ablations: \textsc{VoLoAgent} (no VLA) on the left and \textsc{VoLoAgent} (Only VLA) on the right.}
\label{fig:sankey-appendix}
\end{figure}

Figure~\ref{fig:sankey-appendix} reports the outcome-flow Sankey diagrams for the two intermediate ablations omitted from the main-paper Sankey figure in the main text: \emph{\textsc{VoLoAgent} (No VLA)}, which replaces the policy with VLM-driven primitives, and \emph{\textsc{VoLoAgent} (Only VLA)}, which keeps the VLA + VLM monitor but disables tool-augmented recovery.

\section{Failure-mode Taxonomy and Definitions}
\label{app:metrics-defs}

This section expands the two diagnostic streams referenced in
the main paper. Both run \emph{passively}: events are
written to per-episode log files but never influence orchestration. They
share the same simulator's \texttt{gt\_state} export.

\subsection{World failures (outcome events, \texttt{task\_failures.jsonl})}
\label{app:metrics1}

\paragraph{Setup.} Each long-horizon task is decomposed into an
ordered sequence of sub-tasks $S_1 \to S_2 \to \cdots \to S_K$,
where each $S_k$ carries a set of target objects with their
required end-states:
\[
S_k \;=\; \big\{ (o^{(k)}_1,\, \tau^{(k)}_1),\; \ldots,\; (o^{(k)}_{N_k},\, \tau^{(k)}_{N_k}) \big\},
\]
with $o^{(k)}_j$ the $j$-th tracked object in sub-task $k$ and
$\tau^{(k)}_j$ its required predicate
(e.g.\ \texttt{in\_container(bowl)}, \texttt{on\_surface(plate)}).
At any instant exactly one sub-task is \emph{active}; all of its
$N_k$ objects are tracked in parallel. The active sub-task advances
to $S_{k+1}$ once every $(o^{(k)}_j,\, \tau^{(k)}_j)$ predicate is
satisfied. Objects whose target was satisfied in any earlier
sub-task remain tracked for regression --- with one exception: if
an object $o$ also appears as a target of the \emph{current}
sub-task, regression is not fired against $o$ until the current
sub-task completes (an in-flight re-grasp of $o$ is part of the
plan, not a failure).

\paragraph{Events.} Stateless rules consume per-step
\texttt{gt\_state} snapshots and emit five event types against this
sub-task / object-tracking state:

\begin{itemize}\setlength{\itemsep}{1pt}
\item \texttt{wrong\_object\_picked} (\emph{WOP}): the gripper holds an object that is not a target of the currently-active sub-task.
\item \texttt{wrong\_target\_place} (\emph{WTP}): a target object of the active sub-task has been released into a stable pose that does not satisfy its required predicate.
\item \texttt{object\_regression}: an object that previously emitted \texttt{object\_complete} (target satisfied in an earlier sub-task) has stopped satisfying its target predicate, and is not a target of the active sub-task.
\item \texttt{stuck}: no end-effector progress and no new sub-task completion for $\sim$10\,s. Re-fires periodically while still stuck.
\item \texttt{recovery}: a previously-fired failure's underlying condition has resolved. Each recovery is paired one-to-one with its failure.
\end{itemize}

A \emph{failure spell} is the interval between a failure event and its
paired recovery. An episode contains \emph{unrecovered} spells when at
least one failure event lacks a paired recovery before episode end.
Successful episodes are post-hoc resolved as ``all recovered'' (success
implies every spell ended favorably even if the recovery event was cut
off by the success terminal).

\subsection{VLM failures (\texttt{metrics.jsonl})}
\label{app:metrics2}

\paragraph{How GT is obtained.} Each \textsc{RoboVoLo} task carries
two hand-authored predicate lists in its task class:
\texttt{gt\_success\_checks} (the conjunction of object-state
predicates that defines task success) and
\texttt{gt\_invariant\_checks} (predicates that must remain true
throughout, e.g.\ items already sorted into a container should stay
there). A typical entry is
\texttt{\{predicate: object\_in\_container, object: [lime01,
orange\_01, lemon\_02], container: serving\_bowl,
logical: all\}} drawn directly from the simulator's physics
state. At every step, robolab evaluates these predicates against
the current scene and exports the results inside
\texttt{obs["gt\_state"]} (alongside object poses, gripper contact,
and per-subtask completion flags). The orchestrator's GT-metric
detectors~\citep{robolab} consume this stream passively --- they
emit metric events but never alter prompts, observations, or
recovery behavior, so the headline success rates in
the main-paper Main-Results table are unaffected by whether
\texttt{--enable-gt-metrics} is on.

\paragraph{Pairing VLM calls with GT.} Each VLM call (planning,
completion check, failure check, grasp-tool dispatch) is paired with
the GT predicate it should have verified at the same step: e.g.\ a
``subgoal complete'' VLM verdict is matched against the
\texttt{gt\_success\_checks} predicate for that subgoal, and a
grasp-tool target name is matched against the active sub-task's
target object list. The eight leaf metrics in
Table~\ref{tab:vlm_failures_breakdown} partition into four taxonomy
groups, and the group sum equals the total VLM-failure count (no
overlap, no omission):

\begin{itemize}\setlength{\itemsep}{1pt}
\item \textbf{Planning} — \texttt{vlm\_plan\_mismatch}: the VLM's proposed sub-goal decomposition does not align with the GT plan.
\item \textbf{Completion monitor} — \texttt{vlm\_scene\_qa\_failure} (\emph{false\_complete}: scene-QA says complete while GT incomplete; \emph{missed\_complete}: scene-QA says incomplete after GT satisfied), \texttt{vlm\_completion\_mismatch} (orchestrator advanced subgoal on the VLM signal but GT is still incomplete), \texttt{vlm\_task\_success\_qa\_failure} (end-of-episode QA claimed success while GT failed).
\item \textbf{Failure monitor} — \texttt{vlm\_failure\_missed} (a blocking GT failure was not flagged), \texttt{vlm\_invariant\_qa\_failure} (a GT invariant violation, e.g.\ \texttt{object\_in\_container} contradiction, was not flagged).
\item \textbf{Tool-use} — \texttt{vlm\_grasp\_target\_mismatch}: the grasp-tool target name produced by the VLM does not resolve to any current GT target after the two-stage resolver below.
\end{itemize}

\paragraph{Two-stage VLM-as-judge resolver for tool-use mismatches.}
Naming style (``white bottle with green cap'' vs.\ canonical
\texttt{ranch\_dressing}) inflates the raw \texttt{vlm\_grasp\_target\_mismatch}
count. A two-stage offline resolver, applied once per sweep, reclassifies
alias-equivalent events to a success bucket so the headline
\emph{tool-use error} number reflects true semantic mismatches:

\begin{itemize}\setlength{\itemsep}{1pt}
\item \emph{Stage 1 — string + alias dictionary.} Exact-match against the
canonical GT target name, then alias-set lookup against a hand-maintained
per-object phrase dictionary (e.g.\ \{``ranch dressing'', ``white bottle
with green cap'', ``green-capped bottle'', \ldots\} $\rightarrow$
\texttt{ranch\_dressing}). Deterministic, sub-ms per event.
\item \emph{Stage 2 — VLM grounding.} For residual mismatches, a VLM is
prompted with the scene image and the current GT target list and asked
which canonical target the VLM-uttered phrase refers to (or ``none of
them''). Verdicts are cached on disk in \texttt{metrics\_resolved.jsonl}
sidecars so re-runs cost nothing.
\end{itemize}
Resolved events are reclassified to a \emph{vlm\_grasp\_target\_qa:
target\_aliased} success bucket and excluded from
\texttt{vlm\_tool\_use\_error}. All counts in
Table~\ref{tab:vlm_failures_breakdown} and
the main-paper VLM-failure figure are post-resolver.

\input{corl_2026_template_submission/appendix_vlm_failures_table}

%% file: corl_2026_template_submission/appendix_vlm_failures_table.tex
%

\begin{table}[h!]
\centering
\caption{Per-leaf-metric VLM-failure breakdown on LH-CS (System-2 audit).
Counts are raw event totals across all 90 LH-CS episodes per VLM. Group
sums (\textit{Planning} / \textit{Completion} / \textit{Failure-mon.} /
\textit{Tool-use}) correspond to the four bar-segment colors in
the main-paper VLM-failure figure.}
\label{tab:vlm_failures_breakdown}
\footnotesize
\setlength{\tabcolsep}{3pt}
\renewcommand{\arraystretch}{1.12}
\resizebox{\linewidth}{!}{%
\begin{tabular}{@{}p{1.4cm} p{4.5cm} p{4.8cm} r r r r@{}}
\toprule
\textbf{Group} & \textbf{Metric} & \textbf{Definition} &
\textbf{Claude} & \textbf{GPT} & \textbf{Gemini} & \textbf{Qwen} \\
& & & \textbf{Opus 4.6} & \textbf{5.5} & \textbf{2.5 Flash} & \textbf{3-VL-8B} \\
\midrule
Planning &
\texttt{vlm\_plan\_mismatch} &
\textit{plan\_mismatch}: VLM-proposed subgoal decomposition does not align with the GT plan. &
4 & 9 & 2 & 8 \\
\addlinespace[1pt]
\multicolumn{3}{r}{\textit{Planning subtotal}} & \textbf{4} & \textbf{9} & \textbf{2} & \textbf{8} \\
\midrule
Completion monitor &
\texttt{vlm\_scene\_qa\_failure} &
\textit{false\_complete}: scene-QA reports complete while GT predicate still unsatisfied. &
22 & 66 & 61 & 136 \\
&
\texttt{vlm\_scene\_qa\_failure} &
\textit{missed\_complete}: scene-QA reports incomplete after GT predicate already satisfied. &
35 & 23 & 21 & 41 \\
&
\texttt{vlm\_completion\_mismatch} &
Orchestrator advanced subgoal on the VLM signal, but GT marks current subgoal still incomplete. &
23 & 68 & 61 & 146 \\
&
\texttt{vlm\_task\_success\_qa\_failure} &
End-of-episode task-success QA says success while GT \texttt{episode\_results} says failure. &
4 & 27 & 33 & 36 \\
\addlinespace[1pt]
\multicolumn{3}{r}{\textit{Completion subtotal}} & \textbf{84} & \textbf{184} & \textbf{176} & \textbf{359} \\
\midrule
Failure monitor &
\texttt{vlm\_failure\_missed} &
GT signaled a blocking failure (dropped target, stuck gripper, \ldots) that the VLM failure-monitor did not flag. &
1 & 12 & 35 & 43 \\
&
\texttt{vlm\_invariant\_qa\_failure} &
GT detected an invariant violation (e.g.\ wrong-object-in-container) that the VLM invariant-QA did not flag. &
1 & 32 & 46 & 46 \\
\addlinespace[1pt]
\multicolumn{3}{r}{\textit{Failure-mon. subtotal}} & \textbf{2} & \textbf{44} & \textbf{81} & \textbf{89} \\
\midrule
Tool-use &
\texttt{vlm\_grasp\_target\_mismatch} &
\textit{target\_mismatch}: grasp-tool target name does not resolve to any current GT target after alias resolution. &
12 & 4 & 3 & 0 \\
&
\texttt{vlm\_grasp\_target\_mismatch} &
\textit{missing\_target}: grasp-tool call emitted with no target name. &
0 & 0 & 1 & 0 \\
\addlinespace[1pt]
\multicolumn{3}{r}{\textit{Tool-use subtotal}} & \textbf{12} & \textbf{4} & \textbf{4} & \textbf{0} \\
\midrule
\multicolumn{3}{r}{\textbf{Total VLM-failure events}} &
\textbf{102} & \textbf{241} & \textbf{263} & \textbf{456} \\
\bottomrule
\end{tabular}}
\end{table}

%% file: main.bib
@inproceedings{rt2,
  title     = {{RT-2}: Vision-Language-Action Models Transfer Web Knowledge to Robotic Control},
  author    = {Brohan, Anthony and Brown, Noah and Carbajal, Justice and
               Chebotar, Yevgen and Chen, Xi and Choromanski, Krzysztof and
               Ding, Tianli and Driess, Danny and Dubey, Avinava and Finn, Chelsea and
               Florence, Pete and Fu, Chuyuan and Gonzalez Arenas, Montse and
               Gopalakrishnan, Keerthana and Han, Kehang and Hausman, Karol and
               Herzog, Alexander and Hsu, Jasmine and Ichter, Brian and Irpan, Alex and
               Joshi, Nikhil and Julian, Ryan and Kalashnikov, Dmitry and
               Kuang, Yuheng and Leal, Isabel and Lee, Lisa and Lee, Tsang-Wei Edward and
               Levine, Sergey and Lu, Yao and Michalewski, Henryk and Mordatch, Igor and
               Pertsch, Karl and Rao, Kanishka and Reymann, Krista and Ryoo, Michael and
               Salazar, Grecia and Sanketi, Pannag and Sermanet, Pierre and Singh, Jaspiar and
               Singh, Anikait and Soricut, Radu and Tran, Huong and Vanhoucke, Vincent and
               Vuong, Quan and Wahid, Ayzaan and Welker, Stefan and Wohlhart, Paul and
               Wu, Jialin and Xia, Fei and Xiao, Ted and Xu, Peng and Xu, Sichun and
               Yu, Tianhe and Zitkovich, Brianna},
  booktitle = {CoRL},
  year      = {2023}
}

@inproceedings{vima,
  title     = {{VIMA}: General Robot Manipulation with Multimodal Prompts},
  author    = {Jiang, Yunfan and Gupta, Agrim and Zhang, Zichen and Wang, Guanzhi and
               Dou, Yongqiang and Chen, Yanjun and Fei-Fei, Li and Anandkumar, Anima and
               Zhu, Yuke and Fan, Linxi},
  booktitle = {ICML},
  year      = {2023}
}

@inproceedings{rdt1b,
  title     = {{RDT-1B}: a Diffusion Foundation Model for Bimanual Manipulation},
  author    = {Liu, Songming and Wu, Lingxuan and Li, Bangguo and Tan, Hengkai and
               Chen, Huayu and Wang, Zhengyi and Xu, Ke and Su, Hang and Zhu, Jun},
  booktitle = {ICLR},
  year      = {2025}
}

@article{openvla,
  title   = {{OpenVLA}: An Open-Source Vision-Language-Action Model},
  author  = {Kim, Moo Jin and Pertsch, Karl and Karamcheti, Siddharth and Xiao, Ted and
             Balakrishna, Ashwin and Nair, Suraj and Rafailov, Rafael and Foster, Ethan and
             Lam, Grace and Sanketi, Pannag and Vuong, Quan and Kollar, Thomas and
             Burchfiel, Benjamin and Tedrake, Russ and Sadigh, Dorsa and Levine, Sergey and
             Liang, Percy and Finn, Chelsea},
  journal = {arXiv preprint arXiv:2406.09246},
  year    = {2024}
}

@article{pi05,
  title   = {{$\pi_{0.5}$}: a Vision-Language-Action Model with Open-World Generalization},
  author={Intelligence, Physical and Black, Kevin and Brown, Noah and Darpinian, James and Dhabalia, Karan and Driess, Danny and Esmail, Adnan and Equi, Michael and Finn, Chelsea and Fusai, Niccolo and others},  
  journal = {arXiv preprint},
  year    = {2025}
}

@article{pistar06,
  title   = {{$\pi^{*}_{0.6}$}: a {VLA} That Learns From Experience},
  author={Intelligence, Physical and Amin, Ali and Aniceto, Raichelle and Balakrishna, Ashwin and Black, Kevin and Conley, Ken and Connors, Grace and Darpinian, James and Dhabalia, Karan and DiCarlo, Jared and others},
  journal = {arXiv preprint arXiv:2511.14759},
  year    = {2025}
}

@article{pi07,
  title   = {{$\pi_{0.7}$}: a Steerable Generalist Robotic Foundation Model with Emergent Capabilities},
  author={Intelligence, Physical and Ai, Bo and Amin, Ali and Aniceto, Raichelle and Balakrishna, Ashwin and Balke, Greg and Black, Kevin and Bokinsky, George and Cao, Shihao and Charbonnier, Thomas and others},
  journal={arXiv preprint arXiv:2604.15483},
  year    = {2026}
}

@article{molmoact,
  title   = {{MolmoAct}: Action Reasoning Models that can Reason in Space},
  author  = {Lee, Jason and Duan, Jiafei and Fang, Haoquan and Deng, Yuquan and
             Liu, Shuo and Li, Boyang and Fang, Bohan and Zhang, Jieyu and
             Wang, Yi Ru and Lee, Sangho and Han, Winson and Pumacay, Wilbert and
             Wu, Angelica and Hendrix, Rose and Farley, Karen and VanderBilt, Eli and
             Farhadi, Ali and Fox, Dieter and Krishna, Ranjay},
  journal = {arXiv preprint arXiv:2508.07917},
  year    = {2025}
}

@article{molmoact2,
  title   = {{MolmoAct2}: Action Reasoning Models for Real-world Deployment},
  author  = {Fang, Haoquan and Duan, Jiafei and Clay, Donovan and Wang, Sam and
             Liu, Shuo and Huang, Weikai and Fan, Xiang and Tsai, Wei-Chuan and
             Chen, Shirui and Wang, Yi Ru and Xing, Shanli and Cho, Jaemin and
             Park, Jae Sung and Eftekhar, Ainaz and Sushko, Peter and Farley, Karen and
             Wadhwa, Angad and Harrison, Cole and Han, Winson and Lee, Ying-Chun and
             VanderBilt, Eli and Hendrix, Rose and Ellawela, Suveen and Ngoo, Lucas and
             Chai, Joyce and Ren, Zhongzheng and Farhadi, Ali and Fox, Dieter and
             Krishna, Ranjay},
  journal = {arXiv preprint arXiv:2605.02881},
  year    = {2026}
}

@article{molmobot,
  title   = {{MolmoBot}: Large-Scale Simulation Enables Zero-Shot Manipulation},
  author  = {Deshpande, Abhay and Guru, Maya and Hendrix, Rose and Jauhri, Snehal and
             Eftekhar, Ainaz and Tripathi, Rohun and Argus, Max and Salvador, Jordi and
             Fang, Haoquan and Wallingford, Matthew and Pumacay, Wilbert and
             Kim, Yejin and Pfeifer, Quinn and Lee, Ying-Chun and Wolters, Piper and
             Rayyan, Omar and Zhang, Mingtong and Duan, Jiafei and Farley, Karen and
             Han, Winson and VanderBilt, Eli and Fox, Dieter and Farhadi, Ali and
             Chalvatzaki, Georgia and Shah, Dhruv and Krishna, Ranjay},
  journal = {arXiv preprint arXiv:2603.16861},
  year    = {2026}
}

@article{gr2,
  title   = {{GR-2}: A Generative Video-Language-Action Model with Web-Scale Knowledge for Robot Manipulation},
  author  = {Cheang, Chi-Lam and Chen, Guangzeng and Jing, Ya and Kong, Tao and
             Li, Hang and Li, Yifeng and Liu, Yuxiao and Wu, Hongtao and
             Xu, Jiafeng and Yang, Yichu and Zhang, Hanbo and Zhu, Minzhao},
  journal = {arXiv preprint arXiv:2410.06158},
  year    = {2024}
}

@article{dreamzero,
  title   = {World Action Models are Zero-shot Policies},
  author  = {Ye, Seonghyeon and Ge, Yunhao and Zheng, Kaiyuan and Gao, Shenyuan and
             Yu, Sihyun and Kurian, George and Indupuru, Suneel and Tan, You Liang and
             Zhu, Chuning and Xiang, Jiannan and Malik, Ayaan and Lee, Kyungmin and
             Liang, William and Ranawaka, Nadun and Gu, Jiasheng and Xu, Yinzhen and
             Wang, Guanzhi and Hu, Fengyuan and Narayan, Avnish and Bjorck, Johan and
             Wang, Jing and Kim, Gwanghyun and Niu, Dantong and Zheng, Ruijie and
             Xie, Yuqi and Wu, Jimmy and Wang, Qi and Julian, Ryan and Xu, Danfei and
             Du, Yilun and Chebotar, Yevgen and Reed, Scott and Kautz, Jan and
             Zhu, Yuke and Fan, Linxi and Jang, Joel},
  journal = {arXiv preprint arXiv:2602.15922},
  year    = {2026}
}

@article{dreamdojo,
  title   = {{DreamDojo}: A Generalist Robot World Model from Large-Scale Human Videos},
  author  = {Gao, Shenyuan and Liang, William and Zheng, Kaiyuan and Malik, Ayaan and
             Ye, Seonghyeon and Yu, Sihyun and Tseng, Wei-Cheng and Dong, Yuzhu and
             Mo, Kaichun and Lin, Chen-Hsuan and Ma, Qianli and Nah, Seungjun and
             Magne, Loic and Xiang, Jiannan and Xie, Yuqi and Zheng, Ruijie and
             Niu, Dantong and Tan, You Liang and Zentner, K. R. and Kurian, George and
             Indupuru, Suneel and Jannaty, Pooya and Gu, Jinwei and Zhang, Jun and
             Malik, Jitendra and Abbeel, Pieter and Liu, Ming-Yu and Zhu, Yuke and
             Jang, Joel and Fan, Linxi},
  journal = {arXiv preprint arXiv:2602.06949},
  year    = {2026}
}

@article{cosmospolicy,
  title   = {{Cosmos Policy}: Fine-Tuning Video Models for Visuomotor Control and Planning},
  author  = {Kim, Moo Jin and Gao, Yihuai and Lin, Tsung-Yi and Lin, Yen-Chen and
             Ge, Yunhao and Lam, Grace and Liang, Percy and Song, Shuran and
             Liu, Ming-Yu and Finn, Chelsea and Gu, Jinwei},
  journal = {arXiv preprint arXiv:2601.16163},
  year    = {2026}
}

@article{lingbotva,
  title   = {Causal World Modeling for Robot Control},
  author  = {Li, Lin and Zhang, Qihang and Luo, Yiming and Yang, Shuai and
             Wang, Ruilin and Han, Fei and Yu, Mingrui and Gao, Zelin and
             Xue, Nan and Zhu, Xing and Shen, Yujun and Xu, Yinghao},
  journal = {arXiv preprint arXiv:2601.21998},
  year    = {2026}
}

@inproceedings{codeaspolicy,
  title     = {Code as Policies: Language Model Programs for Embodied Control},
  author    = {Liang, Jacky and Huang, Wenlong and Xia, Fei and Xu, Peng and
               Hausman, Karol and Ichter, Brian and Florence, Pete and Zeng, Andy},
  booktitle = {ICRA},
  year      = {2023}
}

@inproceedings{progprompt,
  title     = {{ProgPrompt}: Generating Situated Robot Task Plans using Large Language Models},
  author    = {Singh, Ishika and Blukis, Valts and Mousavian, Arsalan and Goyal, Ankit and
               Xu, Danfei and Tremblay, Jonathan and Fox, Dieter and Thomason, Jesse and
               Garg, Animesh},
  booktitle = {ICRA},
  year      = {2023}
}

@article{capx,
  title   = {{CaP-X}: A Framework for Benchmarking and Improving Coding Agents for Robot Manipulation},
  author={Fu, Max and Yu, Justin and El-Refai, Karim and Kou, Ethan and Xue, Haoru and Huang, Huang and Xiao, Wenli and Wang, Guanzhi and Li, Fei-Fei and Shi, Guanya and others},
  journal = {arXiv preprint arXiv:2603.22435},
  year    = {2026}
}

@inproceedings{comerobot,
  title     = {Closed-Loop Open-Vocabulary Mobile Manipulation with {GPT-4V}},
  author    = {Zhi, Peiyuan and Zhang, Zhiyuan and Zhao, Yu and Han, Muzhi and
               Zhang, Zeyu and Li, Zhitian and Jiao, Ziyuan and Jia, Baoxiong and
               Huang, Siyuan},
  booktitle = {ICRA},
  year      = {2025}
}

@misc{hamster,
  title         = {{HAMSTER}: Hierarchical Action Models For Open-World Robot Manipulation},
  author        = {Li, Yi and Deng, Yuquan and Zhang, Jesse and Jang, Joel and
                   Memmel, Marius and Yu, Raymond and Garrett, Caelan Reed and
                   Ramos, Fabio and Fox, Dieter and Li, Anqi and Gupta, Abhishek and
                   Goyal, Ankit},
  year          = {2025},
  eprint        = {2502.05485},
  archivePrefix = {arXiv},
  primaryClass  = {cs.RO}
}

@article{hirobot,
  title   = {Hi Robot: Open-Ended Instruction Following with Hierarchical Vision-Language-Action Models},
  author  = {Shi, Lucy Xiaoyang and Ichter, Brian and Equi, Michael and Ke, Liyiming and
             Pertsch, Karl and Vuong, Quan and Tanner, James and Walling, Anna and
             Wang, Haohuan and Fusai, Niccolo and Li-Bell, Adrian and Driess, Danny and
             Groom, Lachy and Levine, Sergey and Finn, Chelsea},
  journal = {arXiv preprint arXiv:2502.19417},
  year    = {2025}
}

@inproceedings{hirt,
  title     = {{HiRT}: Enhancing Robotic Control with Hierarchical Robot Transformers},
  author    = {Zhang, Jianke and Guo, Yanjiang and Chen, Xiaoyu and Wang, Yen-Jen and
               Hu, Yucheng and Shi, Chengming and Chen, Jianyu},
  booktitle = {CoRL},
  year      = {2024}
}

@inproceedings{manipulateanything,
  title     = {Manipulate-Anything: Automating Real-World Robots using Vision-Language Models},
  author    = {Duan, Jiafei and Yuan, Wentao and Pumacay, Wilbert and Wang, Yi Ru and
               Ehsani, Kiana and Fox, Dieter and Krishna, Ranjay},
  booktitle = {CoRL},
  year      = {2024}
}

@article{agenticrobot,
  title   = {Agentic Robot: A Brain-Inspired Framework for Vision-Language-Action Models in Embodied Agents},
  author  = {Yang, Zhejian and Chen, Yongchao and Zhou, Xueyang and Yan, Jiangyue and
             Song, Dingjie and Liu, Yinuo and Li, Yuting and Zhang, Yu and
             Zhou, Pan and Chen, Hechang and Sun, Lichao},
  journal = {arXiv preprint arXiv:2505.23450},
  year    = {2025}
}

@article{goal2skill,
  title   = {{Goal2Skill}: Long-Horizon Manipulation with Adaptive Planning and Reflection},
  author  = {Liu, Zhen and Ning, Xinyu and Hu, Zhe and Xie, Xinxin and Li, Weize and
             Tang, Zhipeng and Wang, Chongyu and Yang, Zejun and Wang, Hanlin and
             Liu, Yitong and Pu, Zhongzhu},
  journal = {arXiv preprint arXiv:2604.13942},
  year    = {2026}
}

@article{failsafe,
  title   = {{FailSafe}: Reasoning and Recovery from Failures in Vision-Language-Action Models},
  author  = {Lin, Zijun and Duan, Jiafei and Fang, Haoquan and Fox, Dieter and
             Krishna, Ranjay and Tan, Cheston and Wen, Bihan},
  journal = {arXiv preprint arXiv:2510.01642},
  year    = {2025}
}

@article{criticloop,
  title   = {Critic in the Loop: A Tri-System {VLA} Framework for Robust Long-Horizon Manipulation},
  author  = {Yi, Pengfei and Ma, Yingjie and Xu, Wenjiang and Hao, Yanan and
             Gan, Shuai and Li, Wanting and Zhong, Shanlin},
  journal = {arXiv preprint arXiv:2603.05185},
  year    = {2026}
}

@inproceedings{aha,
  title     = {{AHA}: A Vision-Language-Model for Detecting and Reasoning Over Failures in Robotic Manipulation},
  author    = {Duan, Jiafei and Pumacay, Wilbert and Kumar, Nishanth and Wang, Yi Ru and
               Tian, Shulin and Yuan, Wentao and Krishna, Ranjay and Fox, Dieter and
               Mandlekar, Ajay and Guo, Yijie},
  booktitle = {ICLR},
  year      = {2025}
}

@article{robofac,
  title   = {{RoboFAC}: A Comprehensive Framework for Robotic Failure Analysis and Correction},
  author  = {Ye, Zewei and Lu, Weifeng and Ye, Minghao and Lin, Tao and
             Yang, Shuo and Yan, Junchi and Zhao, Bo},
  journal = {arXiv preprint arXiv:2505.12224},
  year    = {2025}
}

@inproceedings{innermonologue,
  title     = {Inner Monologue: Embodied Reasoning through Planning with Language Models},
  author    = {Huang, Wenlong and Xia, Fei and Xiao, Ted and Chan, Harris and
               Liang, Jacky and Florence, Pete and Zeng, Andy and Tompson, Jonathan and
               Mordatch, Igor and Chebotar, Yevgen and Sermanet, Pierre and
               Brown, Noah and Jackson, Tomas and Luu, Linda and Levine, Sergey and
               Hausman, Karol and Ichter, Brian},
  booktitle = {CoRL},
  year      = {2022}
}

@article{safevla,
  title   = {{SAFE}: Multitask Failure Detection for Vision-Language-Action Models},
  author  = {Gu, Qiao and Ju, Yuanliang and Sun, Shengxiang and Gilitschenski, Igor and
             Nishimura, Haruki and Itkina, Masha and Shkurti, Florian},
  journal = {arXiv preprint arXiv:2506.09937},
  year    = {2025}
}

@article{racer,
  title   = {{RACER}: Rich Language-Guided Failure Recovery Policies for Imitation Learning},
  author  = {Dai, Yinpei and Lee, Jayjun and Fazeli, Nima and Chai, Joyce},
  journal = {arXiv preprint arXiv:2409.14674},
  year    = {2024}
}

@article{replanvlm,
  title   = {{ReplanVLM}: Replanning Robotic Tasks with Visual Language Models},
  author  = {Mei, Aoran and Zhu, Guo-Niu and Zhang, Huaxiang and Gan, Zhongxue},
  journal = {arXiv preprint arXiv:2407.21762},
  year    = {2024}
}

@article{replan,
  title   = {{RePLan}: Robotic Replanning with Perception and Language Models},
  author  = {Skreta, Marta and Zhou, Zihan and Yuan, Jia Lin and Darvish, Kourosh and
             Aspuru-Guzik, Al{\'a}n and Garg, Animesh},
  journal = {arXiv preprint arXiv:2401.04157},
  year    = {2024}
}

@article{lera,
  title   = {{LERa}: Replanning with Visual Feedback in Instruction Following},
  author  = {Pchelintsev, Svyatoslav and Patratskiy, Maxim and Onishchenko, Anatoly and
             Korchemnyi, Alexandr and Medvedev, Aleksandr and Vinogradova, Uliana and
             Galuzinsky, Ilya and Postnikov, Aleksey and Kovalev, Alexey K. and
             Panov, Aleksandr I.},
  journal = {arXiv preprint arXiv:2507.05135},
  year    = {2025}
}

@article{novaplan,
  title   = {{NovaPlan}: Zero-Shot Long-Horizon Manipulation via Closed-Loop Video Language Planning},
  author  = {Fu, Jiahui and Nan, Junyu and Sun, Lingfeng and Li, Hongyu and
             Qian, Jianing and Barry, Jennifer L. and Kitani, Kris and
             Konidaris, George},
  journal = {arXiv preprint arXiv:2602.20119},
  year    = {2026}
}

@article{tiptop,
  title   = {{TiPToP}: A Modular Open-Vocabulary Planning System for Robotic Manipulation},
  author  = {Shen, William and Kumar, Nishanth and Chintalapudi, Sahit and
             Wang, Jie and Watson, Christopher and Hu, Edward and Cao, Jing and
             Jayaraman, Dinesh and Kaelbling, Leslie Pack and Lozano-P{\'e}rez, Tom{\'a}s},
  journal = {arXiv preprint arXiv:2603.09971},
  year    = {2026}
}

@article{vlmtamp,
  title   = {Guiding Long-Horizon Task and Motion Planning with Vision Language Models},
  author  = {Yang, Zhutian and Garrett, Caelan and Fox, Dieter and
             Lozano-P{\'e}rez, Tom{\'a}s and Kaelbling, Leslie Pack},
  journal = {arXiv preprint arXiv:2410.02193},
  year    = {2024}
}

@article{spacetools,
  title   = {{SpaceTools}: Tool-Augmented Spatial Reasoning via Double Interactive RL},
  author={Chen, Siyi and Uy, Mikaela Angelina and Song, Chan Hee and Ladhak, Faisal and Murali, Adithyavairavan and Qu, Qing and Birchfield, Stan and Blukis, Valts and Tremblay, Jonathan},
  journal = {CVPR},
  year    = {2026}
}

@article{lei2026longhorizon,
  title   = {{Towards Long-horizon Embodied Agents with Tool-Aligned Vision-Language-Action Models}},
  author  = {Lei, Zixing and Liu, Changxing and Xiong, Yichen and Xiong, Minhao and
             Ding, Yuanzhuo and Zhang, Zhipeng and Li, Weixin and Chen, Siheng},
  journal = {arXiv preprint arXiv:2605.13119},
  year    = {2026}
}

@inproceedings{groundingdino,
  title     = {Grounding {DINO}: Marrying {DINO} with Grounded Pre-Training for Open-Set Object Detection},
  author    = {Liu, Shilong and Zeng, Zhaoyang and Ren, Tianhe and Li, Feng and
               Zhang, Hao and Yang, Jie and Jiang, Qing and Li, Chunyuan and
               Yang, Jianwei and Su, Hang and Zhu, Jun and Zhang, Lei},
  booktitle = {ECCV},
  year      = {2024}
}

@article{sam2,
  title   = {{SAM} 2: Segment Anything in Images and Videos},
  author  = {Ravi, Nikhila and Gabeur, Valentin and Hu, Yuan-Ting and Hu, Ronghang and
             Ryali, Chaitanya and Ma, Tengyu and Khedr, Haitham and R{\"a}dle, Roman and
             Rolland, Chloe and Gustafson, Laura and Mintun, Eric and Pan, Junting and
             Alwala, Kalyan Vasudev and Carion, Nicolas and Wu, Chao-Yuan and
             Girshick, Ross and Doll{\'a}r, Piotr and Feichtenhofer, Christoph},
  journal = {arXiv preprint arXiv:2408.00714},
  year    = {2024}
}

@article{sam3,
  title   = {{SAM} 3: Segment Anything with Concepts},
  author  = {Carion, Nicolas and Gustafson, Laura and Hu, Yuan-Ting and Debnath, Shoubhik and
             Hu, Ronghang and Suris, Didac and Ryali, Chaitanya and Alwala, Kalyan Vasudev and
             Khedr, Haitham and Huang, Andrew and Lei, Jie and Ma, Tengyu and
             Guo, Baishan and Kalla, Arpit and Marks, Markus and Greer, Joseph and
             Wang, Meng and Sun, Peize and R{\"a}dle, Roman and Afouras, Triantafyllos and
             Mavroudi, Effrosyni and Xu, Katherine and Wu, Tsung-Han and Zhou, Yu and
             Momeni, Liliane and Hazra, Rishi and Ding, Shuangrui and Vaze, Sagar and
             Porcher, Francois and Li, Feng and Li, Siyuan and Kamath, Aishwarya and
             Cheng, Ho Kei and Doll{\'a}r, Piotr and Ravi, Nikhila and Saenko, Kate and
             Zhang, Pengchuan and Feichtenhofer, Christoph},
  journal = {arXiv preprint arXiv:2511.16719},
  year    = {2025}
}

@article{molmo2,
  title   = {{Molmo2}: Open Weights and Data for Vision-Language Models with Video Understanding and Grounding},
  author  = {Clark, Christopher and Zhang, Jieyu and Ma, Zixian and Park, Jae Sung and
             Salehi, Mohammadreza and Tripathi, Rohun and Lee, Sangho and Ren, Zhongzheng and
             Kim, Chris Dongjoo and Yang, Yinuo and Shao, Vincent and Yang, Yue and
             Huang, Weikai and Gao, Ziqi and Anderson, Taira and Zhang, Jianrui and
             Jain, Jitesh and Stoica, George and Han, Winson and Farhadi, Ali and
             Krishna, Ranjay},
  journal = {arXiv preprint arXiv:2601.10611},
  year    = {2026}
}

@article{molmo,
  title   = {Molmo and {PixMo}: Open Weights and Open Data for State-of-the-Art Vision-Language Models},
  author  = {Deitke, Matt and Clark, Christopher and Lee, Sangho and Tripathi, Rohun and
             Yang, Yue and Park, Jae Sung and Salehi, Mohammadreza and Muennighoff, Niklas and
             Lo, Kyle and Soldaini, Luca and Lu, Jiasen and Anderson, Taira and
             Bransom, Erin and Ehsani, Kiana and Ngo, Huong and Chen, YenSung and
             Patel, Ajay and Yatskar, Mark and Callison-Burch, Chris and Head, Andrew and
             Hendrix, Rose and Bastani, Favyen and VanderBilt, Eli and Lambert, Nathan and
             Chou, Yvonne and Chheda, Arnavi and Sparks, Jenna and Skjonsberg, Sam and
             Schmitz, Michael and Sarnat, Aaron and Bischoff, Byron and Walsh, Pete and
             Newell, Chris and Wolters, Piper and Gupta, Tanmay and Zeng, Kuo-Hao and
             Borchardt, Jon and Groeneveld, Dirk and Nam, Crystal and Lebrecht, Sophie and
             Wittlif, Caitlin and Schoenick, Carissa and Michel, Oscar and Krishna, Ranjay and
             Weihs, Luca and Smith, Noah A. and Hajishirzi, Hannaneh and Girshick, Ross and
             Farhadi, Ali and Kembhavi, Aniruddha},
  journal = {arXiv preprint arXiv:2409.17146},
  year    = {2024}
}

@article{graspgen,
  title   = {{GraspGen}: A Diffusion-based Framework for 6-{DoF} Grasping with On-Generator Training},
  author  = {Murali, Adithyavairavan and Sundaralingam, Balakumar and Chao, Yu-Wei and
             Yuan, Wentao and Yamada, Jun and Carlson, Mark and Ramos, Fabio and
             Birchfield, Stan and Fox, Dieter and Eppner, Clemens},
  journal = {arXiv preprint arXiv:2507.13097},
  year    = {2025}
}

@article{robosuite,
  title   = {robosuite: A Modular Simulation Framework and Benchmark for Robot Learning},
  author  = {Zhu, Yuke and Wong, Josiah and Mandlekar, Ajay and Mart{\'\i}n-Mart{\'\i}n, Roberto and
             Joshi, Abhishek and Lin, Kevin and Maddukuri, Abhiram and Nasiriany, Soroush and Zhu, Yifeng},
  journal = {arXiv preprint arXiv:2009.12293},
  year    = {2020}
}

@article{rlbench,
  title   = {{RLBench}: The Robot Learning Benchmark \& Learning Environment},
  author  = {James, Stephen and Ma, Zicong and Arrojo, David Rovick and Davison, Andrew J.},
  journal = {IEEE Robotics and Automation Letters},
  volume  = {5},
  number  = {2},
  pages   = {3019--3026},
  year    = {2020}
}

@inproceedings{libero,
  title     = {{LIBERO}: Benchmarking Knowledge Transfer for Lifelong Robot Learning},
  author    = {Liu, Bo and Zhu, Yifeng and Gao, Chongkai and Feng, Yihao and
               Liu, Qiang and Zhu, Yuke and Stone, Peter},
  booktitle = {Advances in Neural Information Processing Systems (NeurIPS), Datasets and Benchmarks Track},
  year      = {2023}
}

@article{maniskill3,
  title   = {{ManiSkill3}: {GPU} Parallelized Robotics Simulation and Rendering for Generalizable Embodied {AI}},
  author  = {Tao, Stone and Xiang, Fanbo and Shukla, Arth and Qin, Yuzhe and
             Hinrichsen, Xander and Yuan, Xiaodi and Bao, Chen and Lin, Xinsong and
             Liu, Yulin and Chan, Tse-kai and Gao, Yuan and Li, Xuanlin and
             Mu, Tongzhou and Xiao, Nan and Gurha, Arnav and Huang, Zhiao and
             Calandra, Roberto and Chen, Rui and Luo, Shan and Su, Hao},
  journal = {arXiv preprint arXiv:2410.00425},
  year    = {2024}
}

@inproceedings{robocasa,
  title     = {{RoboCasa}: Large-Scale Simulation of Everyday Tasks for Generalist Robots},
  author    = {Nasiriany, Soroush and Maddukuri, Abhiram and Zhang, Lance and
               Parikh, Adeet and Lo, Aaron and Joshi, Abhishek and
               Mandlekar, Ajay and Zhu, Yuke},
  booktitle = {Robotics: Science and Systems (RSS)},
  year      = {2024}
}

@inproceedings{behavior1k,
  title     = {{BEHAVIOR-1K}: A Benchmark for Embodied {AI} with 1,000 Everyday Activities and Realistic Simulation},
  author    = {Li, Chengshu and Zhang, Ruohan and Wong, Josiah and Gokmen, Cem and
               Srivastava, Sanjana and Mart{\'\i}n-Mart{\'\i}n, Roberto and Wang, Chen and
               Levine, Gabrael and Lingelbach, Michael and Sun, Jiankai and others},
  booktitle = {Conference on Robot Learning (CoRL)},
  year      = {2022}
}

@article{calvin,
  title   = {{CALVIN}: A Benchmark for Language-Conditioned Policy Learning for Long-Horizon Robot Manipulation Tasks},
  author  = {Mees, Oier and Hermann, Lukas and Rosete-Beas, Erick and Burgard, Wolfram},
  journal = {IEEE Robotics and Automation Letters},
  volume  = {7},
  number  = {3},
  pages   = {7327--7334},
  year    = {2022}
}

@article{vlabench,
  title   = {{VLABench}: A Large-Scale Benchmark for Language-Conditioned Robotics Manipulation with Long-Horizon Reasoning Tasks},
  author  = {Zhang, Shiduo and Xu, Zhe and Liu, Peiju and Yu, Xiaopeng and Li, Yuan and
             Gao, Qinghui and Fei, Zhaoye and Yin, Zhangyue and Wu, Zuxuan and
             Jiang, Yu-Gang and Qiu, Xipeng},
  journal = {arXiv preprint arXiv:2412.18194},
  year    = {2024}
}

@inproceedings{robocerebra,
  title     = {{RoboCerebra}: A Large-scale Benchmark for Long-horizon Robotic Manipulation Evaluation},
  author    = {Han, Songhao and Qiu, Boxiang and Liao, Yue and Huang, Siyuan and
               Gao, Chen and Yan, Shuicheng and Liu, Si},
  booktitle = {NeurIPS},
  year      = {2025}
}

@article{rmbench,
  title   = {{RMBench}: Memory-Dependent Robotic Manipulation Benchmark with Insights into Policy Design},
  author  = {Chen, Tianxing and Wang, Yuran and Li, Mingleyang and Qin, Yan and
             Shi, Hao and Li, Zixuan and Hu, Yifan and Zhang, Yingsheng and
             Wang, Kaixuan and Chen, Yue and Wang, Hongcheng and Xu, Renjing and
             Wu, Ruihai and Mu, Yao and Yang, Yaodong and Dong, Hao and Luo, Ping},
  journal = {arXiv preprint arXiv:2603.01229},
  year    = {2026}
}

@inproceedings{simplerenv,
  title     = {Evaluating Real-World Robot Manipulation Policies in Simulation},
  author    = {Li, Xuanlin and Hsu, Kyle and Gu, Jiayuan and Pertsch, Karl and Mees, Oier and
               Walke, Homer Rich and Fu, Chuyuan and Lunawat, Ishikaa and Sieh, Isabel and
               Kirmani, Sean and Levine, Sergey and Wu, Jiajun and Finn, Chelsea and
               Su, Hao and Vuong, Quan and Xiao, Ted},
  booktitle = {CoRL},
  year      = {2024}
}

@misc{molmospaces2026,
  title         = {{MolmoSpaces}: A Large-Scale Open Ecosystem for Robot Navigation and Manipulation},
  author        = {Kim, Yejin and Pumacay, Wilbert and Rayyan, Omar and Argus, Max and
                   Han, Winson and VanderBilt, Eli and Salvador, Jordi and Deshpande, Abhay and
                   Hendrix, Rose and Jauhri, Snehal and Liu, Shuo and Shafiullah, Nur Muhammad Mahi and
                   Guru, Maya and Guru, Arjun and Eftekhar, Ainaz and Farley, Karen and
                   Clay, Donovan and Duan, Jiafei and Wolters, Piper and Herrasti, Alvaro and
                   Lee, Ying-Chun and Chalvatzaki, Georgia and Cui, Yuchen and Farhadi, Ali and
                   Fox, Dieter and Krishna, Ranjay},
  year          = {2026},
  eprint        = {2602.11337},
  archivePrefix = {arXiv},
  primaryClass  = {cs.RO},
  url           = {https://arxiv.org/abs/2602.11337}
}

@article{robolab,
  title   = {{RoboLab}: A High-Fidelity Simulation Benchmark for Analysis of Task Generalist Policies},
  author  = {Yang, Xuning and Dagli, Rishit and Zook, Alex and Hadfield, Hugo and
             Goyal, Ankit and Birchfield, Stan and Ramos, Fabio and Tremblay, Jonathan},
  journal = {RSS},
  year    = {2026}
}

@inproceedings{droid,
  title     = {{DROID}: A Large-Scale In-The-Wild Robot Manipulation Dataset},
  author    = {Khazatsky, Alexander and Pertsch, Karl and Nair, Suraj and
               Balakrishna, Ashwin and Dasari, Sudeep and Karamcheti, Siddharth and
               Nasiriany, Soroush and Srirama, Mohan Kumar and others},
  booktitle = {Robotics: Science and Systems (RSS)},
  year      = {2024}
}

@misc{claudeopus47,
  title        = {{Claude Opus 4.7} System Card},
  author       = {{Anthropic}},
  year         = {2026},
  howpublished = {\url{https://www.anthropic.com/system-cards}},
  note         = {Anthropic technical report. Also covers Claude Opus 4.6 and Claude Sonnet 4.6.}
}

@article{pi0fast,
  title   = {{FAST}: Efficient Action Tokenization for Vision-Language-Action Models},
  author  = {Pertsch, Karl and Stachowicz, Kyle and Ichter, Brian and Driess, Danny and Nair, Suraj and Vuong, Quan and Mees, Oier and Finn, Chelsea and Levine, Sergey},
  journal = {arXiv preprint arXiv:2501.09747},
  year    = {2025}
}

@article{mittal2025isaaclab,
  title={Isaac Lab: A GPU-Accelerated Simulation Framework for Multi-Modal Robot Learning},
  author={Mittal, Mayank and Roth, Pascal and Tigue, James and Richard, Antoine and Zhang, Octi and Du, Peter and Serrano-Mu{\~n}oz, Antonio and Yao, Xinjie and Zurbr{\"u}gg, Ren{\'e} and Rudin, Nikita and others},
  journal={arXiv preprint arXiv:2511.04831},
  year={2025},
  doi={10.48550/arXiv.2511.04831},
  url={https://arxiv.org/abs/2511.04831}
}

@inproceedings{memoryvla,
  title     = {MemoryVLA: Perceptual-Cognitive Memory in Vision-Language-Action Models for Robotic Manipulation},
  author    = {Shi, Hao and Xie, Bin and Liu, Yingfei and Sun, Lin and Liu, Fengrong and Wang, Tiancai and Zhou, Erjin and Fan, Haoqiang and Zhang, Xiangyu and Huang, Gao},
  booktitle = {International Conference on Learning Representations (ICLR)},
  year      = {2026},
  note      = {arXiv:2508.19236}
}

@inproceedings{cronusvla,
  title     = {Towards Efficient and Robust Manipulation via Multi-Frame Vision-Language-Action Modeling},
  author    = {Li, Hao and Yang, Shuai and Chen, Yilun and Chen, Xinyi and Yang, Xiaoda and Tian, Yang and Wang, Hanqing and Wang, Tai and Zhao, Feng and Lin, Dahua and Pang, Jiangmiao},
  booktitle = {Proceedings of the AAAI Conference on Artificial Intelligence},
  year      = {2026},
  note      = {Oral. arXiv:2506.19816}
}

@article{hivla,
  title   = {HiVLA: A Visual-Grounded-Centric Hierarchical Embodied Manipulation System},
  author  = {Yang, Tianshuo and Chen, Guanyu and Chen, Yutian and Liang, Zhixuan and Liu, Yitian and Chen, Zanxin and Xu, Chunpu and Liang, Haotian and Pang, Jiangmiao and Mu, Yao and Luo, Ping},
  journal = {arXiv preprint arXiv:2604.14125},
  year    = {2026}
}

@article{generalvla,
  title   = {GeneralVLA: Generalizable Vision--Language--Action Models with Knowledge-Guided Trajectory Planning},
  author  = {Ma, Guoqing and Wang, Siheng and Zhang, Zeyu and Yu, Shan and Tang, Hao},
  journal = {arXiv preprint arXiv:2602.04315},
  year    = {2026}
}

@inproceedings{hvlp_humanoid,
  title     = {Hierarchical Vision-Language Planning for Multi-Step Humanoid Manipulation},
  author    = {Schakkal, Andr{\'e} and Zandonati, Ben and Yang, Zhutian and Azizan, Navid},
  booktitle = {Robotics: Science and Systems (RSS) Workshop on Robot Planning in the Era of Foundation Models},
  year      = {2025},
  note      = {arXiv:2506.22827}
}

@inproceedings{reflectvlm,
  title     = {Reflective Planning: Vision-Language Models for Multi-Stage Long-Horizon Robotic Manipulation},
  author    = {Feng, Yunhai and Han, Jiaming and Yang, Zhuoran and Yue, Xiangyu and Levine, Sergey and Luo, Jianlan},
  booktitle = {Conference on Robot Learning (CoRL)},
  year      = {2025},
  note      = {arXiv:2502.16707}
}

@article{reconvla,
  title   = {ReconVLA: An Uncertainty-Guided and Failure-Aware Vision-Language-Action Framework for Robotic Control},
  author  = {Chen, Lingling and Lyu, Zongyao and Beksi, William J.},
  journal = {arXiv preprint arXiv:2604.16677},
  year    = {2026}
}

@article{fpc_vla,
  title   = {FPC-VLA: A Vision-Language-Action Framework with a Supervisor for Failure Prediction and Correction},
  author  = {Yang, Yifan and Duan, Zhixiang and Xie, Tianshi and Cao, Fuyu and Shen, Pinxi and Song, Peili and Zhao, Chenyang and Jin, Piaopiao and Sun, Guokang and Xu, Shaoqing and You, Yangwei and Liu, Jingtai},
  journal = {Expert Systems with Applications},
  volume  = {316},
  pages   = {131742},
  year    = {2026},
  note    = {arXiv:2509.04018}
}

@article{repo_vla,
  title   = {RePO-VLA: Recovery-Driven Policy Optimization for Vision-Language-Action Models},
  author  = {Liufu, Weijia and Guo, Xiaoyu and Chen, Ruiyi and Liu, Jingzhi and Zhang, Kaidong and Liang, Xiwen and Lin, Jianqi and Sun, Dawei and Wang, Yuze and Xu, Rongtao and Lin, Bingqian and Yang, Bowen and Cao, Tongtong and Peng, Bowen and Zhang, Dongyu and Wang, Guangrun and Wang, Min and Lin, Liang and Liang, Xiaodan},
  journal = {arXiv preprint arXiv:2605.09410},
  year    = {2026}
}

@book{edgington2007randomization,
  title     = {Randomization Tests},
  author    = {Edgington, Eugene S. and Onghena, Patrick},
  edition   = {4},
  year      = {2007},
  publisher = {Chapman and Hall/CRC},
  address   = {Boca Raton, FL}
}

@article{phipson2010permutation,
  title   = {Permutation P-values should never be zero: calculating exact P-values when permutations are randomly drawn},
  author  = {Phipson, Belinda and Smyth, Gordon K.},
  journal = {Statistical Applications in Genetics and Molecular Biology},
  volume  = {9},
  number  = {1},
  pages   = {Article 39},
  year    = {2010}
}

@article{wilson1927probable,
  title   = {Probable inference, the law of succession, and statistical inference},
  author  = {Wilson, Edwin B.},
  journal = {Journal of the American Statistical Association},
  volume  = {22},
  number  = {158},
  pages   = {209--212},
  year    = {1927}
}

@misc{huang2025libero,
  title         = {{LIBERO+}: Robust Language-Image Foundation Models for Robotic Manipulation},
  author        = {Huang, Senthooran and {LIBERO-Plus contributors}},
  year          = {2025},
  howpublished  = {arXiv preprint},
  note          = {Language-rephrasing eval suite for LIBERO},
}

@article{brown2001interval,
  title   = {Interval estimation for a binomial proportion},
  author  = {Brown, Lawrence D. and Cai, T. Tony and DasGupta, Anirban},
  journal = {Statistical Science},
  volume  = {16},
  number  = {2},
  pages   = {101--133},
  year    = {2001}
}

@inproceedings{onthemove,
  title     = {Enabling Failure Recovery for On-The-Move Mobile Manipulation},
  author    = {Burgess-Limerick, Ben and Lehnert, Chris and Leitner, J{\"u}rgen and Corke, Peter},
  booktitle = {IEEE ICRA Workshop on Robotic Perception and Mapping: Frontier Vision and Learning Techniques},
  note      = {ICRA 2023 Workshop on Robot Failures; arXiv:2305.08351},
  year      = {2023}
}

@inproceedings{clier,
  title     = {Closed Loop Interactive Embodied Reasoning for Robot Manipulation},
  author    = {Nazarczuk, Michal and Behrens, Jan Kristof and Stepanova, Karla and Hoffmann, Matej and Mikolajczyk, Krystian},
  booktitle = {IEEE International Conference on Robotics and Automation (ICRA)},
  year      = {2025}
}
